\documentclass{article} 
\usepackage{iclr2027_conference,times}

\usepackage{amsmath,amsfonts,bm}

\def\eqref#1{equation~\ref{#1}}

\def\1{\bm{1}}

\DeclareMathAlphabet{\mathsfit}{\encodingdefault}{\sfdefault}{m}{sl}
\SetMathAlphabet{\mathsfit}{bold}{\encodingdefault}{\sfdefault}{bx}{n}

\usepackage{graphicx}
\usepackage{subcaption}
\DeclareCaptionFont{scalepanel}{\fontsize{8}{9.5}\selectfont}
\usepackage{placeins}
\usepackage{booktabs}
\PassOptionsToPackage{hyphens}{url}
\usepackage[hidelinks]{hyperref}
\usepackage{xurl}
\usepackage{xcolor}

\newcommand{\pmstd}[1]{\,{\scriptstyle\pm\,#1}}

\definecolor{redblue}{RGB}{142,26,196}

\definecolor{haokcolor}{HTML}{00796B}

\title{Delta-Matching: Closing the Final Gap of\\Native 8-bit Training for LLMs}

\author{Haozhan Tang\thanks{Correspondence to: Haozhan Tang \textless\href{mailto:haozhant@andrew.cmu.edu}{\texttt{haozhant@andrew.cmu.edu}}\textgreater.} \\
Carnegie Mellon University \\
\And
Hao Kang \\
Carnegie Mellon University \\
\And
Han Cai \\
NVIDIA \\
\And
Song Han \\
NVIDIA \& Massachusetts Institute of Technology \\
\And
Chenyan Xiong \\
Carnegie Mellon University \\
}

\iclrfinalcopy
\begin{document}

\maketitle
\fancyhead{}
\pagestyle{fancy}
\thispagestyle{fancy}

\begin{abstract}
Reliable FP8 attention remains a barrier to fully native 8-bit large language model training.
We derive how forward-backward inconsistencies produce stale delta and empirically show how it distorts training dynamics.
Our stale-delta hybrid runs show a modest loss gap at 569M parameters but substantial loss increases and downstream degradation at 1.67B and 5.29B.
QK normalization, NoPE (no positional encoding), and lower-learning-rate context extension mitigate or delay degradation without eliminating it.
This pattern suggests accumulated optimization error that smaller models and short runs can conceal.
We propose Delta-Matching, proving that it restores the softmax gradient's zero-row-sum invariant under the stated numerical assumptions.
It enables native block-scaled FP8 in every forward and backward attention-core matmul without architectural changes, smaller global batches, or auxiliary forward outputs.
Across tested architectures, scales, and training stages, Delta-Matching matches BF16/FP32 mixed-precision training loss and overall downstream performance.
We will release our implementation, trained models, and data recipes\footnote{Project Page:
\url{https://jet-ai-projects.github.io/delta-matching}.}.
\end{abstract}

\begin{figure}[!htbp]
\centering
\begin{minipage}[c]{0.53\textwidth}
  \centering
  \includegraphics[width=\linewidth]{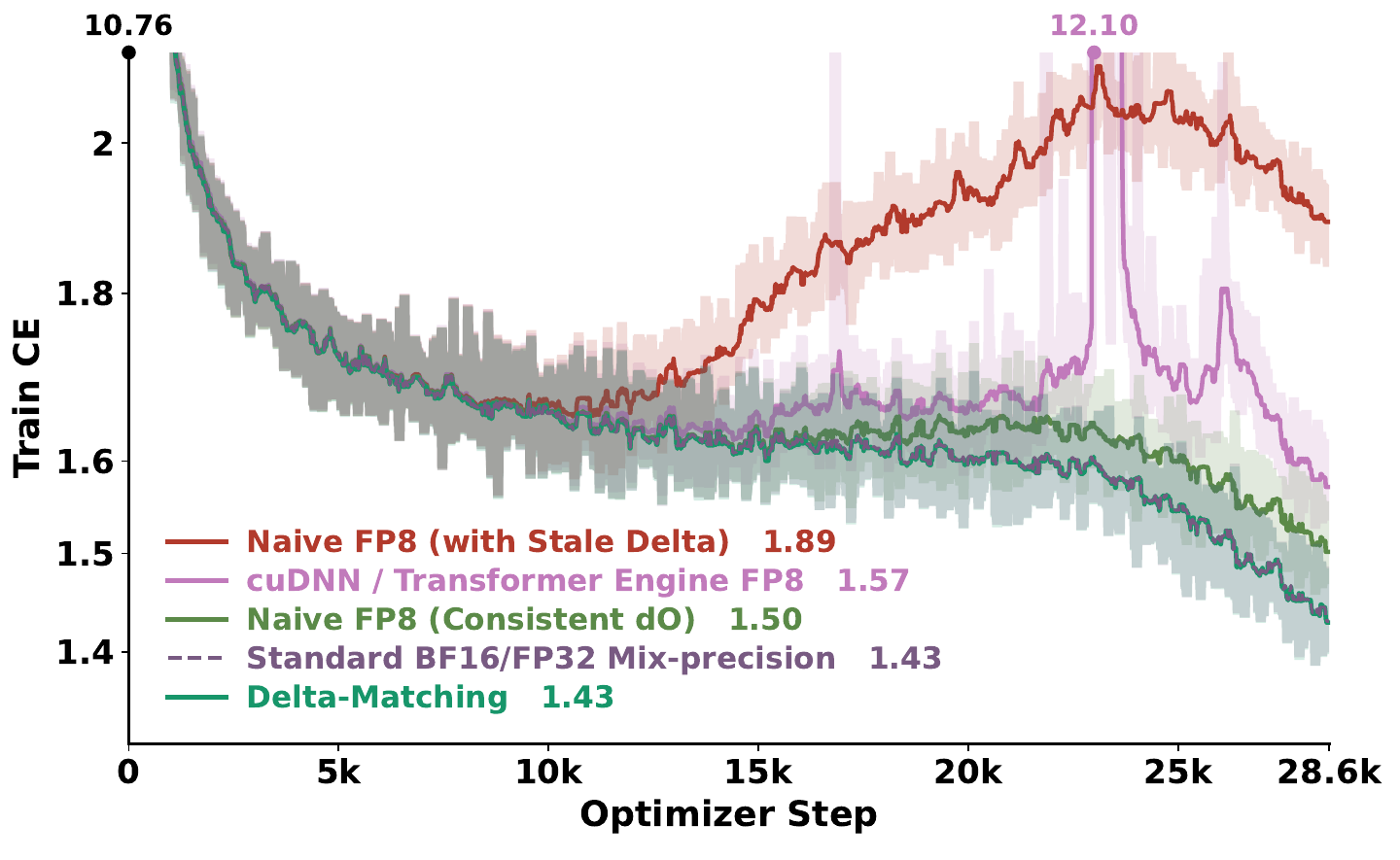}
\end{minipage}%
\begin{minipage}[c]{0.39\textwidth}
  \centering
  \includegraphics[width=\linewidth]{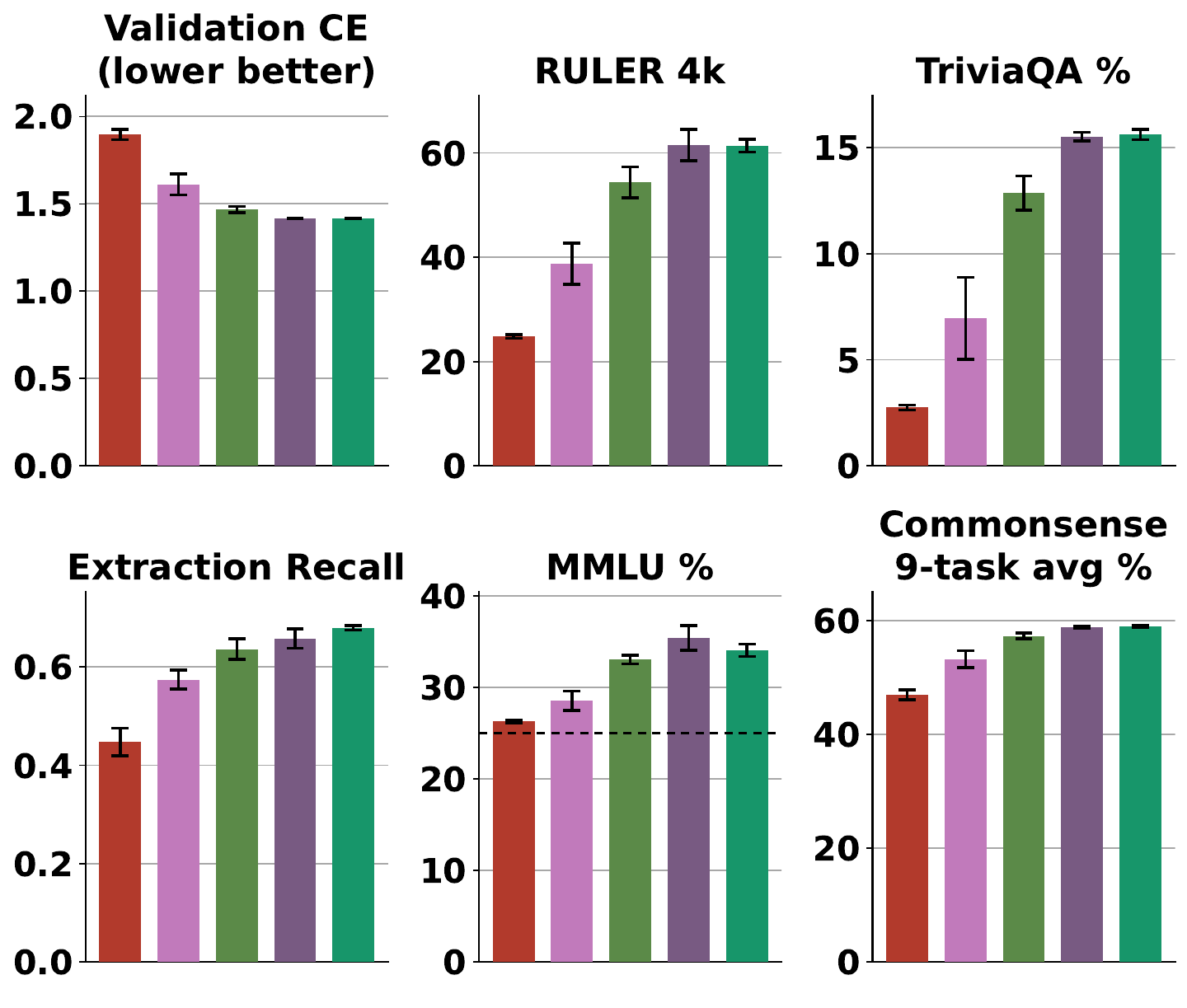}
\end{minipage}
\caption{
A 1.67B GDN/GQA hybrid trained on 30B Nemotron-CC tokens.
\textbf{Left.} Delta-Matching tracks BF16/FP32 training loss; FP8 baselines degrade.
Lines show 250-step rolling medians with batch min--max bands on a zoomed logarithmic scale.
\textbf{Right.} Delta-Matching matches BF16/FP32 validation loss and overall downstream performance.
Section~\ref{sec:exp_setup} details training, evaluation, and plotting conventions.
}

\label{fig:teaser}
\end{figure}
\section{Introduction}
\label{sec:introduction}

FP8 promises faster, more memory-efficient large language model (LLM) training, with roughly twice BF16's peak Tensor Core throughput on Hopper~\citep{nvidia2023hopper}.
Existing systems quantize training-state storage~\citep{peng2023fp8,xi2025coat} and linear-layer matmuls in both passes~\citep{liu2024deepseek}.
Despite effective low-precision inference~\citep{zhang2025sageattention,shah2024flashattention3} and native FP8 training kernels~\citep{nvidiateattention,nvidiacudnnattention,nvidia2026cudnnmxfp8blog}, reliable attention training across architectures, tokens per step, scales, and durations remains unresolved.

We trace an FP8 attention failure to a saved-output shortcut for the backward row correction~\citep{dao2024flashattention2}.
Separately rounded forward and backward operands can invalidate this shortcut, producing \emph{stale delta} that violates the softmax score gradient's zero-row-sum invariant (Section~\ref{sec:violate_invariant}).
Residuals can sharpen or flatten rows; despite near cancellation across rows, we observe persistent pressure on selected channel gains (Section~\ref{sec:delta_training_dynamics}).

We propose \textbf{Delta-Matching}, deriving the correction from unquantized probabilities and FP32 products of FP8 backward operands to restore the invariant under the stated numerical assumptions (Section~\ref{sec:delta_matching}).
Matching output-gradient precision alone leaves a probability-quantization inconsistency that still degrades training (Section~\ref{sec:exp_cmp}).
Delta-Matching enables native block-scaled FP8 in every forward and backward attention-core matmul, without architectural changes, a smaller global batch, or an auxiliary forward output.

The failure persists in hybrids pairing Gated DeltaNet (GDN) with multi-head latent attention (MLA)~\citep{yang2025gdn,liu2024deepseek}, and Kimi Delta Attention (KDA) with grouped-query attention (GQA)~\citep{team2025kimilinear,ainslie2023gqa}.
QK normalization, proposed to mitigate related BF16 instability~\citep{qiu2026whybf16fa}, prevents catastrophic collapse in our FP8 GDN/GQA hybrid but leaves a substantial performance gap.
Increasing head dimension from 128 to 256 does not help; removing attention's positional encoding (NoPE) while retaining position-sensitive recurrent layers only delays degradation (Sections~\ref{sec:exp_attn_arch} and~\ref{sec:exp_model_arch}).

Scale and duration also matter. At 569M parameters, GDN/GQA with stale delta has a modest validation-loss gap, largely preserving downstream performance.
SageBwd pretraining~\citep{zhang2026sagebwd} and the BF16 study by \citet{qiu2026whybf16fa} focus on this sub-billion regime.
At 1.67B and 5.29B, models initially track BF16/FP32 before loss rises and downstream performance degrades, with earlier onset at 5.29B.
During 8K-to-64K context extension at a lower learning rate following the Kimi K2 and DeepSeek-V3 recipe~\citep{team2025kimik2,liu2024deepseek}, the stale-delta baseline's training-loss gap over BF16/FP32 exceeds $0.01$ only after roughly 19{,}000 optimizer steps (20B tokens; Section~\ref{sec:exp_stage}).
These patterns suggest error accumulation over updates, accelerated at larger scales, which small-model or short-run evaluations can obscure.

Figure~\ref{fig:teaser} summarizes 1.67B GDN/GQA pretraining on 30B tokens. We make three contributions.
\begin{itemize}
    \item We identify a stale-delta failure mechanism in FP8 attention training through theoretical analysis and controlled experiments.
    \item We formulate Delta-Matching for native FP8 attention and prove that it restores the softmax-backward invariant under the stated numerical assumptions.
    \item Across tested attention and model architectures, scales, training stages, and linear-layer precisions, Delta-Matching matches BF16/FP32 mixed-precision (MP) attention references in training loss and overall downstream performance.
\end{itemize}

These findings clarify 8-bit training and offer a path to fully native 8-bit LLM training.
We will open-source our implementation, trained models, and data recipes to accelerate LLM research.

\section{Background and Related Work}
\label{sec:background}

\begingroup
\setlength{\intextsep}{8pt plus 1pt minus 1pt}
\makeatletter
\renewcommand{\paragraph}{\@startsection{paragraph}{4}{\z@}
  {0.75ex plus 0.25ex minus 0.1ex}{-1em}{\normalfont\normalsize\bfseries}}
\makeatother

\paragraph{Low-precision training.}
\label{sec:background_training}

FP8 uses OCP-standardized E4M3 and E5M2 formats~\citep{micikevicius2022fp8,micikevicius2023ocp}.
MXFP8's shared E8M0 power-of-two scale for 32 values~\citep{rouhani2023ocpmx,rouhani2023microscaling} makes E4M3 practical in both passes~\citep{mishra2025recipes}.
H100 SXM peaks at 1,979 TFLOPS in FP8 versus 989 TFLOPS in BF16~\citep{nvidia2023hopper}, and B200 reaches 4.5 versus 2.2 PFLOPS~\citep{nvidia2025blackwell}.
Blackwell adds native MXFP8/NVFP4 block scaling~\citep{nvidia_ptx_blockscaling}; NVFP4 offers $2\times$ FP8 throughput, rising to $3\times$ on Blackwell Ultra~\citep{nvidia2025blackwellultra}.
Preliminary Rubin specifications list up to 17.5, 35, and 4 PFLOPS for FP8, NVFP4, and BF16, respectively~\citep{nvidia2026rubinnvl72}.
These are peak dense Tensor Core rates. Transformer Engine (TE) reports a roughly $1.7\times$ FP8 GEMM speedup on Hopper; higher-precision operators limit overall gains~\citep{nvidia_te_speedups}.

FP8-LM and COAT quantize training-state storage~\citep{peng2023fp8,xi2025coat}, while DeepSeek-V3 and Ling-1T apply FP8 to linear-layer GEMMs~\citep{liu2024deepseek,team2025every}.
NVIDIA studies simulated MXFP8~\citep{mishra2025recipes} and NVFP4 pretraining~\citep{abecassis2025pretraining,nvidia2025nemotron3}; Kimi K3 uses MXFP4 for routed-expert weights during post-training~\citep{team2026kimi}.
Appendix~\ref{app:low_precision_train} details training scales, precision exceptions, and further hardware specifications.
These systems use higher-precision attention cores (Figure~\ref{fig:train_pipe_precision}).

\paragraph{Low-precision attention.}
\label{sec:background_attention}

SageAttention's default accurate inference path uses INT8 $QK^\top$ with FP16 $PV$~\citep{zhang2025sageattention}; SageAttention2 combines INT4 $Q,K$ with FP8 $P,V$~\citep{zhang2024sageattention2}; FlashAttention-3 provides an FP8 forward-only kernel on Hopper~\citep{shah2024flashattention3}.

Training remains unsettled (Figure~\ref{fig:attention_core}).
FlashAttention-4 evaluates FP16/BF16 forward and backward kernels~\citep{zadouri2026flashattention4}.
SageAttention3 matches full precision on tested fine-tuning tasks~\citep{zhang2026sageattention3}; SageBwd matches it on a 325M model at 260K tokens per optimizer step but retains a loss gap at 2.1M~\citep{zhang2026sagebwd}.
Other approaches modify the architecture~\citep{hernandez2026towards}, retain selected products in BF16~\citep{ding2026fullfp4}, or evaluate emulated training only through language-modeling loss~\citep{rouhani2023microscaling}.
\citet{qiu2026whybf16fa} identify saved-output rounding errors in BF16/FP32 training and test the direct backward row correction as a stability intervention; related work studies early-warning monitors~\citep{huang2026monitors}.
Attn-QAT uses forward NVFP4 fake quantization and an auxiliary output for a consistent BF16 backward correction~\citep{zhang2026attnqat}.
Despite FP8/MXFP8 kernels for both passes~\citep{nvidiateattention,nvidiacudnnattention,nvidia2026cudnnmxfp8blog}, reliable native attention-core training remains understudied across hybrid architectures, tokens per optimizer step, and parameter scales.

\begin{figure}[!htbp]
\centering
\begin{subfigure}[b]{0.27\textwidth}
  \centering
  \includegraphics[width=\linewidth]{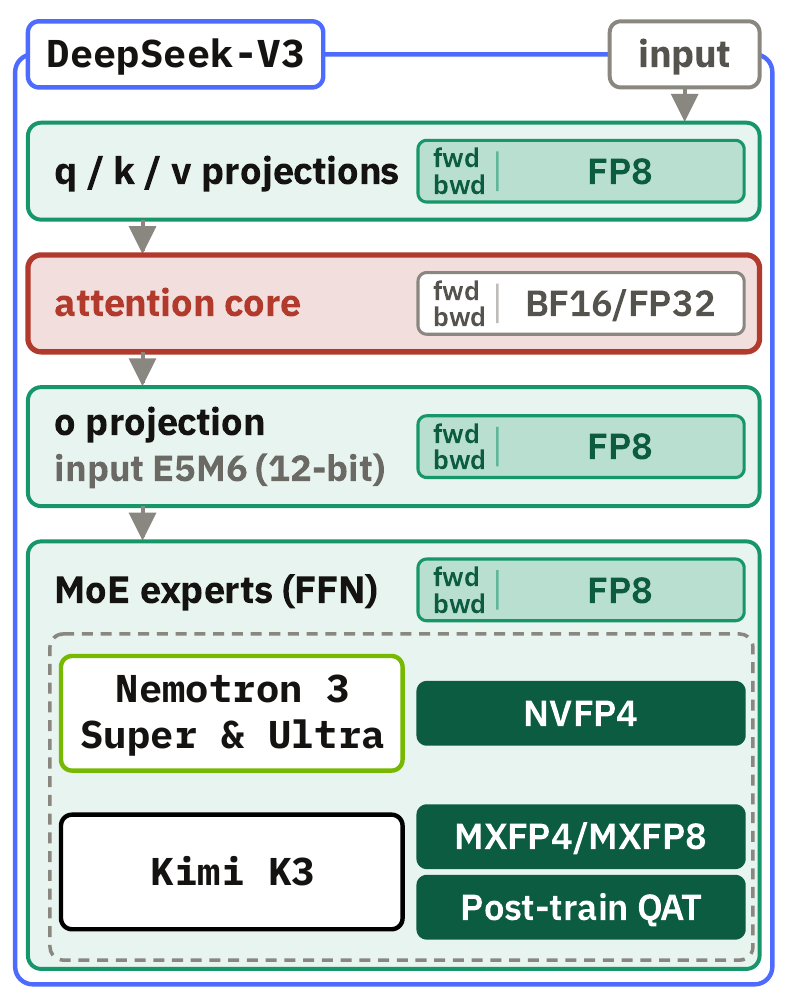}
  \caption{}
  \label{fig:train_pipe_precision}
\end{subfigure}%
\begin{subfigure}[b]{0.61\textwidth}
  \centering
  \includegraphics[width=\linewidth]{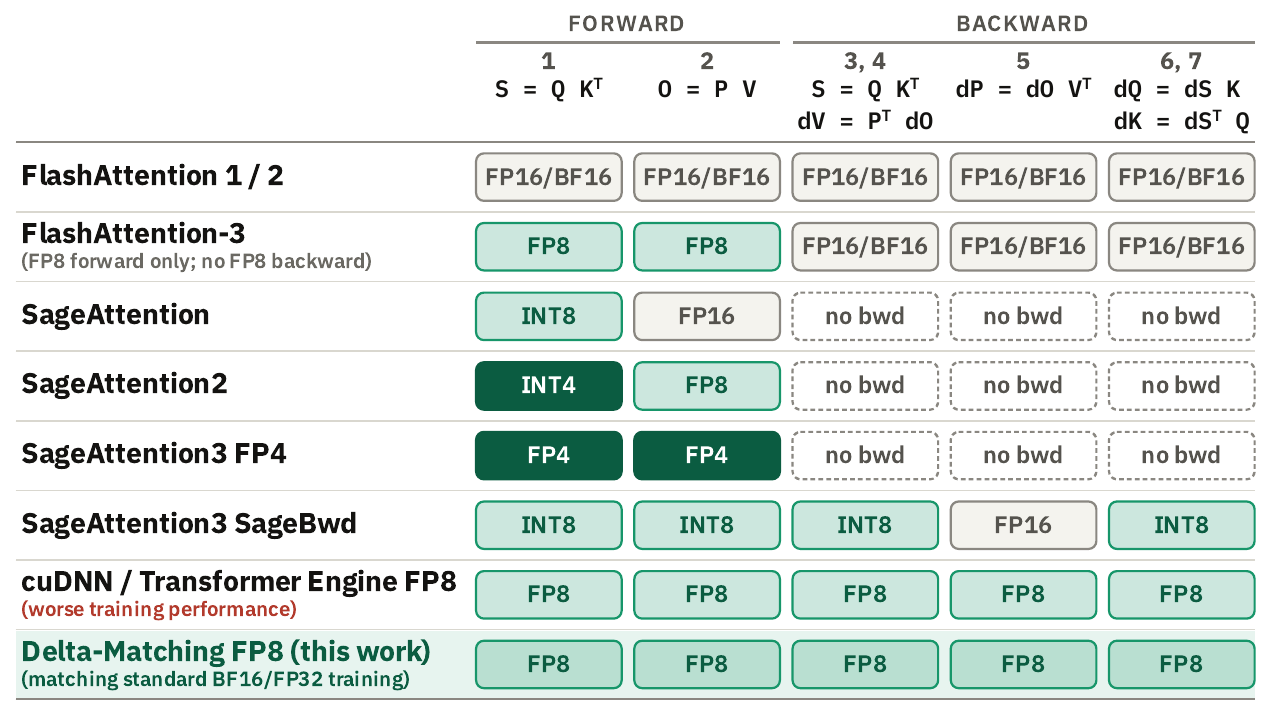}
  \caption{}
  \label{fig:attention_core}
\end{subfigure}
\caption{(a) DeepSeek-V3 uses FP8 for projections and FFNs in both passes, retaining higher precision for the $O(N^2)$ attention core~\citep{liu2024deepseek,team2025every}. Sub-8-bit recipes (dashed) mainly target FFNs~\citep{nvidia2025nemotron3,team2026kimi}.
(b) Operand precision for seven attention-core matmuls (two forward, five backward). Of the two compared methods quantizing all seven to 8 bits, only Delta-Matching matches BF16/FP32 training.}
\label{fig:background}
\end{figure}

\paragraph{Softmax, linear, and hybrid attention.}
\label{sec:background_attention_arch}

Softmax attention computes probabilities $P$ and output $O$, with $dX$ denoting the loss gradient for $X$.
Appendix~\ref{app:softmax_attention} gives the full forward and backward equations.
Row-wise softmax remains in FP32 for stability~\citep{ding2021cogview,zeng2022glm,qiu2026whybf16fa}.
FlashAttention recomputes $P$ blockwise during backward instead of storing it, using the exact-arithmetic saved-output correction $\delta_i=\langle dO_i,O_i\rangle$~\citep{dao2022flashattention,dao2024flashattention2}.
GQA shares key-value heads across query-head groups~\citep{shazeer2019fast,ainslie2023gqa}, MLA jointly compresses keys and values into latent vectors~\citep{liu2024deepseek}, and QK normalization controls logit growth~\citep{henry2020qknorm,dehghani2023scaling,wortsman2024small}.
Linear attention uses a fixed-size recurrent state~\citep{katharopoulos2020transformers}.
GDN combines gating and delta-rule updates~\citep{yang2025gdn}; related recurrent architectures include Kimi Delta Attention (KDA)~\citep{team2025kimilinear} and Mamba-2~\citep{dao2024mamba2}.
Hybrids interleave recurrent and softmax layers~\citep{blakeman2025nemotronh,qwen2025next,qwen2025nextcard}.
We evaluate GQA-only models and GDN/GQA, GDN/MLA, KDA/GQA, and Mamba-2/GQA hybrids.
Appendix~\ref{app:attention_architectures} provides further architecture details.
\par
\endgroup

\section{The Stale Delta and Delta-Matching}
\label{sec:the_delta}

We derive FP8 violations of the zero-row-sum invariant and formulate Delta-Matching to restore it under the stated assumptions.
Section~\ref{sec:delta_training_dynamics} examines the resulting training dynamics.

\subsection{The Zero-Row-Sum Invariant}
\label{sec:zero_row_sum_invariant}
The softmax backward pass computes $dS=P\odot(dP-\delta)$,
where $\delta_i=\sum_j P_{ij}dP_{ij}$ is broadcast across row $i$.
Since $\sum_j P_{ij}=1$ in exact arithmetic,
\begin{equation}
\label{eq:delta_zero_row_sum}
\sum_j dS_{ij}
=\sum_j P_{ij}(dP_{ij}-\delta_i)
=\delta_i-\delta_i\sum_j P_{ij}=0.
\end{equation}
This expresses softmax shift invariance, since adding a constant to a score row leaves its probabilities unchanged.
It also gives $\sum_j dK_{jc}=0$ across tokens in each channel $c$ for keys entering $S=QK^\top/\sqrt{d}$ (Appendix~\ref{app:proofs}).

\subsection{Violation of the Invariant under Quantization}
\label{sec:violate_invariant}
Let $P$ and $dO$ denote FP32 softmax probabilities and BF16 output gradients before FP8 quantization.
Hats denote dequantized FP8 operands, including block scales; the products $\widehat{O}=\widehat{P}\widehat{V}$ and $\widehat{dP}=\widehat{dO}\widehat{V}^\top$ use FP32 accumulation and output.
Naive FP8 uses FlashAttention's saved-output correction $\delta_i^{\mathrm{stale}}=\langle dO_i,\widehat{O}_i\rangle$~\citep{dao2024flashattention2}.
Retaining BF16 $dO$ avoids its FP8 rounding but mismatches the $\widehat{dP}$ operands.
For normalized $P$, identical effective $\widehat{V}$ in both passes, and no accumulation, output-storage, or reduction rounding, $dS_{ij}^{\mathrm{stale}}=P_{ij}(\widehat{dP}_{ij}-\delta_i^{\mathrm{stale}})$ has row sum

\begin{equation}
\label{eq:delta_quantization_residual}
\sum_j dS_{ij}^{\mathrm{stale}}
=\sum_k\widehat{dO}_{ik}\Big(\sum_j(\underbrace{P_{ij}-\widehat{P}_{ij}}_{\substack{\text{probability}\\\text{quantization error}}})\widehat{V}_{jk}\Big)
+\sum_k(\underbrace{\widehat{dO}_{ik}-dO_{ik}}_{\substack{\text{gradient}\\\text{quantization error}}})\Big(\sum_j\widehat{P}_{ij}\widehat{V}_{jk}\Big).
\end{equation}

The probability-quantization and gradient-quantization errors produce \textbf{stale delta} when they do not cancel (derivation in Appendix~\ref{app:proofs}).
\citet{qiu2026whybf16fa} found a related output-rounding error in smaller-scale BF16/FP32 mixed-precision GPT-2 training.

We probe the FP8 kernel on BF16-recomputed activations from final 1.67B GDN/GQA
checkpoints, covering six attention layers and 32 training windows.
Before $dS$ quantization, stale delta produces normalized row-sum residuals of
$1.1\times10^{-2}$--$4.7\times10^{-2}$, measured as window-averaged row medians
on a common set of nonzero-gradient rows (Appendix~\ref{app:ds_quantization}).

\subsection{Delta-Matching}
\label{sec:delta_matching}
Building on the direct backward correction~\citep{qiu2026whybf16fa}, we formulate \textbf{Delta-Matching} for native FP8 attention.
It uses FP32 probabilities before quantization and FP32 products of FP8 backward operands to correct both mismatches.
Using $\widehat{dO}$ in the saved-output shortcut (\emph{consistent $dO$}) removes only the gradient-quantization term under the assumptions of \eqref{eq:delta_quantization_residual}, leaving
\begin{equation}
\label{eq:delta_consistent_do_residual}
\sum_j P_{ij}\bigl(\widehat{dP}_{ij}-\langle\widehat{dO}_i,\widehat{O}_i\rangle\bigr)
=\sum_k\widehat{dO}_{ik}\Bigl(\sum_j(\underbrace{P_{ij}-\widehat{P}_{ij}}_{\substack{\text{probability}\\\text{quantization error}}})\widehat{V}_{jk}\Bigr).
\end{equation}
The forward output uses $\widehat{P}$ but the score gradient uses $P$, so consistent $dO$ retains a mismatch and still degrades training (Section~\ref{sec:exp_cmp}).

For fixed normalized $P$ and $\widehat{dP}$, zero row sums in $dS_{ij}=P_{ij}(\widehat{dP}_{ij}-c_i)$ uniquely determine $c_i=\delta_i^{\mathrm{matched}}$, with
\begin{equation}
\label{eq:matched_delta_definition}
\delta_i^{\mathrm{matched}}=\sum_j P_{ij}\widehat{dP}_{ij}.
\end{equation}
The resulting gradient satisfies
\begin{equation}
\label{eq:matched_delta_zero_row_sum}
\sum_j dS_{ij}^{\mathrm{matched}}
=\sum_j P_{ij}(\widehat{dP}_{ij}-\delta_i^{\mathrm{matched}})
=\delta_i^{\mathrm{matched}}\Big(1-\sum_j P_{ij}\Big)=0.
\end{equation}

Delta-Matching restores the invariant up to FP32 rounding before $dS$ quantization.
In the same probe, median residuals rise from $1.4\times10^{-7}$--$7.1\times10^{-7}$
to $2.7\times10^{-3}$--$3.4\times10^{-3}$ after FP8 casting with fixed pre-cast normalization.
Casting can introduce a signed offset, but stale-delta residuals remain $5.3$--$6.3\times$ larger across layers
(Appendix~\ref{app:ds_quantization}).
Ordinary FP8 operand errors remain, but the correction needs neither higher-precision attention matmuls nor an auxiliary forward output.
It closely follows BF16/FP32 attention updates without stale-delta sharpening or flattening and suppresses excessive gain growth (Figure~\ref{fig:delta_dynamics}).
With both tested momentum-based optimizers, AdamW and Muon, stale delta
distorts gain dynamics, while Delta-Matching tracks the corresponding BF16/FP32 reference
(Appendix~\ref{app:muon_mechanism}).

Despite recomputation, our kernel achieves lower combined forward and backward latency on H100 than Hopper-optimized BF16 FlashAttention-3~\citep{shah2024flashattention3}.
Appendices~\ref{app:kernel_numerics} and~\ref{app:kernel_throughput} detail kernel numerics and timing; kernel development is not our focus.

The alternative mitigations we test leave training-performance gaps or markedly different dynamics from BF16/FP32 (Appendix~\ref{app:alternative_mitigations}).

\section{Experiments and Discussion}
\label{sec:experiments}

\subsection{Experiment Setup}
\label{sec:exp_setup}

\paragraph{Configuration.}
Our 1.67B-parameter base hybrid combines 18 Gated DeltaNet~\citep{yang2025gdn}
layers and 6 grouped-query attention~\citep{ainslie2023gqa} layers.
We pretrain it on 30B Nemotron-CC tokens~\citep{su2025nemotroncc,blakeman2025nemotron3nano} at 8K context.
Attention uses 16 query and 4 key-value heads of dimension 128, compatible
with the evaluated cuDNN FP8 configuration.
Each main experiment uses two initialization seeds.
Appendices~\ref{app:train_experiments}, \ref{app:data_recipe}, and
\ref{app:conventions_uncertainty} give experiment configurations, data recipes,
and reporting conventions.

\paragraph{Evaluation.}
All methods use BF16 evaluation with FlashAttention-2
(Appendix~\ref{app:eval_configuration}).
Validation cross-entropy (CE) uses held-out Nemotron-CC data.
In-context recall uses regenerated RULER tasks at 4K and 8K~\citep{hsieh2024ruler} and the extraction-recall
suite of~\citet{arora2024simple} over SWDE~\citep{lockard2019openceres},
FDA~\citep{arora2023evaporate}, and SQuAD completion~\citep{rajpurkar2018squad}.
General capabilities use five-shot TriviaQA exact match~\citep{joshi2017triviaqa},
zero-shot MMLU~\citep{hendrycks2020mmlu}, and nine zero-shot commonsense benchmarks,
LAMBADA~\citep{paperno2016lambada}, HellaSwag~\citep{zellers2019hellaswag},
PIQA~\citep{bisk2020piqa}, ARC-Easy and ARC-Challenge~\citep{clark2018arc},
SciQ~\citep{welbl2017sciQ}, OpenBookQA~\citep{mihaylov2018openbookqa},
WinoGrande~\citep{sakaguchi2021winogrande}, and COPA~\citep{roemmele2011commonsensecausalreasoning,wang2019copa}.
CS9 is their unweighted mean.
Appendix~\ref{app:eval} describes the evaluation protocols.
SE denotes standard error; bold marks the best FP8 mean per column within each configuration.

\paragraph{Loss plots.}
Unless otherwise noted, plots show first-seed training CE on a log-scaled $y$-axis zoomed to the post-warmup range, clipping higher values.
Curves and shaded bands show 250-step rolling medians and batch min--max ranges, respectively.
The step-zero dot marks the shared initial CE; legend values give final CE.

\subsection{Delta-Matching against FP8 Baselines and Attention Variants}
\label{sec:exp_cmp}
\label{sec:exp_attn_arch}

We compare cuDNN/TE FP8~\citep{nvidiacudnnattention,nvidia2026cudnnmxfp8blog,nvidiateattention},
naive FP8 with stale delta, consistent $dO$, and Delta-Matching in the default
configuration (Section~\ref{sec:exp_setup}).
The BF16/FP32 MP reference uses BF16 attention inputs, FP32 accumulation~\citep{micikevicius2017mixed},
and FP8 FFNs (P1).
Full FP8 also quantizes attention-core and projection GEMMs (P4),
not other operations or training state (Appendix~\ref{app:train_precision}).
Consistent $dO$ matches the correction's output-gradient precision to the backward pass
but retains the probability mismatch (Section~\ref{sec:delta_matching}).

We also compare BF16/FP32, stale delta, and Delta-Matching on three GDN/GQA
variants (Appendix~\ref{app:train_architectures}).
NoPE removes GQA's RoPE but retains order-sensitive Gated DeltaNet layers;
no QK normalization removes GQA's query and key RMSNorm.
Both use the default precision scopes.
Head dimension 256 preserves total query and KV widths using 8 query and 2 KV heads;
all methods retain BF16 FFNs and attention projections
(P0 versus P2 in Appendix~\ref{app:train_precision}).

\begin{table}[!htbp]
\centering
\caption{
\textbf{FP8 attention training across methods and attention variants.}
Mean $\pm$ SE over two seeds; downstream scores are percentages.
Each configuration has its own BF16/FP32 reference; bold marks its best FP8 mean.
CS9 averages nine individual benchmark accuracies, reported in Appendix~\ref{app:commonsense_benchmarks}.
}
\label{tab:exp_cmp}
\label{tab:exp_cmp_overall}
\label{tab:exp_attn_arch}
\scriptsize
\setlength{\tabcolsep}{1.5pt}
\begin{tabular*}{\linewidth}{@{\extracolsep{\fill}}llccccccc@{}}
\toprule
& & & \multicolumn{3}{c}{In-Context Recall $\uparrow$} & \multicolumn{3}{c}{General Capabilities $\uparrow$} \\
\cmidrule(lr){4-6}\cmidrule(lr){7-9}
Method & Configuration & Validation CE $\downarrow$ & RULER-4K & RULER-8K & Extraction & TriviaQA & MMLU & CS9 \\
\midrule
BF16/FP32 MP & Default & $1.4178\pmstd{0.0004}$ & $61.5\pmstd{3.0}$ & $52.3\pmstd{2.3}$ & $65.7\pmstd{2.0}$ & $15.5\pmstd{0.2}$ & $35.4\pmstd{1.4}$ & $58.8\pmstd{0.2}$ \\
& NoPE & $1.4210\pmstd{0.0000}$ & $58.0\pmstd{3.0}$ & $51.0\pmstd{2.6}$ & $65.4\pmstd{0.8}$ & $16.0\pmstd{0.4}$ & $35.2\pmstd{0.4}$ & $58.5\pmstd{0.1}$ \\
& No QK-norm & $1.4214\pmstd{0.0012}$ & $58.5\pmstd{1.9}$ & $46.5\pmstd{5.2}$ & $65.7\pmstd{0.0}$ & $16.1\pmstd{0.3}$ & $35.0\pmstd{0.7}$ & $59.1\pmstd{0.2}$ \\
& Head 256 & $1.4135\pmstd{0.0003}$ & $62.9\pmstd{1.6}$ & $52.8\pmstd{2.1}$ & $66.2\pmstd{2.4}$ & $16.4\pmstd{0.2}$ & $36.0\pmstd{0.2}$ & $59.4\pmstd{0.8}$ \\
\midrule
cuDNN/TE FP8 & Default & $1.6105\pmstd{0.0600}$ & $38.8\pmstd{3.9}$ & $31.0\pmstd{3.9}$ & $57.4\pmstd{1.9}$ & $\phantom{0}7.0\pmstd{1.9}$ & $28.5\pmstd{1.0}$ & $53.2\pmstd{1.5}$ \\
Naive FP8 (Stale Delta) & Default & $1.8970\pmstd{0.0296}$ & $24.8\pmstd{0.3}$ & $19.8\pmstd{2.1}$ & $44.7\pmstd{2.8}$ & $\phantom{0}2.7\pmstd{0.1}$ & $26.3\pmstd{0.1}$ & $47.0\pmstd{0.9}$ \\
& NoPE & $1.6572\pmstd{0.0973}$ & $38.2\pmstd{2.7}$ & $29.3\pmstd{2.7}$ & $53.3\pmstd{2.7}$ & $\phantom{0}6.3\pmstd{1.8}$ & $28.9\pmstd{1.9}$ & $51.8\pmstd{2.7}$ \\
& No QK-norm & \multicolumn{7}{c}{\textit{Training diverged}} \\
& Head 256 & $2.0709\pmstd{0.0289}$ & $11.5\pmstd{0.2}$ & $\phantom{0}7.7\pmstd{0.7}$ & $29.6\pmstd{3.4}$ & $\phantom{0}1.4\pmstd{0.0}$ & $26.1\pmstd{0.2}$ & $44.0\pmstd{0.4}$ \\
Naive FP8 (Consistent dO) & Default & $1.4667\pmstd{0.0175}$ & $54.4\pmstd{3.0}$ & $43.2\pmstd{2.3}$ & $63.6\pmstd{2.1}$ & $12.9\pmstd{0.8}$ & $33.0\pmstd{0.5}$ & $57.3\pmstd{0.5}$ \\
\midrule
Delta-Matching (\textit{ours}) & Default & $\mathbf{1.4162}\pmstd{0.0006}$ & $\mathbf{61.4}\pmstd{1.2}$ & $\mathbf{47.8}\pmstd{0.5}$ & $\mathbf{68.0}\pmstd{0.5}$ & $\mathbf{15.6}\pmstd{0.2}$ & $\mathbf{34.0}\pmstd{0.7}$ & $\mathbf{59.0}\pmstd{0.2}$ \\
& NoPE & $\mathbf{1.4216}\pmstd{0.0021}$ & $\mathbf{58.6}\pmstd{1.1}$ & $\mathbf{49.7}\pmstd{1.4}$ & $\mathbf{64.7}\pmstd{0.1}$ & $\mathbf{15.7}\pmstd{0.1}$ & $\mathbf{32.7}\pmstd{1.2}$ & $\mathbf{58.9}\pmstd{0.2}$ \\
& No QK-norm & $\mathbf{1.4208}\pmstd{0.0014}$ & $\mathbf{60.5}\pmstd{0.6}$ & $\mathbf{50.3}\pmstd{0.4}$ & $\mathbf{64.7}\pmstd{0.9}$ & $\mathbf{15.9}\pmstd{0.0}$ & $\mathbf{34.5}\pmstd{1.4}$ & $\mathbf{58.9}\pmstd{0.5}$ \\
& Head 256 & $\mathbf{1.4146}\pmstd{0.0014}$ & $\mathbf{62.3}\pmstd{1.2}$ & $\mathbf{53.9}\pmstd{2.9}$ & $\mathbf{66.9}\pmstd{0.8}$ & $\mathbf{16.1}\pmstd{0.2}$ & $\mathbf{34.9}\pmstd{1.7}$ & $\mathbf{58.9}\pmstd{0.2}$ \\
\bottomrule
\end{tabular*}
\end{table}

\begin{figure}[!htbp]
\centering
\begin{subfigure}[b]{0.33\textwidth}
  \centering
  \includegraphics[width=\linewidth]{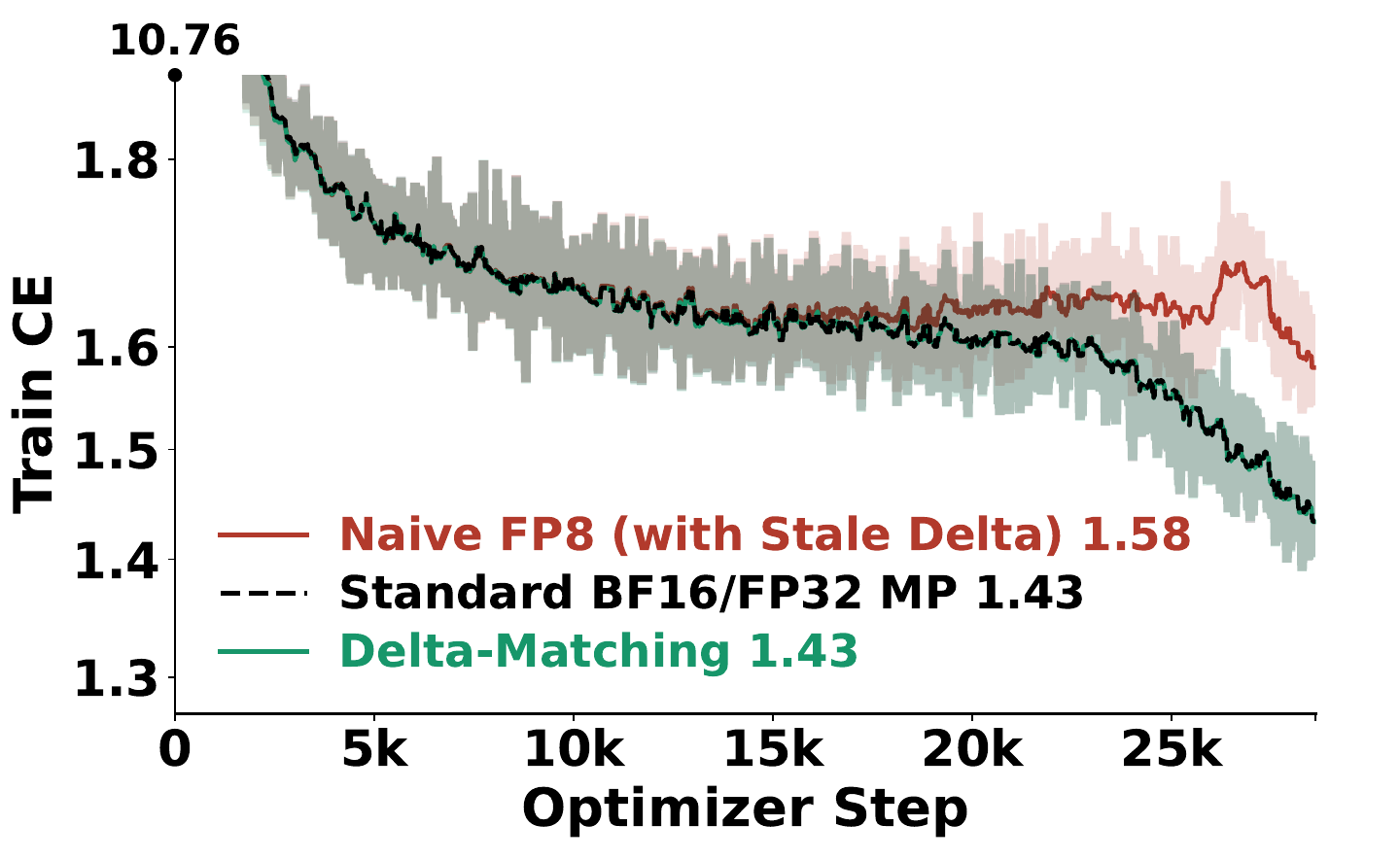}
  \caption{NoPE.}
  \label{fig:exp_attn_nope}
\end{subfigure}\hfill
\begin{subfigure}[b]{0.33\textwidth}
  \centering
  \includegraphics[width=\linewidth]{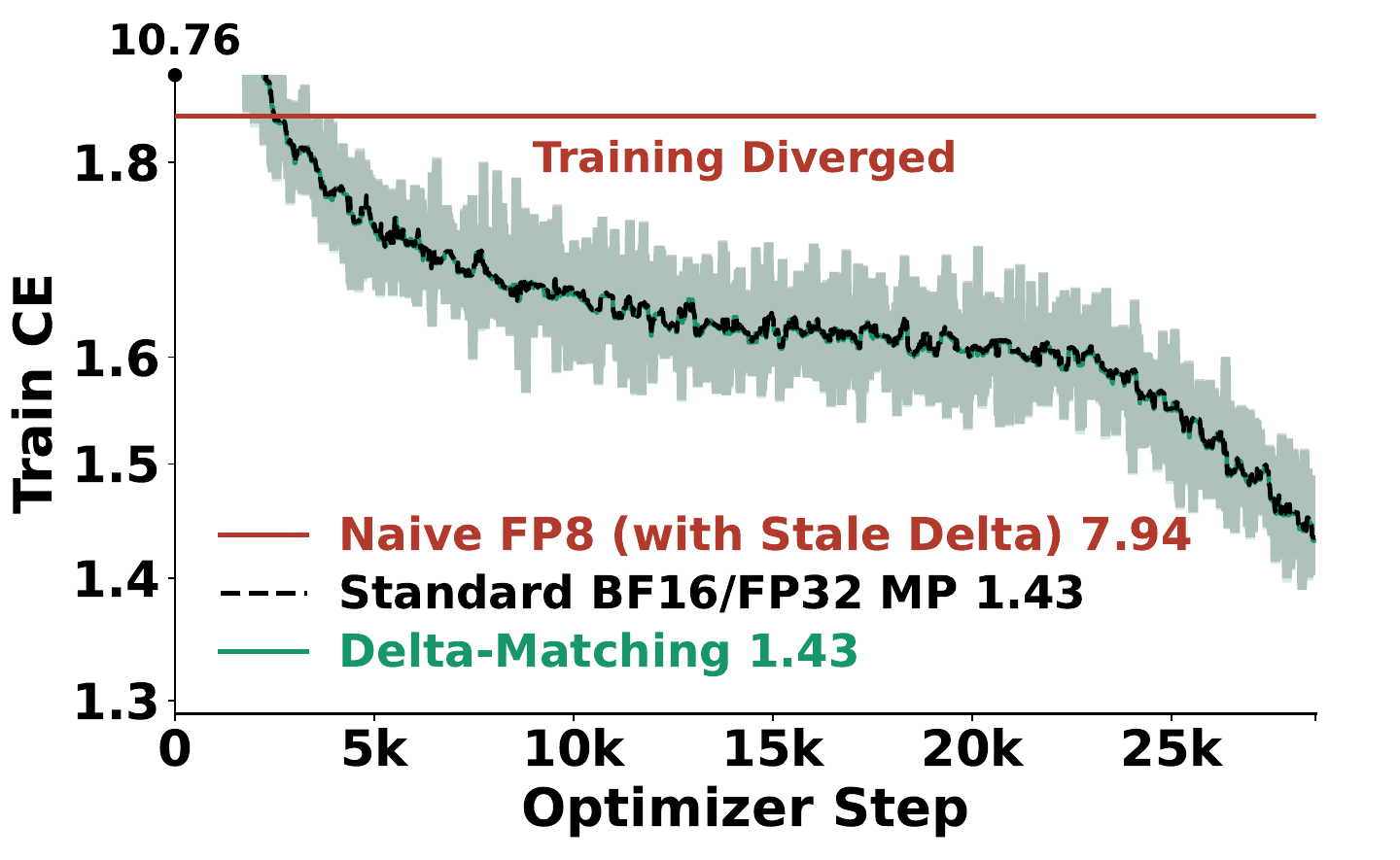}
  \caption{No QK-norm.}
  \label{fig:exp_attn_noqk}
\end{subfigure}\hfill
\begin{subfigure}[b]{0.33\textwidth}
  \centering
  \includegraphics[width=\linewidth]{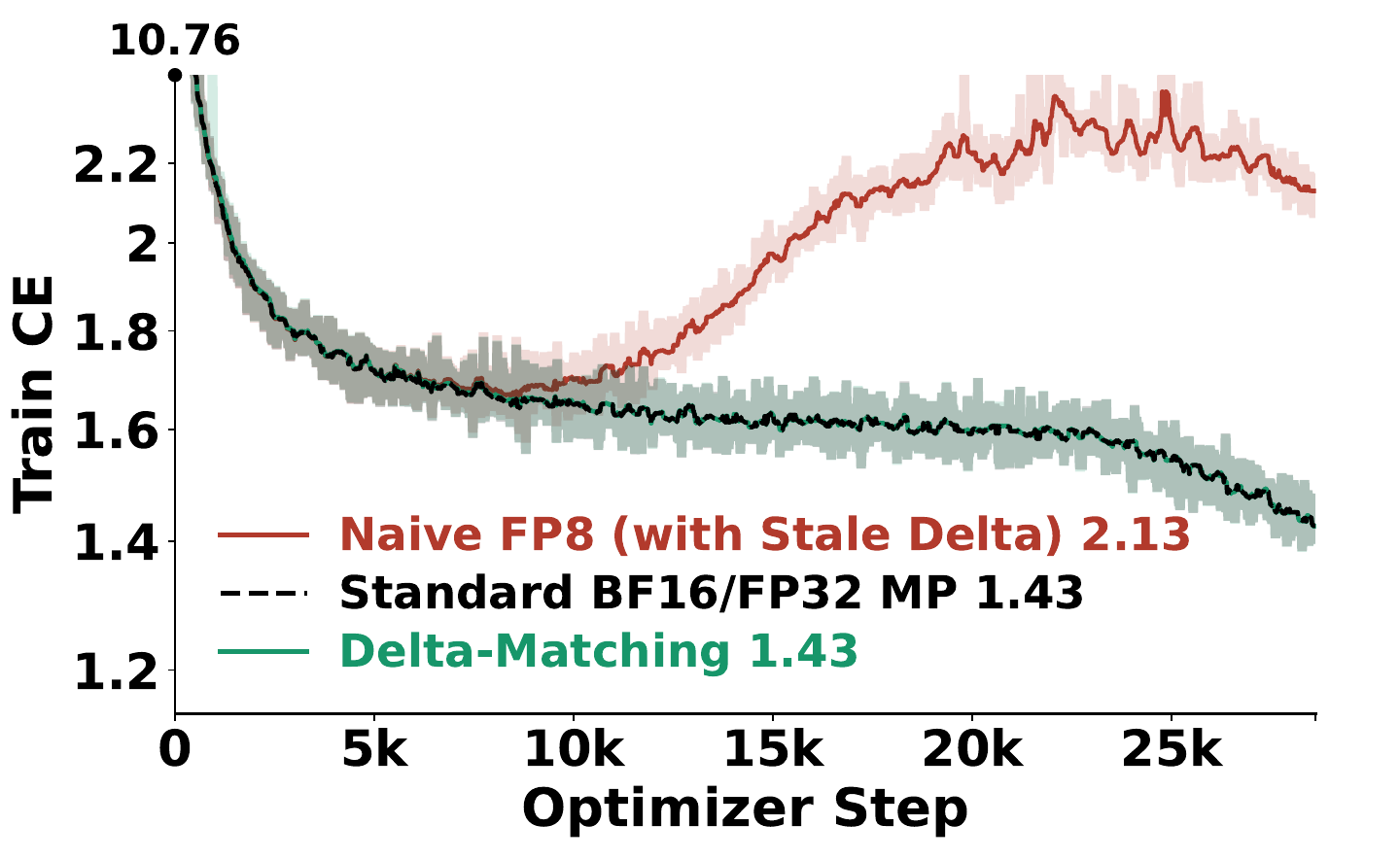}
  \caption{Head 256.}
  \label{fig:exp_attn_hd256}
\end{subfigure}
\caption{
Training cross-entropy for three GDN/GQA attention variants, each with its own BF16/FP32 reference.
The default configuration appears in Figure~\ref{fig:teaser}.
}
\label{fig:exp_attn_arch}
\end{figure}

Figures~\ref{fig:teaser} and~\ref{fig:exp_attn_arch} show training-loss trajectories
for the default configuration and three variants, respectively.

In the default configuration, consistent $dO$ reduces validation CE from $1.8970$
with stale delta to $1.4667$, still above BF16/FP32's $1.4178$;
cuDNN/TE FP8 reaches $1.6105$ (Table~\ref{tab:exp_cmp}).
Delta-Matching reaches $1.4162$ and the best FP8 aggregate downstream means.
Protecting $dP$ in BF16 also preserves near-reference quality, but retains one
higher-precision attention matmul and markedly different training
dynamics (Appendix~\ref{app:protect_fragile_operation}).

NoPE delays but does not eliminate stale-delta degradation; head dimension
256 also leaves a large loss gap (Figure~\ref{fig:exp_attn_arch}).
Their validation CE gaps from their BF16/FP32 references are
$0.2362$ and $0.6574$, respectively, while stale-delta training diverges
without QK normalization (Table~\ref{tab:exp_cmp}).
Delta-Matching stays within $0.0011$ validation CE of each variant's reference,
with broadly comparable downstream performance.

In the default configuration, Delta-Matching's individual commonsense benchmark
means stay within $1.0$ percentage point of BF16/FP32
(Appendix~\ref{app:commonsense_benchmarks}).
RULER-8K is lower at $47.8$ versus $52.3$; Section~\ref{sec:exp_stage} examines context extension.
At head dimension 256, Delta-Matching also matches BF16/FP32 with P4's FP8 GEMMs
(Appendix~\ref{app:fp8_linear}).
In a single-seed full-training comparison with Muon
(Appendix~\ref{app:muon_full_training}), stale delta degrades performance
while Delta-Matching matches the BF16/FP32 reference.

\subsection{How Invariant Violations Change Training Dynamics}
\label{sec:delta_training_dynamics}

We examine pretraining dynamics in the 1.67B GDN/GQA hybrid
(Appendix~\ref{app:delta_dynamics_setup}).

\begin{figure}[!htbp]
\centering
\begin{subfigure}[b]{0.32\textwidth}
  \centering
  \includegraphics[width=\linewidth]{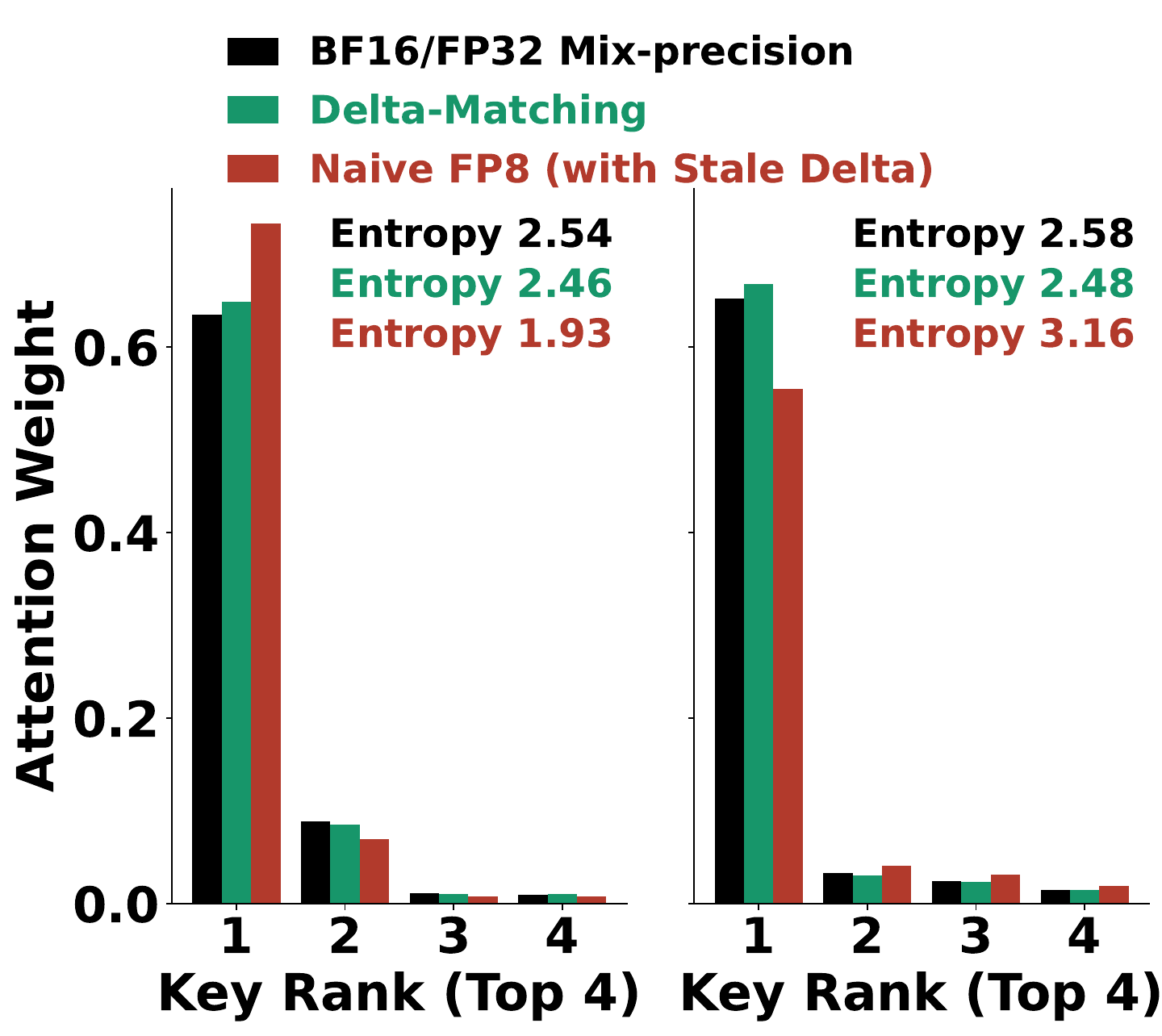}
  \caption{Stale delta sharpens or flattens attention rows.}
  \label{fig:delta_row_update}
\end{subfigure}\hfill
\begin{subfigure}[b]{0.33\textwidth}
  \centering
  \includegraphics[width=\linewidth]{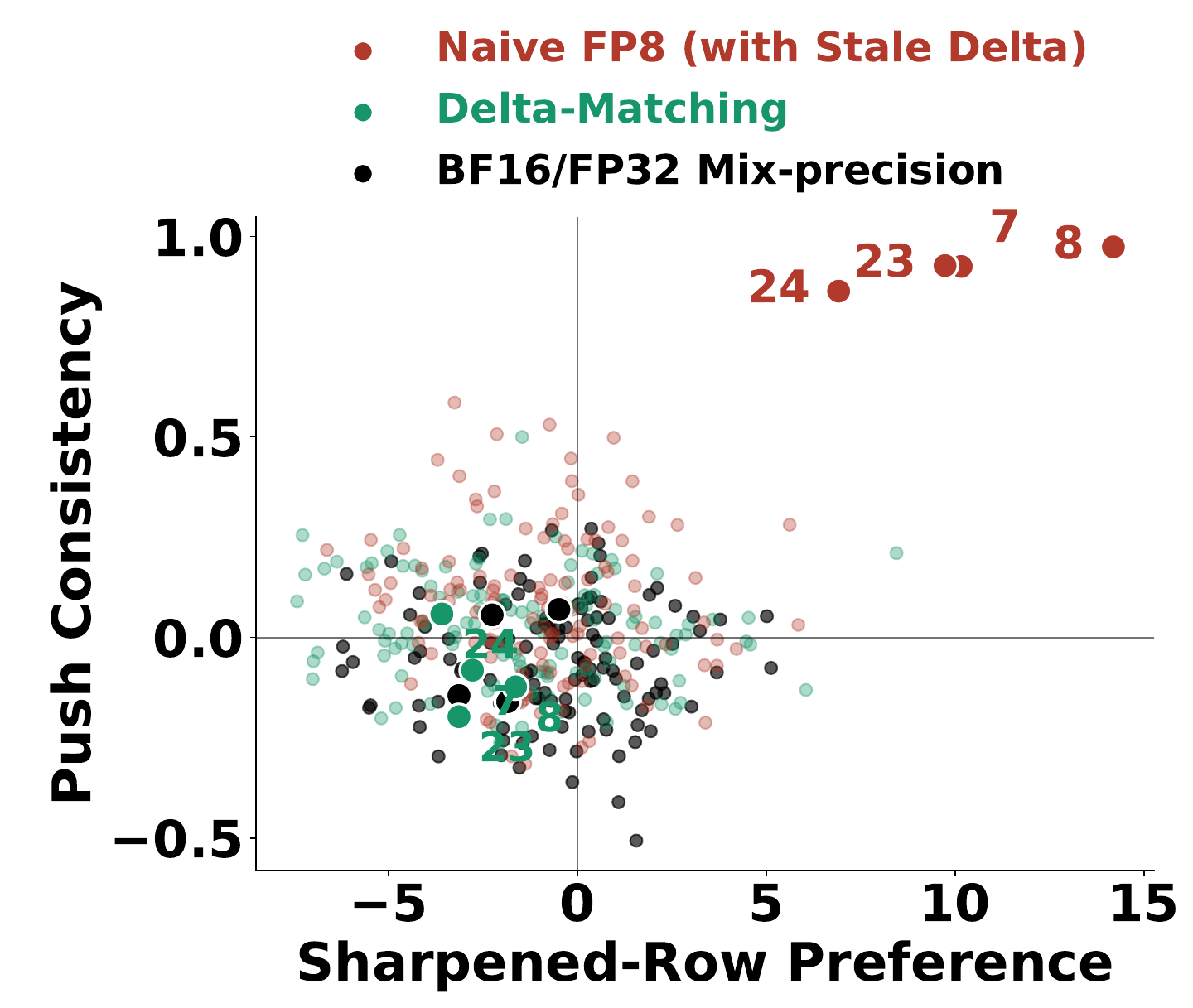}
  \caption{Favored channels show consistent gain increases.}
  \label{fig:delta_favored_pushed}
\end{subfigure}\hfill
\begin{subfigure}[b]{0.32\textwidth}
  \centering
  \includegraphics[width=\linewidth]{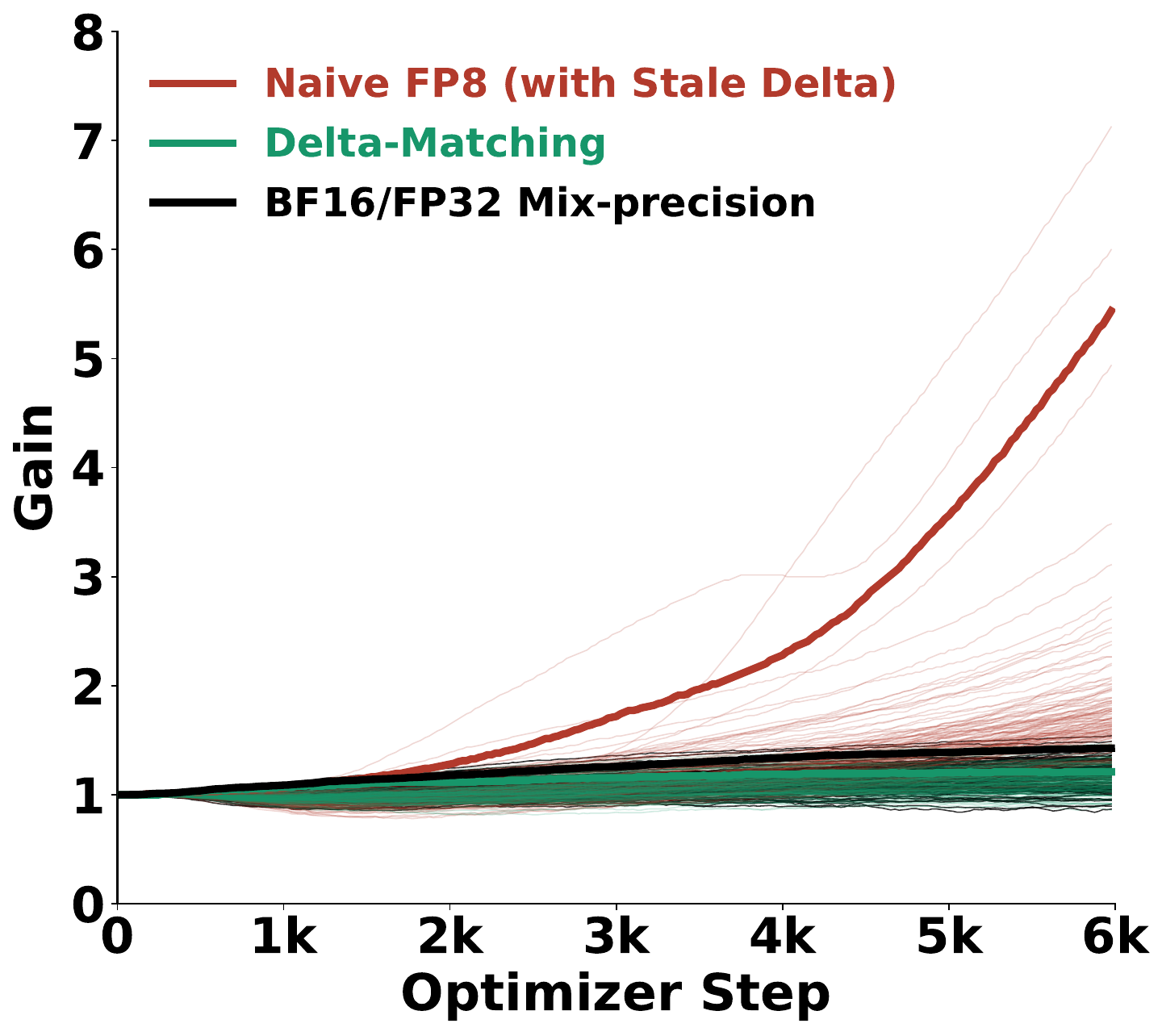}
  \caption{Key RMSNorm gain trajectories over optimizer steps.}
  \label{fig:delta_gain_growth}
\end{subfigure}
\caption{\textbf{Stale delta distorts attention updates and amplifies selected channels.}
Last-layer data from a 1.67B GDN/GQA hybrid; (a,b) at step 6,000.
\textbf{(a)} From shared FP8 logits, stale delta sharpens (left) or flattens (right) attention; Delta-Matching closely follows BF16/FP32 despite different forward logits.
\textbf{(b)} Only stale delta combines strong sharpened-row preference with consistent gain increases in highlighted channels.
\textbf{(c)} Stale delta selectively amplifies key RMSNorm gains, which stay controlled with Delta-Matching and BF16/FP32.
Thick lines track channel 7; thin lines track the others.
Appendix~\ref{app:figure3_quantities} defines quantities and row selection.
}
\label{fig:delta_dynamics}
\end{figure}

The stale-delta residual is the score-gradient row sum $\varepsilon_i=\sum_j dS_{ij}^{\mathrm{stale}}$.
For fixed $P$ and $\widehat{dP}$, stale delta adds $\varepsilon_iP_{ij}$ to the matched-delta gradient.
In isolation, this term biases a small logit update toward sharpening when $\varepsilon_i<0$ and flattening when $\varepsilon_i>0$ (Figure~\ref{fig:delta_row_update}).
Residuals can nearly cancel across rows without their effects canceling within channels.
Channels favored by sharpened rows also show consistent upward pressure on key RMSNorm gains (Figure~\ref{fig:delta_favored_pushed}).

Selective amplification compounds during training (Figure~\ref{fig:delta_gain_growth}).
Eventually, a few channels account for more than 99\% of $\sum_c(g_{q,c}g_{k,c})^2$ in affected layers, where $g_{q,c}$ and $g_{k,c}$ are channel $c$'s query and key RMSNorm gains.
This concentration can amplify selected logit contributions and promote softmax saturation, focusing attention on a few keys.

\subsection{Effects of Parameter Scaling on 8-bit Training}
\label{sec:exp_param_scale}

\begin{figure}[!htbp]
\centering
\captionsetup[subfigure]{justification=centering,font=scalepanel}
\begin{subfigure}[t]{0.20\textwidth}
  \vspace{0pt}
  \centering
  \includegraphics[width=\linewidth]{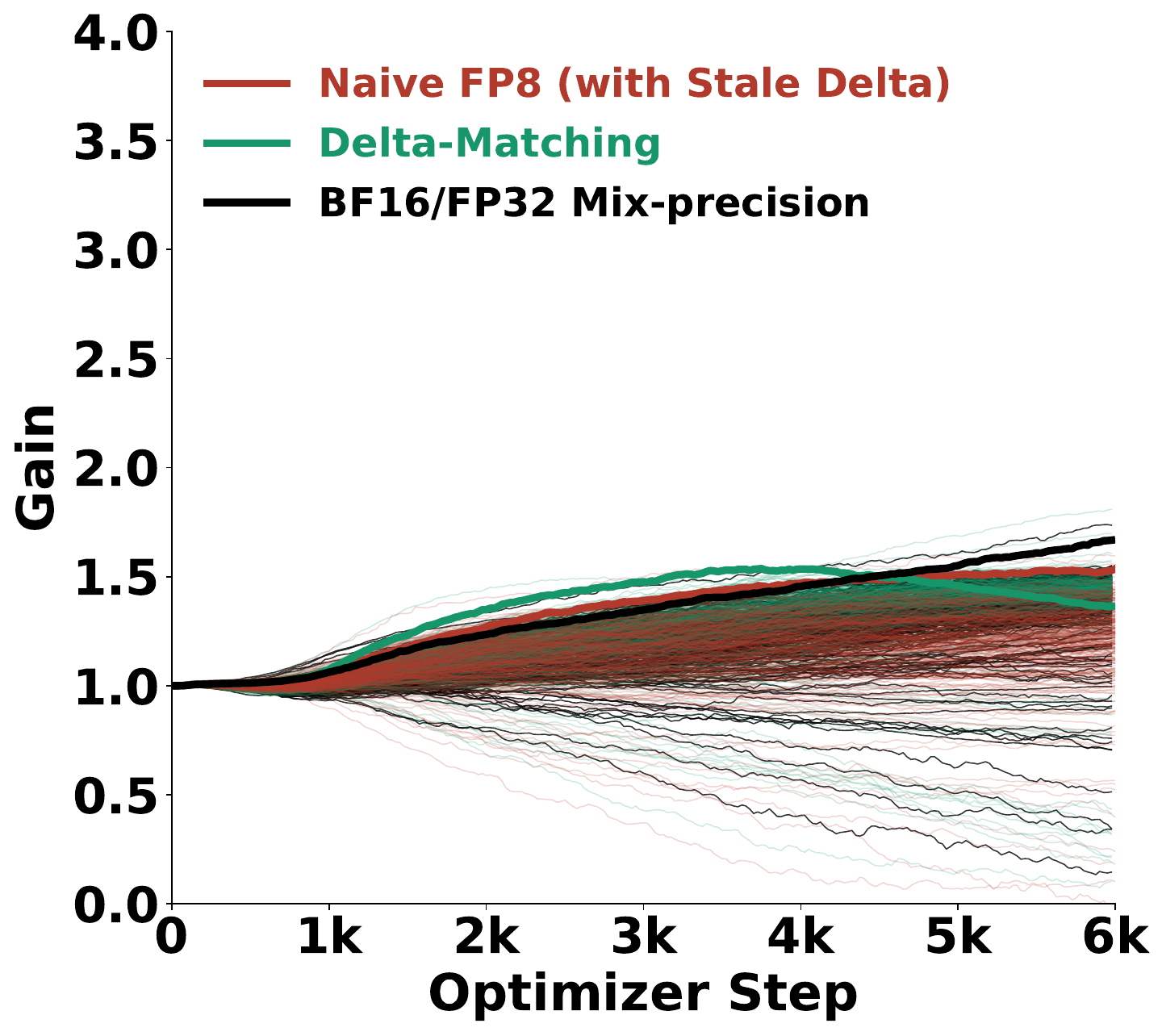}
  \caption{0.569B channel gain.}
  \label{fig:exp_param_scale_569M_gain}
\end{subfigure}\hspace{0.005\textwidth}%
\begin{subfigure}[t]{0.29\textwidth}
  \vspace{0pt}
  \centering
  \includegraphics[width=\linewidth]{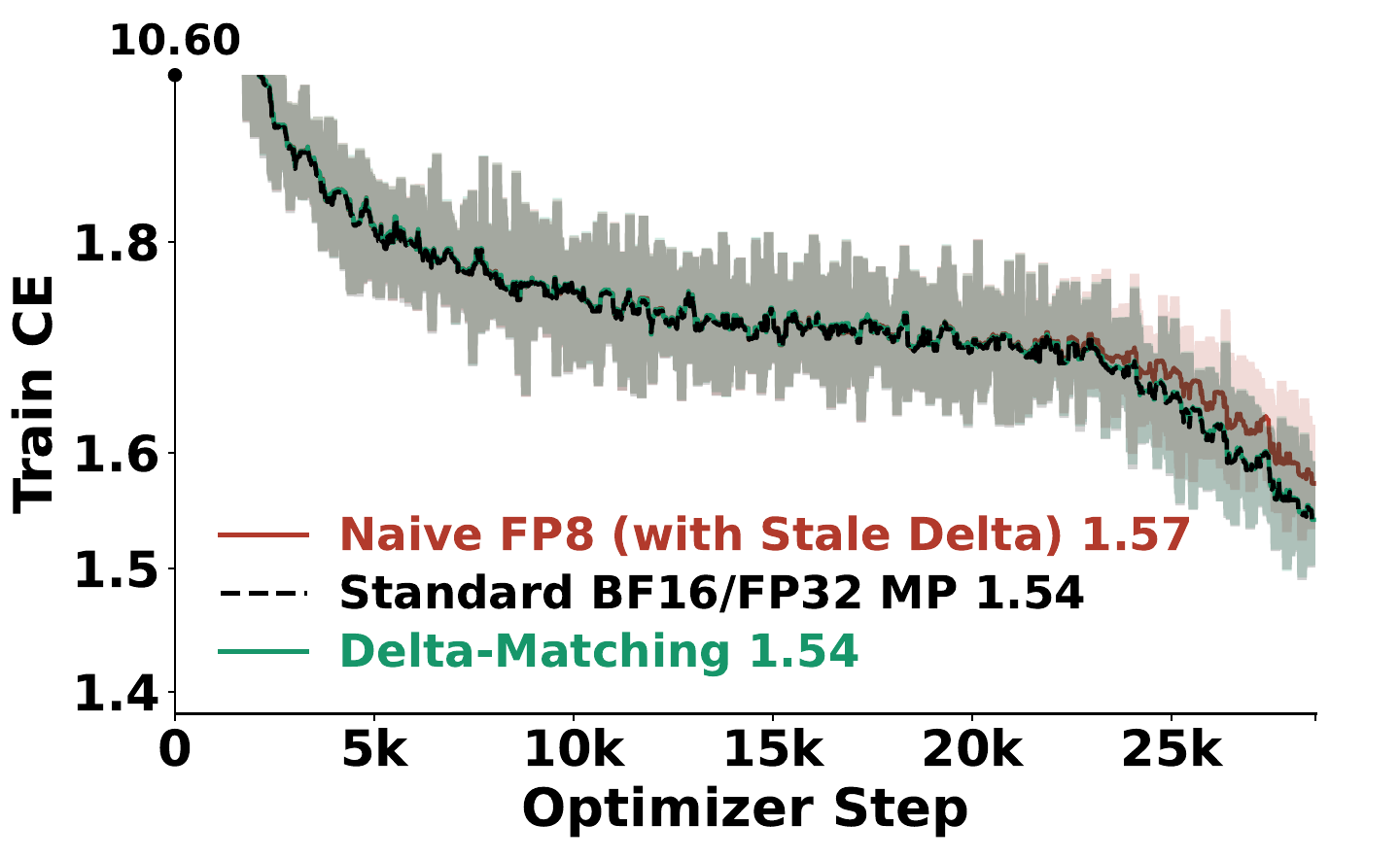}
  \caption{0.569B loss trend.}
  \label{fig:exp_param_scale_569M}
\end{subfigure}\hspace{0.01\textwidth}%
\begin{subfigure}[t]{0.20\textwidth}
  \vspace{0pt}
  \centering
  \includegraphics[width=\linewidth]{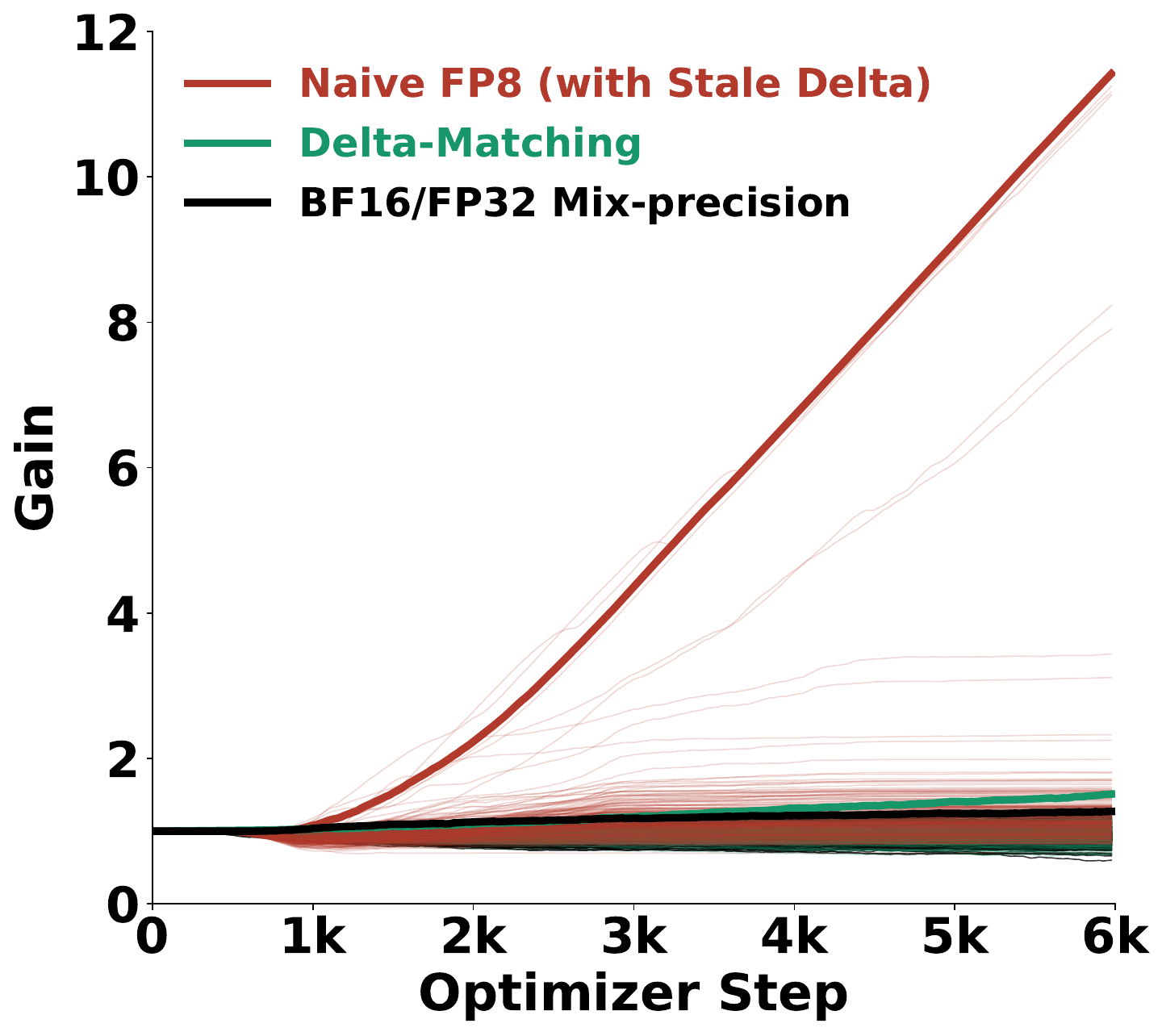}
  \caption{5.29B channel gain.}
  \label{fig:exp_param_scale_5p29B_gain}
\end{subfigure}\hspace{0.005\textwidth}%
\begin{subfigure}[t]{0.29\textwidth}
  \vspace{0pt}
  \centering
  \includegraphics[width=\linewidth]{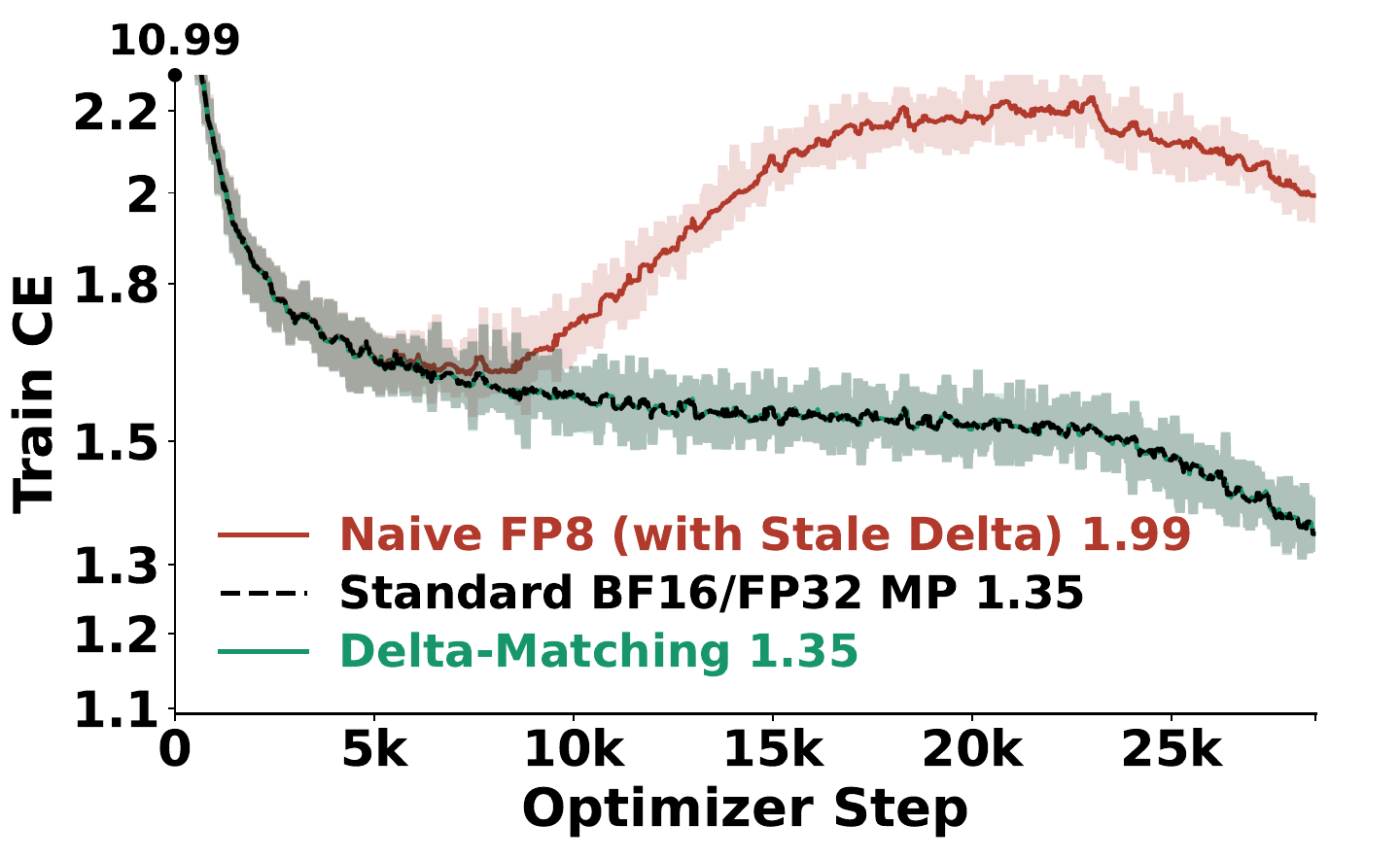}
  \caption{5.29B loss trend.}
  \label{fig:exp_param_scale_5p29B}
\end{subfigure}

\caption{
Channel gain during the first 6,000 optimizer steps (a,c) and training cross-entropy (b,d) for 0.569B (a,b) and 5.29B (c,d) GDN/GQA hybrids pretrained on 30B Nemotron-CC tokens~\citep{su2025nemotroncc,blakeman2025nemotron3nano}.
Delta-Matching tracks BF16/FP32 loss at both scales; naive FP8 has a larger gap at 5.29B.}

\label{fig:exp_param_scale}
\end{figure}

In our GDN/GQA experiments, severe stale-delta degradation emerges only beyond
the sub-billion regime examined in prior attention-training
studies~\citep{zhang2026sagebwd,qiu2026whybf16fa}.
We pretrain 569M and 5.29B GDN/GQA models with head dimension 256 using the
default 1.67B setup and base precision scopes (Section~\ref{sec:exp_cmp}).
At 569M, naive FP8 has a modest validation CE gap with largely preserved
downstream performance. At 5.29B, its CE reaches $1.9327$ versus $1.3349$ for
BF16/FP32. RULER-8K falls from $53.5\%$ to $16.3\%$ and CS9 from $63.1\%$ to $45.4\%$
(Figure~\ref{fig:exp_param_scale} and Table~\ref{tab:exp_param_scale}).

\begin{table}[!htbp]
\centering
\caption{
\textbf{Parameter scales.}
Mean $\pm$ SE for two seeds per model (0.569B, 5.289B); downstream scores in percent.
Individual commonsense benchmark accuracies are in Appendix~\ref{app:commonsense_benchmarks}.
}
\label{tab:exp_param_scale}
\scriptsize
\setlength{\tabcolsep}{1.8pt}
\begin{tabular*}{\linewidth}{@{\extracolsep{\fill}}llccccccc@{}}
\toprule
& & & \multicolumn{3}{c}{In-Context Recall $\uparrow$} & \multicolumn{3}{c}{General Capabilities $\uparrow$} \\
\cmidrule(lr){4-6}\cmidrule(lr){7-9}
Parameters & Method & Validation CE $\downarrow$ & RULER-4K & RULER-8K & Extraction & TriviaQA & MMLU & CS9 \\
\midrule
0.569B & BF16/FP32 MP & $1.5233\pmstd{0.0005}$ & $45.4\pmstd{0.5}$ & $37.8\pmstd{0.4}$ & $65.6\pmstd{0.6}$ & $10.2\pmstd{0.4}$ & $28.9\pmstd{0.5}$ & $54.7\pmstd{0.1}$ \\
& Naive FP8 (Stale Delta) & $1.5399\pmstd{0.0144}$ & $\mathbf{48.7}\pmstd{1.2}$ & $\mathbf{40.6}\pmstd{1.1}$ & $62.3\pmstd{0.9}$ & $\phantom{0}9.6\pmstd{0.6}$ & $28.5\pmstd{1.3}$ & $53.8\pmstd{0.6}$ \\
& Delta-Matching (\textit{ours}) & $\mathbf{1.5228}\pmstd{0.0015}$ & $46.3\pmstd{1.0}$ & $38.2\pmstd{0.8}$ & $\mathbf{62.9}\pmstd{0.2}$ & $\phantom{0}\mathbf{9.9}\pmstd{0.3}$ & $\mathbf{28.7}\pmstd{1.8}$ & $\mathbf{54.6}\pmstd{0.0}$ \\
\midrule
5.289B & BF16/FP32 MP & $1.3349\pmstd{0.0011}$ & $65.6\pmstd{0.2}$ & $53.5\pmstd{1.8}$ & $71.0\pmstd{0.9}$ & $21.7\pmstd{0.4}$ & $42.0\pmstd{0.5}$ & $63.1\pmstd{0.1}$ \\
& Naive FP8 (Stale Delta) & $1.9327\pmstd{0.0349}$ & $23.4\pmstd{2.4}$ & $16.3\pmstd{5.8}$ & $37.1\pmstd{2.3}$ & $\phantom{0}2.0\pmstd{0.1}$ & $26.3\pmstd{0.2}$ & $45.4\pmstd{0.3}$ \\
& Delta-Matching (\textit{ours}) & $\mathbf{1.3374}\pmstd{0.0005}$ & $\mathbf{69.2}\pmstd{2.0}$ & $\mathbf{58.6}\pmstd{0.1}$ & $\mathbf{70.3}\pmstd{0.6}$ & $\mathbf{21.6}\pmstd{0.1}$ & $\mathbf{41.3}\pmstd{0.2}$ & $\mathbf{62.0}\pmstd{0.4}$ \\
\bottomrule
\end{tabular*}
\end{table}

At 569M, stale delta produces little excess channel-gain growth
(Figure~\ref{fig:exp_param_scale_569M_gain}), whereas selected channels grow rapidly
at 5.29B (Figure~\ref{fig:exp_param_scale_5p29B_gain}).
The 5.29B gains rise more steeply than those of the 1.67B model
(Figure~\ref{fig:delta_gain_growth}) over the same first 6,000 optimizer steps.
These trajectories suggest faster selective channel amplification at larger tested
scales under stale delta; Delta-Matching suppresses excessive growth at all three scales.

Delta-Matching stays within $0.0025$ validation CE of BF16/FP32 at both scales
(Table~\ref{tab:exp_param_scale}).
At 5.29B, its RULER means exceed the reference, with slightly lower
extraction recall, MMLU, and CS9.

\subsection{Delta-Matching across Model Architectures}
\label{sec:exp_model_arch}

We compare BF16/FP32, stale delta, and Delta-Matching in three other 1.67B hybrids
trained on 30B tokens using the base precision scopes.
GDN/MLA replaces GQA with multi-head latent attention.
KDA/GQA and Mamba-2/GQA replace Gated DeltaNet~\citep{yang2025gdn} with
Kimi Delta Attention~\citep{team2025kimilinear} and
Mamba-2~\citep{dao2024mamba2}, respectively, and widen the FFN for parameter parity
(Appendix~\ref{app:train_architectures}).
Delta-Matching's correction applies only to the softmax-attention backward pass.

\begin{figure}[!htbp]
\centering
\begin{subfigure}[b]{0.32\textwidth}
  \centering
  \includegraphics[width=\linewidth]{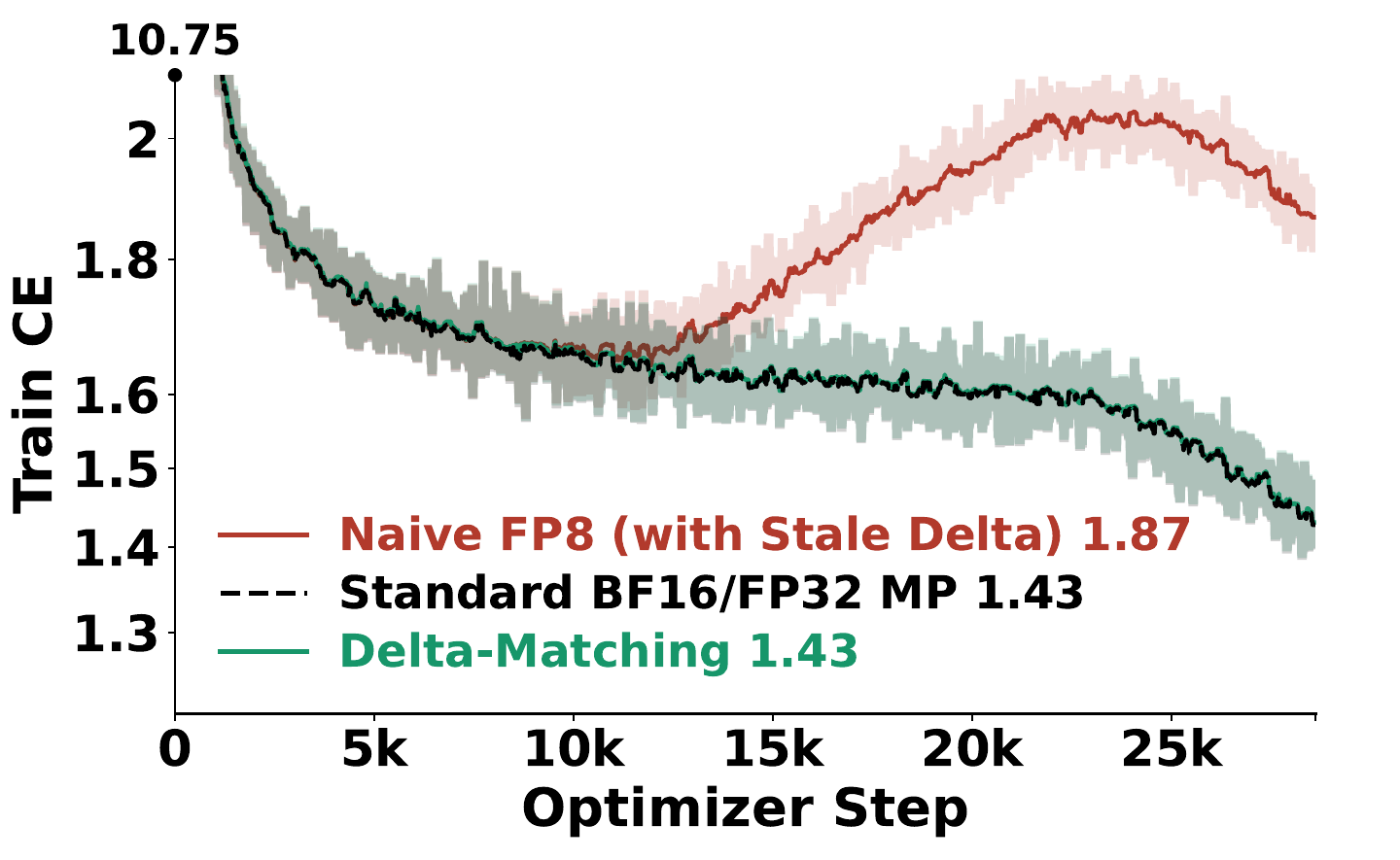}
  \caption{1.67B GDN/MLA hybrid.}
  \label{fig:exp_model_gdn_mla}
\end{subfigure}\hspace{0.015\textwidth}%
\begin{subfigure}[b]{0.32\textwidth}
  \centering
  \includegraphics[width=\linewidth]{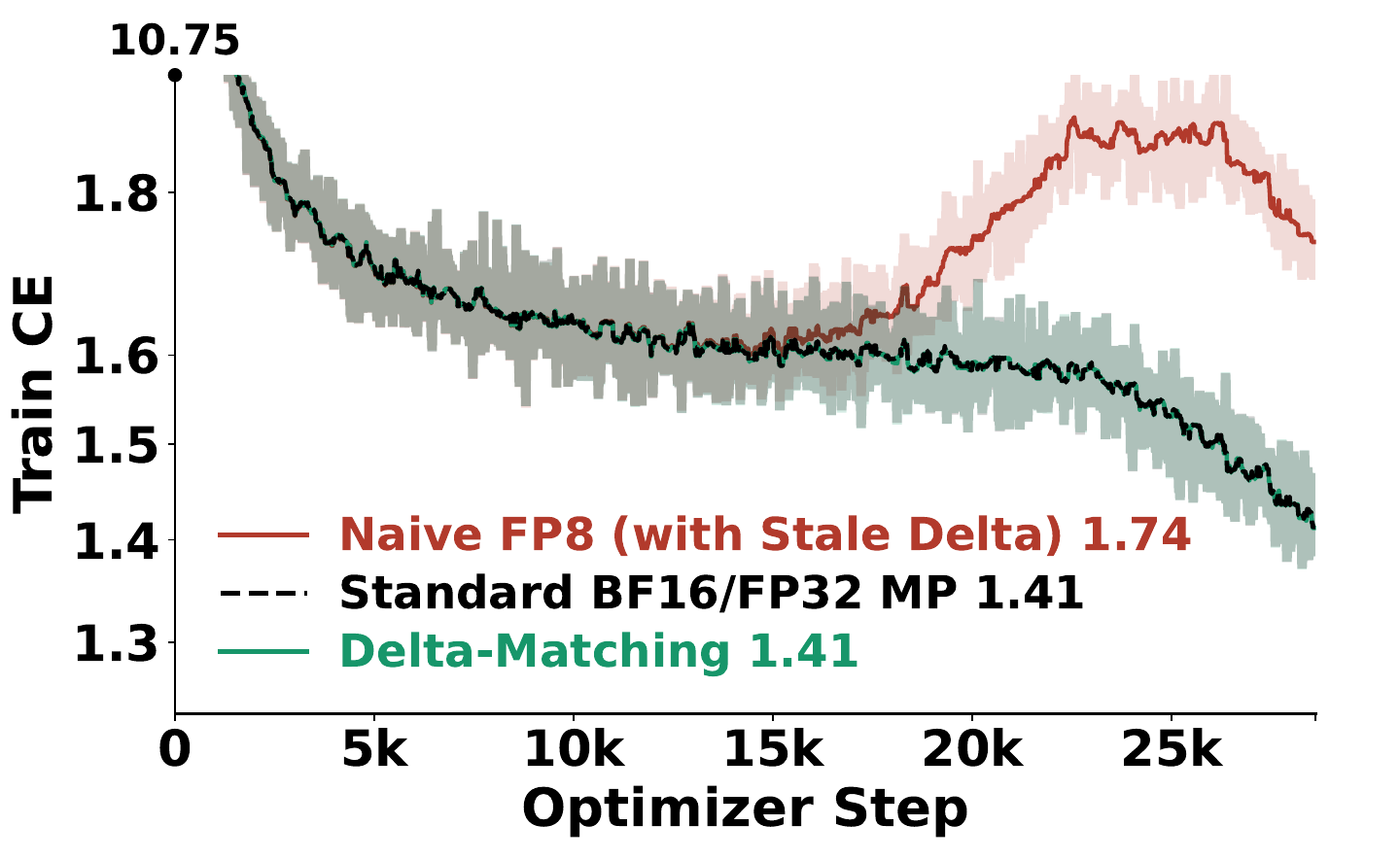}
  \caption{1.67B KDA/GQA hybrid.}
  \label{fig:exp_model_kda_gqa}
\end{subfigure}\hspace{0.015\textwidth}%
\begin{subfigure}[b]{0.32\textwidth}
  \centering
  \includegraphics[width=\linewidth]{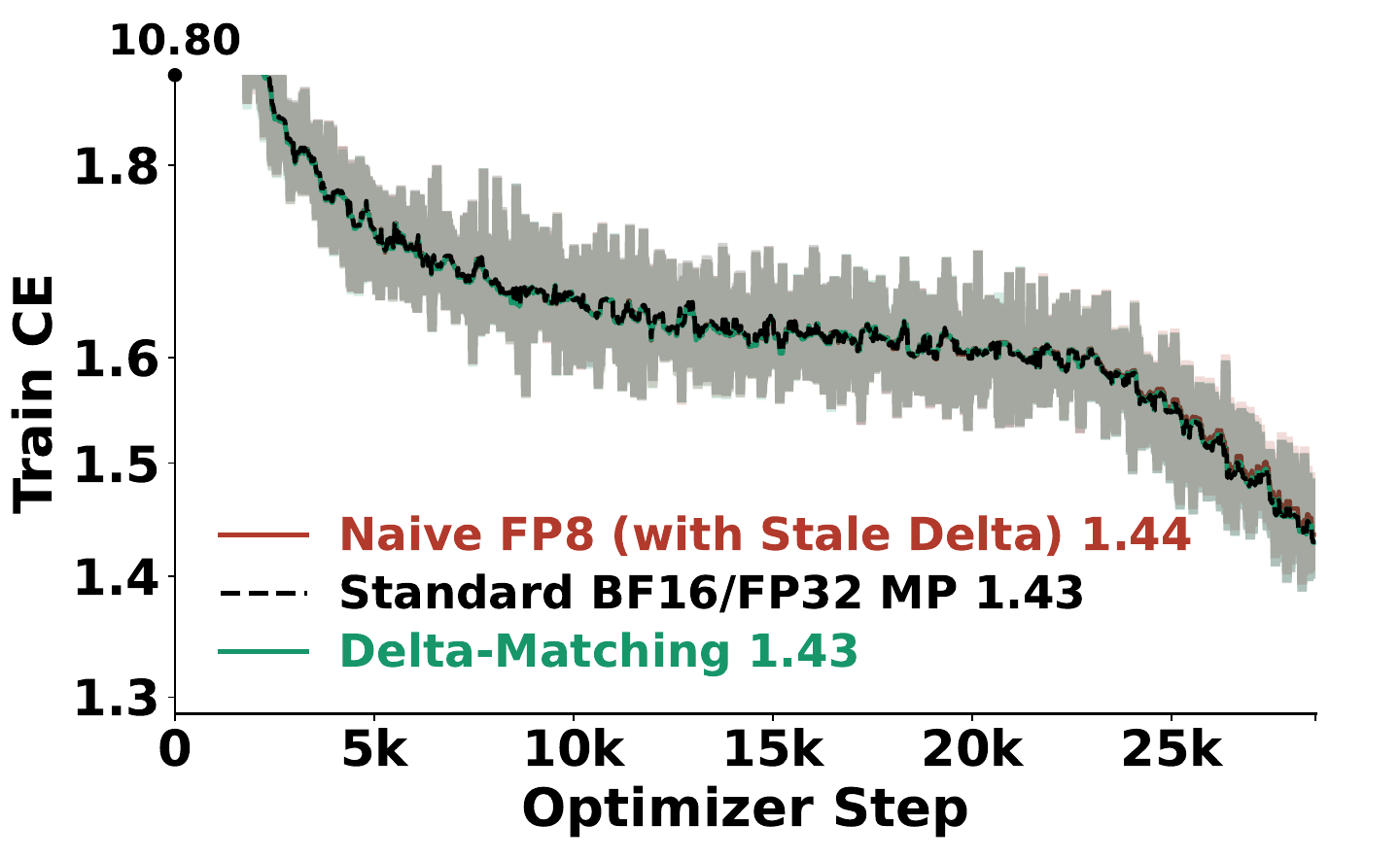}
  \caption{1.67B Mamba-2/GQA hybrid.}
  \label{fig:exp_model_mamba_gqa}
  \label{fig:app_mamba_gqa_loss}
\end{subfigure}
\caption{
Training cross-entropy for 1.67B GDN/MLA, KDA/GQA, and Mamba-2/GQA hybrids over 30B tokens.
Stale delta develops substantial loss gaps in the first two but remains near BF16/FP32
in Mamba-2/GQA at this scale. Delta-Matching tracks all three references.
}
\label{fig:exp_model_arch}
\end{figure}

\begin{table}[!htbp]
\centering
\caption{
\textbf{Model architectures.}
Mean $\pm$ SE over two seeds; downstream scores are percentages.
Individual commonsense benchmark accuracies for GDN/MLA and KDA/GQA appear in
Appendix~\ref{app:commonsense_benchmarks}.
}
\label{tab:exp_model_arch}
\label{tab:app_mamba_gqa_eval}
\scriptsize
\setlength{\tabcolsep}{1.8pt}
\begin{tabular*}{\linewidth}{@{\extracolsep{\fill}}llccccccc@{}}
\toprule
& & & \multicolumn{3}{c}{In-Context Recall $\uparrow$} & \multicolumn{3}{c}{General Capabilities $\uparrow$} \\
\cmidrule(lr){4-6}\cmidrule(lr){7-9}
Architecture & Method & Validation CE $\downarrow$ & RULER-4K & RULER-8K & Extraction & TriviaQA & MMLU & CS9 \\
\midrule
GDN/MLA & BF16/FP32 MP & $1.4178\pmstd{0.0032}$ & $54.9\pmstd{2.2}$ & $47.4\pmstd{0.1}$ & $66.7\pmstd{0.1}$ & $15.7\pmstd{0.3}$ & $36.3\pmstd{1.6}$ & $59.2\pmstd{0.0}$ \\
& Naive FP8 (Stale Delta) & $1.9268\pmstd{0.0844}$ & $25.9\pmstd{2.0}$ & $18.9\pmstd{5.3}$ & $42.9\pmstd{4.4}$ & $\phantom{0}2.0\pmstd{0.7}$ & $26.1\pmstd{0.5}$ & $46.0\pmstd{1.3}$ \\
& Delta-Matching (\textit{ours}) & $\mathbf{1.4181}\pmstd{0.0024}$ & $\mathbf{60.4}\pmstd{0.6}$ & $\mathbf{50.0}\pmstd{1.7}$ & $\mathbf{67.5}\pmstd{0.1}$ & $\mathbf{16.3}\pmstd{0.3}$ & $\mathbf{34.4}\pmstd{0.5}$ & $\mathbf{59.2}\pmstd{0.3}$ \\
\midrule
KDA/GQA & BF16/FP32 MP & $1.3994\pmstd{0.0004}$ & $65.6\pmstd{0.5}$ & $50.0\pmstd{2.3}$ & $67.9\pmstd{0.4}$ & $17.7\pmstd{0.1}$ & $34.8\pmstd{0.1}$ & $60.2\pmstd{0.3}$ \\
& Naive FP8 (Stale Delta) & $1.8224\pmstd{0.1100}$ & $26.2\pmstd{10.9}$ & $22.9\pmstd{9.2}$ & $44.2\pmstd{11.2}$ & $\phantom{0}3.4\pmstd{1.0}$ & $26.2\pmstd{0.3}$ & $48.5\pmstd{2.2}$ \\
& Delta-Matching (\textit{ours}) & $\mathbf{1.3988}\pmstd{0.0001}$ & $\mathbf{60.7}\pmstd{2.4}$ & $\mathbf{50.7}\pmstd{2.2}$ & $\mathbf{66.8}\pmstd{1.3}$ & $\mathbf{18.3}\pmstd{0.2}$ & $\mathbf{35.7}\pmstd{1.9}$ & $\mathbf{60.0}\pmstd{0.1}$ \\
\midrule
Mamba-2/GQA & BF16/FP32 MP & $1.4152\pmstd{0.0008}$ & $56.0\pmstd{1.6}$ & $46.5\pmstd{1.2}$ & $67.9\pmstd{0.8}$ & $15.5\pmstd{0.5}$ & $30.5\pmstd{0.0}$ & $58.6\pmstd{0.5}$ \\
& Naive FP8 (Stale Delta) & $1.4191\pmstd{0.0024}$ & $\mathbf{61.8}\pmstd{0.1}$ & $\mathbf{49.0}\pmstd{1.3}$ & $68.4\pmstd{0.6}$ & $15.5\pmstd{0.2}$ & $\mathbf{31.7}\pmstd{2.0}$ & $58.2\pmstd{0.4}$ \\
& Delta-Matching (\textit{ours}) & $\mathbf{1.4152}\pmstd{0.0004}$ & $58.0\pmstd{1.4}$ & $48.0\pmstd{0.6}$ & $\mathbf{68.8}\pmstd{0.6}$ & $\mathbf{15.8}\pmstd{0.1}$ & $31.2\pmstd{1.1}$ & $\mathbf{58.4}\pmstd{0.6}$ \\
\bottomrule
\end{tabular*}
\end{table}

Figure~\ref{fig:exp_model_arch} shows the corresponding training-loss trajectories.
Stale delta raises validation CE by $0.5090$ for GDN/MLA and $0.4230$ for KDA/GQA relative to
BF16/FP32 (Table~\ref{tab:exp_model_arch}). In these two hybrids, Delta-Matching
stays within $0.0006$ of each reference and improves every reported downstream mean over stale delta, with broadly
comparable performance to BF16/FP32, although KDA/GQA's RULER-4K mean is lower
at $60.7$ versus $65.6$.

In contrast, Mamba-2/GQA shows no substantial stale-delta performance gap at
1.67B parameters and 30B tokens. Its validation CE is $1.4191$ with stale delta
versus $1.4152$ for both BF16/FP32 and Delta-Matching, with broadly comparable
downstream performance (Table~\ref{tab:exp_model_arch}).
Layer probes nevertheless reveal severe gain corruption under stale delta,
with masking results suggesting compensation elsewhere
(Appendix~\ref{app:mamba_layer_probe}).
Similar behavior in some pure-GQA configurations reinforces the need to check
backward consistency and internal dynamics alongside loss and downstream performance
(Appendix~\ref{app:convergent_architectures}).

\FloatBarrier
\subsection{Delta-Matching During Context Extension}

\label{sec:exp_stage}

We study continued pretraining at longer context and a lower learning rate,
extending the 1.67B GDN/GQA hybrid from 8K to 64K on ProLong-derived
data~\citep{gao2025prolong} (Appendix~\ref{app:data_recipe}).
Methods share each initialization's completed BF16/FP32 checkpoint and train
28,000 steps with fresh optimizers at a peak learning rate of $8\times10^{-4}$.

\begin{figure}[!htb]
\centering
\includegraphics[width=\linewidth]{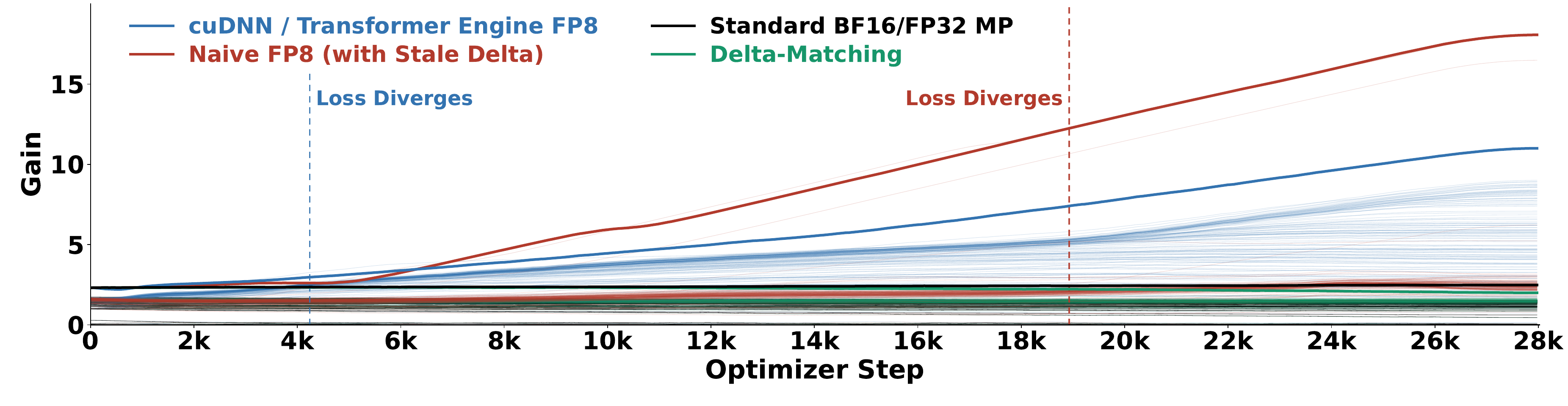}
\par\smallskip
\includegraphics[width=\linewidth]{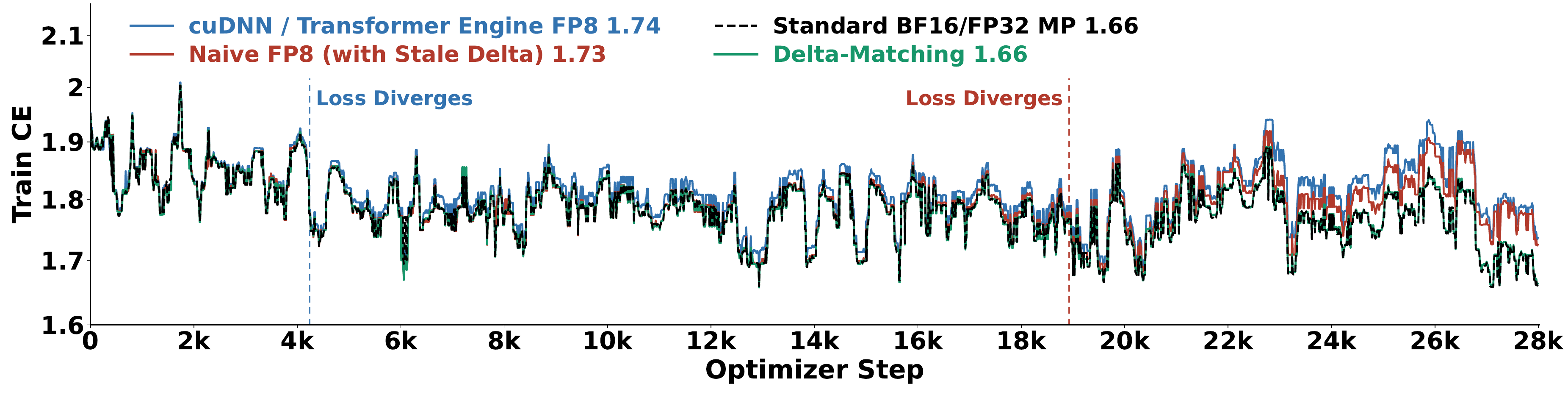}
\caption{
Channel gains (top) and training cross-entropy (bottom) during 8K-to-64K
continued pretraining of a 1.67B GDN/GQA hybrid at a lower learning rate.
Stale-delta FP8 and cuDNN/TE FP8 accumulate excess channel gain before visible
loss gaps emerge, while Delta-Matching tracks BF16/FP32.
Shared optimizer-step axes align the dashed loss-gap onset markers.
The cuDNN/TE FP8 curve is the first-seed rerun.
}
\label{fig:exp_training_stages}
\end{figure}

Both naive FP8 with stale delta and cuDNN/TE FP8 accumulate excess channel gains
before their loss curves visibly separate from BF16/FP32
(Figure~\ref{fig:exp_training_stages}).
For stale delta, the loss gap exceeds $0.01$ only after roughly 19{,}000 steps
(20B tokens). This suggests that, at fixed tokens per step, limited training-token
budgets can conceal FP8 attention instability that emerges after more optimizer updates.
Delta-Matching maintains near-reference channel gains and tracks BF16/FP32 training loss.
The step-18,000 corpus expansion preserved optimizer state but reset TE's amplitude
history (Appendix~\ref{app:train_recipes}), so the observed timing does not isolate
duration, corpus-size, or learning-rate effects.

\begin{table}[!htbp]
\centering
\caption{
\textbf{In-context recall after 8K-to-64K context extension.}
Mean $\pm$ SE over two seeds. For cuDNN/TE FP8, the first seed is a rerun on the
same batches because its original run diverged.
Scores are percentages.
}
\label{tab:exp_training_stages}
\scriptsize
\setlength{\tabcolsep}{1.8pt}
\begin{tabular*}{\linewidth}{@{\extracolsep{\fill}}lcccccc@{}}
\toprule
Method & RULER-4K $\uparrow$ & RULER-8K $\uparrow$ & RULER-16K $\uparrow$ & RULER-32K $\uparrow$ & RULER-64K $\uparrow$ & Extraction $\uparrow$ \\
\midrule
BF16/FP32 MP & $66.3\pmstd{0.9}$ & $59.2\pmstd{3.3}$ & $55.1\pmstd{3.7}$ & $50.0\pmstd{5.1}$ & $46.0\pmstd{4.5}$ & $69.0\pmstd{1.6}$ \\
\midrule
cuDNN/TE FP8 & $58.4\pmstd{8.7}$ & $51.6\pmstd{5.7}$ & $49.1\pmstd{5.4}$ & $44.7\pmstd{3.4}$ & $38.5\pmstd{1.8}$ & $65.7\pmstd{3.0}$ \\
Naive FP8 (Stale Delta) & $62.8\pmstd{1.7}$ & $56.2\pmstd{1.5}$ & $50.0\pmstd{2.2}$ & $45.8\pmstd{0.8}$ & $40.1\pmstd{0.6}$ & $64.5\pmstd{0.4}$ \\
Delta-Matching (\textit{ours}) & $\mathbf{67.6}\pmstd{4.4}$ & $\mathbf{62.2}\pmstd{2.8}$ & $\mathbf{55.4}\pmstd{1.1}$ & $\mathbf{51.6}\pmstd{0.1}$ & $\mathbf{45.6}\pmstd{0.6}$ & $\mathbf{67.9}\pmstd{0.6}$ \\
\bottomrule
\end{tabular*}
\end{table}

Delta-Matching tracks BF16/FP32 on RULER ($45.6\%$ versus $46.0\%$ at 64K)
and has higher mean scores than both FP8 baselines on all recall metrics
(Table~\ref{tab:exp_training_stages}).
The original first-seed cuDNN/TE run diverged, but its rerun completed without divergence.
Both BF16/FP32 and Delta-Matching lose MMLU and TriviaQA performance after extension
(Appendix~\ref{app:stage_capabilities}).

\FloatBarrier
\section{Conclusion}
\label{sec:conclusion}

We derive stale delta's violation of FP8 attention's zero-row-sum invariant and trace its training consequences empirically.
Failures at larger scales and over longer runs are consistent with accumulated optimization error.
QK normalization, NoPE, larger head dimensions, and lower-learning-rate context extension do not eliminate them.
Other mitigations leave performance gaps or altered dynamics; some architectures conceal gain corruption behind near-reference performance.

Delta-Matching restores the invariant under the stated assumptions, enabling native FP8 in all seven attention-core matmuls without auxiliary forward outputs.
Across tested attention variants, architectures, scales, and stages, it matches BF16/FP32 training loss and overall downstream performance.
Appendix~\ref{app:limitations} discusses limitations and future directions.
These findings highlight forward-backward numerical consistency as essential for reliable 8-bit training and provide a solid, practical solution toward fully 8-bit LLM training.

\subsection*{AI use statement}
We used generative AI tools to assist with proof writing, method implementation,
translation, and dataset cleaning and reformatting.
We also used these tools to identify relevant literature,
monitor experiments, and improve the clarity and readability of the manuscript.
We have not used generative AI tools for any other tasks requiring disclosure.
We have reviewed all AI-assisted work. The authors independently checked the
correctness of the proofs, manually designed test cases to validate the
method implementations, and manually inspected selected samples to check
the data processing.
We take responsibility for the final content of this work, including text,
claims, or artifacts produced with the aid of generative AI.

\subsection*{Ethics statement}
We are not aware of any ethical issues associated with this work.

\subsection*{Reproducibility statement}
We state the numerical assumptions and derive the method in
Section~\ref{sec:the_delta}, with a detailed derivation of the quantization
residual in Appendix~\ref{app:proofs}.
Appendix~\ref{app:data_recipe} documents data preparation, and
Appendix~\ref{app:train_config} specifies model architectures, training
precision, optimization recipes, and experiment configurations.
Evaluation settings and benchmark protocols are detailed in
Appendices~\ref{app:eval_configuration} and~\ref{app:eval}, with uncertainty
reporting defined in Appendix~\ref{app:conventions_uncertainty}.
We will release our implementation, trained models, and data recipes to
support reproducibility.

\bibliography{references}
\bibliographystyle{iclr2027_conference}

\appendix
\clearpage
\section*{Appendix Contents}
\pdfbookmark[0]{Appendix Contents}{appendix.contents}

\makeatletter
\begingroup
\setlength{\parskip}{0pt}
\c@tocdepth=3\relax
\hypersetup{linktoc=all}
\@starttoc{apc}
\endgroup
\makeatother
\clearpage

\let\appendixoriginaladdcontentsline\addcontentsline
\renewcommand{\addcontentsline}[3]{%
  \appendixoriginaladdcontentsline{#1}{#2}{#3}%
  \begingroup
  \def\appendixcontentsfile{#1}%
  \def\appendixstandardtoc{toc}%
  \ifx\appendixcontentsfile\appendixstandardtoc
    \appendixoriginaladdcontentsline{apc}{#2}{#3}%
  \fi
  \endgroup
}

\section{Extended Discussion}

We provide additional background on low-precision hardware, training systems,
and attention architectures. We then derive the stale-delta quantization
residual and its implications for key gradients.

\subsection{Low-Precision Hardware and Training Systems}
\label{app:low_precision_train}

\paragraph{Formats and scaling.}
E4M3 and E5M2 are 8-bit floating-point formats~\citep{micikevicius2022fp8} standardized in OCP OFP8~\citep{micikevicius2023ocp}.
The OCP Microscaling (MX) specification adds block-scaled formats such as MXFP8, sharing an 8-bit power-of-two E8M0 scale across each block of 32 values~\citep{rouhani2023ocpmx,rouhani2023microscaling}.
Per-tensor FP8 training typically uses E4M3 for forward tensors and the wider-range E5M2 for gradients; finer-grained MXFP8 scaling makes E4M3 practical for both passes, including activation gradients~\citep{mishra2025recipes}.

\paragraph{Hardware capabilities and realized speedups.}
Tensor Core throughput figures below are peak dense rates.
Hopper introduced FP8 Tensor Cores, with H100 SXM peaking at 1,979 TFLOPS in FP8 versus 989 TFLOPS in BF16 and Transformer Engine managing per-tensor scaling~\citep{nvidia2023hopper,nvidia_te_fp8primer}.
B200 peaks at 4.5 PFLOPS in FP8 versus 2.2 PFLOPS in BF16, while Blackwell adds native Tensor Core block scaling for MXFP8 and NVFP4~\citep{nvidia2025blackwell,nvidia_ptx_blockscaling}.
NVFP4 offers $2\times$ the FP8 throughput on Blackwell and $3\times$ on Blackwell Ultra~\citep{nvidia2025blackwellultra}.
Preliminary Rubin specifications list training throughput of up to 17.5, 35, and 4 PFLOPS for FP8, NVFP4, and BF16, respectively, with 19.2~TB/s of HBM4 bandwidth per GPU; partner availability was announced for the second half of 2026~\citep{nvidia2026rubinnvl72,nvidia2026rubinannouncement}.
Realized gains are smaller. Transformer Engine reports about a $1.7\times$ FP8 GEMM speedup over BF16 on Hopper, with remaining BF16 operators, especially attention, limiting end-to-end speedups~\citep{nvidia_te_speedups}.
FP32 remains standard for softmax because of training stability~\citep{ding2021cogview,zeng2022glm,qiu2026whybf16fa}; H100 exponential throughput is largely precision-independent at about 3.9 TFLOPS, so FP8 primarily benefits the attention matmuls~\citep{shah2024flashattention3}.

\paragraph{Training-system scope and precision exceptions.}
FP8-LM stores gradients, all-reduce traffic, and Adam's first moment in FP8, while COAT quantizes activations and both optimizer moments~\citep{peng2023fp8,xi2025coat}.
DeepSeek-V3 executes forward, activation-gradient, and weight-gradient GEMMs of its quantized linear layers in FP8 but retains the original precision for embeddings, the output head, gating, normalization, and attention operators~\citep{liu2024deepseek}.
NVIDIA reports BF16 parity using simulated MXFP8 linear-layer GEMMs in an 8B model trained on 15T tokens~\citep{mishra2025recipes}, and NVFP4 pretraining of a 12B model on 10T tokens with 16\% of its linear layers retained in BF16~\citep{abecassis2025pretraining}.
Nemotron~3, a hybrid Mamba-MoE architecture, uses native NVFP4 pretraining for up to 25T tokens, but retains QKV and attention output projections in BF16 and reports worse loss when jointly quantizing them with Mamba output projections~\citep{nvidia2025nemotron3}.
At the trillion-parameter scale, Ling-1T applies blockwise FP8 to most linear layers, reporting at most a 0.25\% relative loss difference from BF16 after 900B tokens~\citep{team2025every}.
Kimi K3 uses MXFP4 for routed-expert weights during post-training~\citep{team2026kimi}.
These systems retain higher precision for the attention core by excluding it or restricting low-precision computation to linear or expert GEMMs~\citep{peng2023fp8,xi2025coat,liu2024deepseek,mishra2025recipes,abecassis2025pretraining,team2025every,nvidia2025nemotron3,team2026kimi}.

\subsection{Attention and Hybrid Architectures}
\label{app:attention_architectures}

\paragraph{Softmax attention and its backward pass.}
\label{app:softmax_attention}
Given queries $Q$, keys $K$, values $V$, and head dimension $d$, softmax attention and its backward pass compute~\citep{vaswani2017attention}
\[
\begin{gathered}
S=QK^\top/\sqrt{d},\qquad P=\mathrm{softmax}(S),\qquad O=PV,\\
dV=P^\top dO,\quad dP=dO\,V^\top,\quad dS=P\odot(dP-\delta),\quad dQ=dS\,K/\sqrt{d},\quad dK=dS^\top Q/\sqrt{d}.
\end{gathered}
\]
Here $dX$ is the loss gradient for $X$, $\odot$ is elementwise multiplication, and $\delta_i=\sum_jP_{ij}dP_{ij}$ is broadcast across row $i$.

\paragraph{Softmax attention variants.}
Grouped-query attention (GQA) shares each key-value head across a group of query heads, interpolating between multi-head and multi-query attention~\citep{shazeer2019fast,ainslie2023gqa}.
Multi-head latent attention (MLA) jointly compresses keys and values into a low-dimensional latent representation, reducing the inference KV cache while retaining softmax attention~\citep{liu2024deepseek}.
QK normalization controls logit growth using scaled $\ell_2$ normalization~\citep{henry2020qknorm} or LayerNorm~\citep{dehghani2023scaling,wortsman2024small}.

\paragraph{Recurrent and hybrid architectures.}
Linear attention replaces softmax with kernel inner products, enabling recurrent computation with a constant-size state~\citep{katharopoulos2020transformers}.
Related models use data-dependent gating (GLA, Mamba, and Mamba2)~\citep{yang2023gla,gu2023mamba,dao2024mamba2}, per-channel decay (RWKV)~\citep{peng2023rwkv}, or delta-rule updates (DeltaNet)~\citep{schlag2021linear,yang2024deltanet}.
Gated DeltaNet (GDN) combines gating and the delta rule~\citep{yang2025gdn}, updating its recurrent state as
\begin{equation}
\label{eq:app_gdn_recurrence}
\mathbf{S}_t
= \mathbf{S}_{t-1}\big(\alpha_t(\mathbf{I}
- \beta_t \mathbf{k}_t \mathbf{k}_t^\top)\big)
+ \beta_t \mathbf{v}_t \mathbf{k}_t^\top.
\end{equation}
To address limitations in retrieval~\citep{li2025minimax} and in-context learning~\citep{lenz2025jamba}, hybrid models interleave recurrent and softmax layers.
Nemotron-H devotes approximately 8\% of its layers to attention~\citep{blakeman2025nemotronh}; Kimi Linear uses three KDA layers per MLA layer~\citep{team2025kimilinear}; and Qwen3-Next uses three GDN layers per gated-attention layer~\citep{qwen2025next,qwen2025nextcard}.

\subsection{Proof of the Quantization Residual}
\label{app:proofs}

We derive \eqref{eq:delta_quantization_residual} for a fixed attention head.
Indices $i$, $j$, and $k$ denote query positions, key positions, and value channels, respectively.

\paragraph{Step 1. Fix the operands and assumptions.}
Throughout Section~\ref{sec:the_delta} and this proof, $P$ and $dO$ denote the FP32 softmax probabilities and BF16 output gradient before FP8 quantization.
FP8-quantized operands $\widehat{P}$, $\widehat{dO}$, and $\widehat{V}$ are represented by their dequantized values including block scales.
To isolate operand-quantization effects, we assume exact row normalization,
\begin{equation}
\label{eq:app_delta_normalization}
\sum_j P_{ij}=1.
\end{equation}
For causal attention, masked entries of both $P$ and $\widehat{P}$ are zero, so all sums may include the full key index range.
We assume FP32 accumulation and output for the forward product $\widehat{P}\widehat{V}$ and the backward product $\widehat{dO}\widehat{V}^\top$.
The same effective $\widehat{V}$ is used in both passes, and we neglect FP32 accumulation and reduction rounding, as well as any output-storage rounding.
Thus $\widehat{O}$ and $\widehat{dP}$ denote the FP32 outputs computed from quantized operands, with components
\begin{equation}
\label{eq:app_delta_operand_products}
\widehat{O}_{ik}=\sum_j\widehat{P}_{ij}\widehat{V}_{jk},
\qquad
\widehat{dP}_{ij}=\sum_k\widehat{dO}_{ik}\widehat{V}_{jk}.
\end{equation}
The saved-output correction uses the BF16 output gradient $dO$, while $\widehat{dP}$ uses its FP8-quantized counterpart $\widehat{dO}$.
Thus the stale correction and score gradient are
\begin{equation}
\label{eq:app_delta_stale_definitions}
\delta_i^{\mathrm{stale}}=\sum_k dO_{ik}\widehat{O}_{ik},
\qquad
dS_{ij}^{\mathrm{stale}}
=P_{ij}\bigl(\widehat{dP}_{ij}-\delta_i^{\mathrm{stale}}\bigr).
\end{equation}

\paragraph{Step 2. Expand the row sum.}
Define the row-sum residual as $\varepsilon_i$.
Because $\delta_i^{\mathrm{stale}}$ is constant across row $i$, \eqref{eq:app_delta_normalization} gives
\begin{equation}
\label{eq:app_delta_expand_row_sum}
\begin{aligned}
\varepsilon_i
=\sum_j dS_{ij}^{\mathrm{stale}}
&=\sum_j P_{ij}\widehat{dP}_{ij}
-\delta_i^{\mathrm{stale}}\sum_j P_{ij}\\
&=\sum_j P_{ij}\widehat{dP}_{ij}
-\sum_k dO_{ik}\widehat{O}_{ik}.
\end{aligned}
\end{equation}

\paragraph{Step 3. Substitute the products and interchange sums.}
Substituting \eqref{eq:app_delta_operand_products} into \eqref{eq:app_delta_expand_row_sum} and interchanging the finite sums yields
\begin{equation}
\label{eq:app_delta_reorder_products}
\begin{aligned}
\varepsilon_i
&=\sum_j P_{ij}\Bigl(\sum_k\widehat{dO}_{ik}\widehat{V}_{jk}\Bigr)
-\sum_k dO_{ik}\Bigl(\sum_j\widehat{P}_{ij}\widehat{V}_{jk}\Bigr)\\
&=\sum_k\widehat{dO}_{ik}\Bigl(\sum_j P_{ij}\widehat{V}_{jk}\Bigr)
-\sum_k dO_{ik}\Bigl(\sum_j\widehat{P}_{ij}\widehat{V}_{jk}\Bigr).
\end{aligned}
\end{equation}

\paragraph{Step 4. Add and subtract the common mixed term.}
Define the term that combines $\widehat{dO}$ with the quantized forward probabilities,
\begin{equation}
\label{eq:app_delta_mixed_term}
M_i=\sum_k\widehat{dO}_{ik}\Bigl(\sum_j\widehat{P}_{ij}\widehat{V}_{jk}\Bigr).
\end{equation}
Adding and subtracting $M_i$ in \eqref{eq:app_delta_reorder_products} separates the two operand mismatches,
\begin{equation}
\label{eq:app_delta_factor_errors}
\begin{aligned}
\varepsilon_i
&=\Bigl[\sum_k\widehat{dO}_{ik}\Bigl(\sum_j P_{ij}\widehat{V}_{jk}\Bigr)-M_i\Bigr]
+\Bigl[M_i-\sum_k dO_{ik}\Bigl(\sum_j\widehat{P}_{ij}\widehat{V}_{jk}\Bigr)\Bigr]\\
&=\sum_k\widehat{dO}_{ik}\Bigl(\sum_j(\underbrace{P_{ij}-\widehat{P}_{ij}}_{\substack{\text{probability}\\\text{quantization error}}})\widehat{V}_{jk}\Bigr)
+\sum_k(\underbrace{\widehat{dO}_{ik}-dO_{ik}}_{\substack{\text{gradient}\\\text{quantization error}}})\Bigl(\sum_j\widehat{P}_{ij}\widehat{V}_{jk}\Bigr).
\end{aligned}
\end{equation}
This is \eqref{eq:delta_quantization_residual}.
The first term captures the mismatch between $P$ in the score gradient and $\widehat{P}$ in the saved forward output.
The second captures the mismatch between $\widehat{dO}$ in the backward product and $dO$ in the saved-output correction.
The contributions need not vanish or cancel, so a nonzero row sum is possible but not guaranteed.
The assumptions exclude additional residuals from finite-precision normalization, accumulation, output storage, and reductions.
The derivation applies before any subsequent quantization of $dS$.

\paragraph{Zero column sums of the key gradient.}
For the queries and keys entering $S=QK^\top/\sqrt{d}$, the key gradient is $dK=dS^\top Q/\sqrt{d}$.
Using the exact-arithmetic zero-row-sum invariant in \eqref{eq:delta_zero_row_sum},
\begin{equation}
\label{eq:delta_zero_column_sum_dk}
\sum_j dK_{jc}
=\frac{1}{\sqrt{d}}\sum_j\sum_i dS_{ij}Q_{ic}
=\frac{1}{\sqrt{d}}\sum_i Q_{ic}\sum_j dS_{ij}=0.
\end{equation}
Thus, the key gradients sum to zero across tokens in each channel $c$.

\subsection{Limitations}
\label{app:limitations}

\paragraph{Infrastructure and performance.}
Delta-Matching enables FP8 GEMMs in FFNs, attention projections, and all seven
attention-core matmuls, but our training stack is not fully optimized.
The correction uses a separate pass, leaving room for kernel fusion, improved
memory reuse, and integration with neighboring quantizers.
Further opportunities include reducing context-parallel communication costs by
exchanging attention activations in FP8 and scheduling kernels to overlap
recomputation with other backward computations.
Our throughput measurements isolate attention kernels rather than end-to-end
training performance or distributed scaling (Appendix~\ref{app:kernel_throughput}).
They exclude input quantization and autograd-wrapper overhead for our FP8 kernels,
while the baselines include wrapper overhead and cuDNN/TE FP8 also includes input casting.
Despite speedups over FA3 BF16, Delta-Matching is slower than cuDNN BF16 at
head dimension 128 under this accounting.
We have not compared runtime against the quality-preserving BF16-$dP$ alternative
in Appendix~\ref{app:protect_fragile_operation}.
Turning numerical feasibility into system-level speed and memory savings remains
an important engineering direction.
Nearly full FP8 refers to these GEMM operands, not every operator or stored
tensor (Appendix~\ref{app:train_precision}).

\paragraph{Architecture and training scale.}
Our experiments reach 5.29B parameters with 30B-token base pretraining, but do not
establish how stale-delta degradation behaves at substantially larger scales
or longer durations.
The 2.58B pure-GQA and 1.67B Mamba-2/GQA models exhibit severe attention-layer
gain corruption despite small validation-loss gaps
(Appendix~\ref{app:convergent_architectures}).
We have not established whether further scaling these architectures would turn
that corruption into substantial training-loss or downstream degradation.
Masking probes suggest compensation elsewhere in the network but do not identify
the compensating components or predict when that compensation might fail.

\paragraph{Optimizer coverage.}
Most experiments use AdamW, supplemented by a 28,610-step Muon-based study
with one model configuration and initialization (Appendix~\ref{app:muon}).
Muon updates hidden matrices, while QK-normalization gains and other remaining
parameters still use AdamW.
The row-correction identity is optimizer-independent, but broader Muon coverage
and other optimizer recipes remain untested.

\paragraph{Scope of the numerical guarantee.}
Delta-Matching restores the zero-row-sum invariant under the stated assumptions,
up to FP32 rounding and before subsequent $dS$ quantization
(Section~\ref{sec:delta_matching}).
It does not remove ordinary FP8 operand errors or guarantee convergence and
BF16-equivalent performance for arbitrary models and training recipes.
Our evidence supports error accumulation over optimizer updates, but does not
provide a quantitative law predicting the onset or magnitude of performance loss.

\section{Experiment Setup}
\label{app:exp_setup}

We document the data recipes, training configurations, evaluation protocols,
and FP8 attention kernel numerics. We also define reporting conventions and the
quantities plotted in Figure~\ref{fig:delta_dynamics}.

\subsection{Data Recipe}
\label{app:data_recipe}

\paragraph{Nemotron-CC v2.1.}
We use Nemotron-CC v2.1~\citep{su2025nemotroncc,blakeman2025nemotron3nano},
restricted to its \texttt{High-Quality} and \texttt{High-Quality-DQA} subsets.
Text is tokenized with the 32,000-entry Llama-2 tokenizer~\citep{touvron2023llama}.
Each document is wrapped with BOS and EOS tokens, then concatenated into a token stream.
We construct a validation split by independently holding out each document
with probability 0.02 using a fixed seed, then shuffling the held-out documents.
The validation stream contains 740,467 documents and 807,013,275 tokens.
Packed windows can cross document boundaries without resetting positions or recurrent state.

\paragraph{ProLong-64K.}
For context extension, ProLong-64K data~\citep{gao2025prolong} are retokenized with the same Llama-2
tokenizer and repacked into 65,536-token chunks.
Held-out windows are disjoint from the training pool.
The validation stream contains 256 chunks totaling 16,777,216 tokens.
Appendix~\ref{app:eval} specifies how these streams are scored.

\subsection{Training Configuration}
\label{app:train_config}

\subsubsection{Model Architectures}
\label{app:train_architectures}

Table~\ref{tab:train_architectures} summarizes the model geometries.
Each hybrid repeats three recurrent layers followed by one softmax-attention layer.
All models use SwiGLU FFNs, RMSNorm before each token mixer and FFN,
a final RMSNorm, and tied input and output embeddings with a 32,000-token vocabulary.
RMSNorm uses $\epsilon=10^{-6}$ and FP32 computation.
Unless stated otherwise, attention uses query and key RMSNorm and a sigmoid output gate.
Hybrid GQA applies RoPE to one quarter of each head with $\theta=10^7$.

\begin{table}[!htbp]
\centering
\caption{Model architectures. Parameter counts are in billions, rounded to four decimal places.
$L_r$ and $L_a$ count recurrent and softmax-attention layers;
$h_q/h_{kv}$ gives the query and key-value head counts, and $d$ is the attention head dimension.
For MLA, the head counts describe the expanded tensors entering the attention core.}
\label{tab:train_architectures}
\small
\setlength{\tabcolsep}{3pt}
\begin{tabular*}{\linewidth}{@{\extracolsep{\fill}}lrrrrrrr@{}}
\toprule
Model & Params (B) & $L_r/L_a$ & Hidden & FFN & $h_q/h_{kv}$ & $d$ & RoPE fraction \\
\midrule
GDN/GQA & 0.5693 & 15/5 & 1280 & 3840 & 6/2 & 256 & $1/4$ \\
GDN/GQA & 1.6654 & 18/6 & 2048 & 6144 & 16/4 & 128 & $1/4$ \\
GDN/GQA & 1.6654 & 18/6 & 2048 & 6144 & 8/2 & 256 & $1/4$ \\
GDN/GQA & 5.2886 & 24/8 & 3072 & 9216 & 16/4 & 256 & $1/4$ \\
GDN/MLA & 1.6654 & 18/6 & 2048 & 6144 & 16/16 & 128 & $1/2$ \\
KDA/GQA & 1.6648 & 18/6 & 2048 & 8064 & 16/4 & 128 & $1/4$ \\
Pure GQA & 1.4749 & 0/28 & 2048 & 6144 & 16/8 & 128 & $1$ \\
Mamba-2/GQA & 1.6665 & 18/6 & 2048 & 7104 & 16/4 & 128 & $1/4$ \\
\bottomrule
\end{tabular*}
\end{table}

\paragraph{GDN/GQA and attention variants.}
GDN layers use key dimension 128, value dimension 256, width-4 convolutions on
queries, keys, and values, and gated output RMSNorm.
Key/value head counts are 10/10, 16/16, and 16/32 at 569M, 1.67B, and
5.29B parameters, respectively.
The NoPE and no-QK-norm variants use the 1.67B, head-dimension-128 geometry.
NoPE removes RoPE only from the GQA layers, leaving the recurrent layers unchanged;
no-QK-norm removes only the query and key RMSNorms.
The Head Dim 256 variant uses 8 query and 2 KV heads instead of 16 and 4,
preserving the total query and KV widths.
Its precision and implementation settings also differ from the base configuration,
as specified below.

\paragraph{GDN/MLA and KDA/GQA.}
MLA uses a KV latent of dimension 384 and a shared rotary key of dimension 64.
A linear projection maps the 2048-dimensional hidden state to their combined
448-dimensional representation.
After RMS normalization, the latent is projected to 1024-dimensional keys and
2048-dimensional values.
Each query/key head contains 64 rotary and 64 non-rotary channels, normalized
separately, with $\theta=10^7$.
The GDN layers are unchanged from the base model.
KDA instead uses 16 recurrent heads with key and value dimensions of 128,
rank-128 decay and output gates, width-4 convolutions, and gated output RMSNorm.
Its FFN width increases to 8064 for approximate parameter parity.

\paragraph{Pure GQA and Mamba-2/GQA.}
The pure GQA model has 28 attention layers, no recurrent layers or attention output
gate, and full-head RoPE with $\theta=10^6$.
The Mamba-2 hybrid retains the base GQA configuration and replaces GDN with
Mamba-2 layers using 64 recurrent heads of dimension 64, state size 128,
one group, expansion factor 2, convolution width 4, chunk size 256, and gated RMSNorm.
Its FFN width increases to 7104 for approximate parameter parity.

\subsubsection{Training GEMM Precision}
\label{app:train_precision}

Table~\ref{tab:train_precision} distinguishes attention-core precision from FFN
and attention-projection precision.
P0--P4 describe GEMM operands, not every operation or stored tensor in the model.
FP8 attention retains higher precision for softmax, row reductions, and accumulation.
Recurrent layers remain in BF16 with higher-precision internal computations;
normalization, embeddings, and the output head are not converted to FP8 linear layers.
Master parameters, AdamW states, and gradient reductions remain in FP32.
The full FP8 configuration P4 thus covers the attention core, FFNs, and projections,
not all model components.

\begin{table}[!htbp]
\centering
\caption{Training GEMM operand precision. Attention projections denote QKVO for
GQA and include the latent and up-projections for MLA. Precision scope is independent
of stale, consistent-$dO$, or matched delta construction.}
\label{tab:train_precision}
\small
\setlength{\tabcolsep}{4pt}
\begin{tabular*}{\linewidth}{@{\extracolsep{\fill}}llccc@{}}
\toprule
ID & Configuration & Attention core & FFNs & Projections \\
\midrule
P0 & BF16 reference & BF16 & BF16 & BF16 \\
P1 & FP8 FFNs only & BF16 & FP8 & BF16 \\
P2 & Attention-core-only FP8 & FP8 & BF16 & BF16 \\
P3 & FP8 attention and FFNs & FP8 & FP8 & BF16 \\
P4 & FP8 attention, FFNs, and projections & FP8 & FP8 & FP8 \\
\bottomrule
\end{tabular*}
\end{table}

Our FP8 attention uses E4M3 operands with block scaling.
Appendix~\ref{app:kernel_numerics} specifies our FP8 attention quantizers,
accumulation windows, and backward passes.
FP8 linear layers use tensorwise dynamic scaling, with E4M3 inputs and weights
and E5M2 output gradients for forward, input-gradient, and weight-gradient GEMMs.
The cuDNN/Transformer Engine (TE) baseline uses HYBRID delayed scaling,
a maximum-amplitude history of length 16, and BF16 external inputs and outputs.

Naive FP8 uses the saved-output correction, while consistent $dO$ uses the same
rounded output-gradient values as the backward GEMM in that correction.
Delta-Matching instead forms the correction from the backward operands as defined
in Section~\ref{sec:delta_matching}.
The cuDNN/TE baseline uses its own saved-output correction.

\subsubsection{Optimization and Training Recipes}
\label{app:train_recipes}

\paragraph{Pretraining.}
Unless noted otherwise, stage-1 runs train for 28,610 optimizer steps on the Nemotron-CC training split
in Appendix~\ref{app:data_recipe}.
The global batch contains 128 sequences of 8192 tokens, giving 1,048,576 tokens
per step and approximately 30B tokens overall.
Except for the Muon-based study in Appendix~\ref{app:muon}, we use AdamW~\citep{kingma2014adam, loshchilov2017decoupled} with $\beta_1=0.9$, $\beta_2=0.95$, $\epsilon=10^{-8}$, weight decay
0.1, and global gradient-norm clipping at 1.0.
One-dimensional parameters and embeddings, including the tied output head, are
excluded from weight decay.
The learning rate warms up linearly for 952 steps to $2.4\times10^{-3}$, remains
constant through step 22,888, and decays linearly to zero over the last 5722 steps.
The objective adds a z-loss of $10^{-4}$ times the mean squared log-partition
to next-token cross-entropy.
We use no dropout and keep the data order fixed across runs;
the reported run seeds vary model initialization.

\paragraph{Initialization.}
Linear weights and embeddings use a zero-mean Gaussian with standard deviation
0.02; biases are zero.
Weights named \texttt{o\_proj} or \texttt{down\_proj} are additionally scaled by
$1/\sqrt{2L}$ for $L$ decoder layers.
Recurrent parameters outside Linear modules, including convolutions and state-space
parameters, retain their library initialization.
MLA's value up-projection is an exception, with standard deviation
$0.02\sqrt{2048/384}$.

\paragraph{Context extension.}
Every method starts from the completed BF16/FP32 reference checkpoint of its
initialization seed, with a fresh optimizer, rather than from its own stage-1 arm.
We train on ProLong-64K for 28,000 steps with 16 sequences of 65,536 tokens per
global batch, preserving 1,048,576 tokens per step and adding approximately 29.4B tokens.
The learning rate warms up for 100 steps to $8\times10^{-4}$, stays constant
through step 26,000, and decays linearly to zero at step 28,000.
Other optimizer settings follow pretraining.
YaRN uses extension factor 8, $\beta_{\mathrm{fast}}=32$, and
$\beta_{\mathrm{slow}}=1$; the post-RoPE query is multiplied by
$(1+0.1\ln 8)^2\approx1.459$.
At step 18,000, training resumes from the checkpoint with optimizer state and
data-order continuity, while the selected corpus fraction expands from the first
60\% to the first 85\%.
Transformer Engine's amplitude history is not checkpointed and restarts at this
transition.
All reported context-extension results use the final 28,000-step export.

\subsubsection{Experiment Configurations}
\label{app:train_experiments}

Table~\ref{tab:train_experiments} maps each result group to its model,
precision scope, and hardware.
MP denotes the BF16/FP32 attention reference, which already uses FP8 FFNs in
P1 configurations.
Its attention projections remain BF16, so MP--P4 comparisons also change
projection precision.
The naive and Delta-Matching arms within each main comparison share the same
precision scope and attention implementation, differing in delta construction.
Experiments use two initialization seeds unless otherwise noted.

\begin{table}[!htbp]
\centering
\caption{Configurations behind the result tables and diagnostic figures. P0--P4 are defined in
Table~\ref{tab:train_precision}. Each run uses eight GPUs of the listed type.
The FP8-linear configurations are detailed in Appendix~\ref{app:fp8_linear}.}
\label{tab:train_experiments}
\small
\setlength{\tabcolsep}{3pt}
\begin{tabular*}{\linewidth}{@{\extracolsep{\fill}}llccc@{}}
\toprule
Results & Model or variant & MP & FP8 & GPUs \\
\midrule
Figure~\ref{fig:delta_dynamics} & 1.67B GDN/GQA, $d=128$ & P1 & P4 & H100 \\
Figure~\ref{fig:app_muon} & 1.67B GDN/GQA, Muon, $d=128$ & P1 & P4 & H100 \\
Table~\ref{tab:app_muon_full} & 1.67B GDN/GQA, Muon and AdamW, $d=128$ & P1 & P4 & H100 \\
Table~\ref{tab:exp_cmp_overall} & Default, 1.67B GDN/GQA, $d=128$ & P1 & P4 & H100 \\
Table~\ref{tab:exp_attn_arch} & NoPE & P1 & P4 & H100 \\
Table~\ref{tab:exp_attn_arch} & No QK-norm & P1 & P4 & H200 \\
Table~\ref{tab:exp_attn_arch} & Head Dim 256 & P0 & P2 & H100 \\
Table~\ref{tab:exp_model_arch} & 1.67B GDN/MLA & P1 & P4 & H200 \\
Table~\ref{tab:exp_model_arch} & 1.67B KDA/GQA & P1 & P4 & H200 \\
Table~\ref{tab:exp_model_arch} & 1.67B Mamba-2/GQA & P1 & P4 & H200 \\
Table~\ref{tab:exp_param_scale} & 569M GDN/GQA & P1 & P4 & H100 \\
Table~\ref{tab:exp_param_scale} & 5.29B GDN/GQA & P1 & P4 & H100 \\
Table~\ref{tab:exp_training_stages} & 1.67B GDN/GQA, 64K & P1 & P4 & H100 \\
Table~\ref{tab:exp_fp8_linear} & 1.67B GDN/GQA, $d=256$ & P0/P1 & P3/P4 & H100 \\
Table~\ref{tab:app_pure_gqa_eval} & 1.475B pure GQA & P0 & P2 & H100 \\
\bottomrule
\end{tabular*}
\end{table}

\paragraph{Training dynamics in Figure~\ref{fig:delta_dynamics}.}
\label{app:delta_dynamics_setup}
The BF16/FP32, stale-delta, and Delta-Matching runs share one initialization seed
and data order, following the pretraining recipe in Appendix~\ref{app:train_recipes}.
The diagnostics cover the first 6,000 steps of the 28,610-step learning-rate schedule,
with panels (a, b) using step-6,000 checkpoints and panel (c) recording gains every 25 steps.
Offline probe details are given in Appendix~\ref{app:figure3_quantities}.

\paragraph{Hardware and batching.}
Each run uses a single node with eight H100 GPUs (80GB each) or eight H200 GPUs
(141GB each), with FSDP2 across all eight GPUs.
Master parameters remain in FP32, while forward/backward parameters are BF16
before any FP8 operand casts.
At 8K context, the usual per-GPU batch and accumulation count are 2 and 8;
they are 4 and 4 for 569M and H200 runs, and 1 and 16 for 5.29B.
At 64K, they are 1 and 2.
Selective activation checkpointing is enabled for Head Dim 256, FP8-linear,
pure-GQA, 569M, and context-extension runs, and for the first 5.29B seed;
it is disabled otherwise.

\subsection{Evaluation Configuration}
\label{app:eval_configuration}

All methods are evaluated from BF16 Hugging Face exports through the same
inference path; FP8 is used during training, not benchmark evaluation.
Exports preserve GQA or MLA, the recurrent layers, and the layer pattern.
Attention uses FlashAttention-2~\citep{dao2024flashattention2}, with a BF16 KV
cache for generation; GatedDeltaNet uses \texttt{fla} kernels with FP32 recurrent state.
The bundled tokenizer matches Llama-2, and padding uses the EOS token.
Harness-based benchmarks use unmodified \texttt{lm-evaluation-harness}
0.4.12 task definitions~\citep{gao2021lmevalharness}
with our model wrapper; RULER uses a separate evaluator.
RULER, extraction, and TriviaQA use continuous-batching generation with greedy
selection from FP32 logits; CS9 and MMLU use the model's forward pass.
Their likelihood scores use BF16 log-softmax and BF16 answer sums, with exact
ties resolved in favor of the first choice.
Validation CE runs on one GPU with batch size 4 at 8K and 1 at 64K.
Downstream evaluation is data-parallel over eight GPUs, with per-GPU batch sizes
32 for CS9, extraction, and TriviaQA, 16 for MMLU and base RULER, and 4 for
context-extension RULER.
The environment uses PyTorch 2.12.0+cu130, Transformers 5.9.0,
\texttt{flash\_attn} 2.8.3.post1, and \texttt{flash-linear-attention} 0.5.0.

\subsection{Reporting Conventions and Uncertainty}
\label{app:conventions_uncertainty}

Unless otherwise stated, tables report the arithmetic mean $\pm$ one standard error (SE) across initialization
seeds. The SE is $s/\sqrt{n}$, where $s$ is the sample standard deviation and
$n$ is the seed count specified in each caption.
Reported benchmark error bars do not include benchmark-sampling uncertainty.
For CS9 and extraction, we first compute each aggregate within a seed,
then calculate its SE across seeds rather than averaging the component SEs.
For metric values $x_1,\ldots,x_n$ from the individual seeds,
\begin{equation}
\label{eq:app_sample_std}
s = \sqrt{\frac{1}{n-1}\sum_{i=1}^{n}(x_i-\bar{x})^2},
\qquad
\bar{x} = \frac{1}{n}\sum_{i=1}^{n}x_i.
\end{equation}
For paired validation-loss contrasts, the SE is the sample standard deviation of
per-window loss differences divided by $\sqrt{N}$, with $N=2{,}000$ windows at 8K
or $N=255$ at 64K.
For visual clarity, loss plots show only the first seed of each setting and use a
logarithmic $y$-axis unless otherwise noted; tabulated metrics summarize all reported seeds.

\paragraph{Quantities in Figure~\ref{fig:delta_dynamics}.}
\label{app:figure3_quantities}
Panels (a, b) probe layer 23 on 8,192-token windows.
For all three runs, we obtain activations and output gradients by rerunning
each checkpoint entirely in BF16, including attention and FFN layers.
This does not exactly replay training, which used FP8 FFNs in every run.
\textbf{(a)} Let $P_8$ be the softmax probabilities from logits computed with FP8
$Q,K$ operands, before probability quantization, and $P_t$ the BF16/FP32 reference probabilities.
\emph{Attention weight} is the post-update probability
$P'_i=\mathrm{softmax}(S_i-\eta_i dS_i)$.
Both FP8 updates start from $S_i=\log P_{8,i}$, while the reference starts from
$\log P_{t,i}$, over the causal row.
The matched gradient is $dS_{8,i}=P_{8,i}\odot(dP_{8,i}-\delta_i^{\mathrm{matched}})$,
where $dP_8$ uses FP8 $dO,V$ operands and
$\delta_i^{\mathrm{matched}}=\sum_j P_{8,ij}dP_{8,ij}$.
Stale delta adds $\varepsilon_iP_{8,i}$, with
$\varepsilon_i=\delta_i^{\mathrm{matched}}-\delta_i^{\mathrm{stale}}$;
only this FP8 pair isolates the backward correction.
The reference gradient $dS_t$ uses unquantized operands, and all three updates
share $\eta_i=1/\max_j|dS_{t,ij}|$.
Let $T_i$ index the eight largest entries of the pre-update $P_{t,i}$ for row selection.
The plot shows the four keys with the largest post-update BF16/FP32 probabilities;
\emph{Key rank} orders them by those probabilities.
Entropy annotations describe the full post-update row in nats.
Rows are selected from window 100 of the step-6,000 Delta-Matching checkpoint,
requiring $i\geq1024$, reference entropy between 1 and 3 nats, and
$F_i=\|P_{8,i}-P_{t,i}\|_2/\|P_{t,i}\|_2\leq0.05$.
Within each residual sign, rows are ranked by
$|\varepsilon_i|\,\|P_{8,i}[T_i]\|_2/\|(dS_{8,i}-dS_{t,i})[T_i]\|_2$.
The left row (head 5, row 5780) ranks first among negative residuals; the right
(head 5, row 1185) ranks fourth among positive residuals and first within head 5.
Their ranking ratios are 32.2 and 12.7, and forward differences are 1.8\% and 0.5\%.
Delta-Matching's plotted weights differ from BF16/FP32 by at most 0.016 per key.

\textbf{(b)} Each dot is a channel, colored by run, with channels 7, 23, 8, and 24 enlarged.
The attention-weighted channel contribution is
$L_{ic}=Q_{ic}\sum_j P_{8,ij}K_{jc}/\sqrt{d}$, using unquantized post-RoPE $Q,K$ and $d=128$.
It measures a post-RoPE channel contribution, not a derivative with respect to the pre-RoPE gain.
For each of 192 window/head samples (windows 100--111, each with 16 heads),
$\Delta_{rc}$ is the mean $L_{ic}$ over rows with $\varepsilon_i<0$ minus its mean
over rows with $\varepsilon_i>0$.
\emph{Sharpened-row preference} is $\bar{\Delta}_c/(s_c/\sqrt{192})$, where
$\bar{\Delta}_c$ and $s_c$ are the mean and sample standard deviation of these 192 differences.
This is a descriptive one-sample $t$-statistic, not a mean/SD effect size.
The SE ignores dependence among heads sharing a window, so the statistic
does not provide calibrated statistical significance.
For BF16/FP32 and Delta-Matching, the probe computes the hypothetical stale-FP8
residual on each model's own layer-23 activations and output gradients;
neither run applies this residual during training.
For the stale run, the probe estimates the residual its kernel applies.
\emph{Push consistency} is $-m/\sqrt{v}$, where $m,v$ are each run's actual
step-6,000 Adam first and second gradient moments for the key gain;
positive values indicate upward adaptive-update pressure.
The axes show a descriptive association, not an isolated estimate of the residual's contribution to gain updates.

\textbf{(c)} \emph{Gain} is the learned key RMSNorm weight $g_{k,c}$ applied before
RoPE, shared across heads and initialized to 1.
For token $j$ and channel $c$, RMSNorm computes
\begin{equation}
\label{eq:app_key_rmsnorm}
\widetilde{k}_{j,c}
=g_{k,c}\frac{k_{j,c}}
{\sqrt{\frac{1}{d}\sum_{r=1}^{d}k_{j,r}^{2}+\epsilon_{\mathrm{norm}}}},
\end{equation}
where $\epsilon_{\mathrm{norm}}$ is the numerical stabilizer.
Figure~\ref{fig:delta_gain_growth} records the learned parameter $g_{k,c}$ directly,
not a separately computed statistic.
Thick lines show $g_{k,7}$ and thin lines the other 127 channels, recorded every 25
steps with the same initialization seed, data order, and learning-rate schedule.
The BF16/FP32 trace records FP32 parameters just after the update, one update later
than the FP8 runs' logging of BF16 forward copies.
The traces therefore have a one-step timing offset and small storage-rounding differences.
Around step 6,000, $g_{k,7}$ reaches 5.4 with stale delta versus 1.2 with Delta-Matching.

\subsection{Evaluation Suite}
\label{app:eval}

\paragraph{Common Settings.}
\label{app:eval_reporting}
We specify task variants, prompts, and metrics for reproducibility~\citep{biderman2024evallessons}.
All downstream tasks use their full evaluation split.
Prompts have no added BOS token or chat template.
CS9, MMLU, and extraction are zero-shot; TriviaQA uses five demonstrations.
Generation is greedy, while CS9 and MMLU use likelihood scoring.

\paragraph{Validation Cross-Entropy.}
\label{app:eval_ce}
At 8K, we score the first 2,000 contiguous windows of our shuffled Nemotron-CC
validation stream.
Each window contains 8,193 token IDs and yields 8,192 next-token predictions,
for 16,384,000 predictions in total.
Adjacent windows share one boundary token but no predicted positions.
Within each window, causal attention crosses document boundaries without resetting
positions or recurrent state.
The BF16 forward pass produces logits that are cast to FP32 for token
cross-entropy and batch sums; running totals use FP64.
For $N$ validation windows of $L+1$ token IDs, we report
\begin{equation}
\label{eq:app_validation_ce}
\mathrm{CE}_{\mathrm{val}}(\theta)
= -\frac{1}{NL}\sum_{w=1}^{N}\sum_{t=1}^{L}
\log p_{\theta}\!\left(x_{t+1}^{(w)}\mid x_{1:t}^{(w)}\right),
\end{equation}
where $x_t^{(w)}$ is token $t$ of window $w$ and $p_{\theta}$ is the evaluated
model's next-token distribution. Natural logarithms give mean negative
log-likelihood over $NL$ predicted tokens in nats per token.
All methods use identical windows.
The 64K protocol instead uses held-out ProLong data~\citep{gao2025prolong}.
Its 256 packed chunks of 65,536 tokens yield 255 prediction windows because each
requires 65,537 token IDs, giving 16,711,680 predictions.
Context-extended models are also evaluated on the same 8K Nemotron windows.

\paragraph{Extraction Recall.}
\label{app:eval_extraction}
We use the Based extraction suite~\citep{arora2024simple}, comprising
SWDE~\citep{lockard2019openceres}, FDA~\citep{arora2023evaporate}, and SQuAD
completion~\citep{rajpurkar2018squad}.
The datasets are \texttt{hazyresearch/based-swde-v2},
\texttt{hazyresearch/based-fda}, and \texttt{hazyresearch/based-squad}, with
1,111, 1,102, and 2,984 validation examples, respectively.
We use harness task version 0, which does not strip prompt or target whitespace.
Generation budgets are 48, 48, and 256 tokens; SWDE and FDA stop at a newline,
whereas SQuAD completion stops at a blank line. EOS also terminates generation.
The 7,936-token input limit does not truncate any prompt.
Each example scores one if its single supplied gold value occurs anywhere in
the continuation as a case-insensitive literal substring, with no other normalization.
This is neither SQuAD exact match nor token F1.
The aggregate is the unweighted mean of the three task scores.

\paragraph{RULER.}
\label{app:eval_recall}
We regenerate the 13 RULER tasks~\citep{hsieh2024ruler} with the Llama-2 tokenizer
and a fixed generation seed, using 500 examples per task and length.
Five essay-haystack retrieval tasks use Project Gutenberg novels instead of
Paul Graham essays, and the 4K \texttt{qa\_2} task uses a patched generator.
Our scores are therefore not directly comparable to published RULER results.
The base protocol evaluates lengths 4,096 and 8,192; context-extension evaluation
also includes 16,384, 32,768, and 65,536.
Nominal lengths include the generation budget, and prompts are not truncated.
Generation budgets are 128 tokens for retrieval, 30 for variable tracking,
120 for common-word extraction, 50 for frequent-word extraction, and 32 for QA.
Configured newline stops do not trigger with this tokenizer; generation ends
at the token budget or a runtime EOS token (IDs 2 or 11).
The official scorer measures the fraction of required references found as
case-insensitive substrings for retrieval, variable tracking, and word extraction;
QA scores one if any accepted reference is found.
For each length, we average the 13 task scores, each rounded to two decimals.
Empty predictions score zero, while missing predictions abort evaluation.
Per-task results appear in Appendix~\ref{app:ruler_results}.

\paragraph{MMLU.}
\label{app:eval_mmlu}
We evaluate all 14,042 test questions from \texttt{cais/mmlu} using the
zero-shot default harness task~\citep{hendrycks2020mmlu}.
The model selects among answer labels A--D by conditional likelihood.
The reported \texttt{acc} is weighted by each subject's question count,
not an unweighted mean of the 57 subject scores.
No prompt reaches the 4,096-token input limit.

\paragraph{Commonsense Individual Benchmark Accuracy.}
\label{app:eval_cs9}
We evaluate nine individual commonsense benchmarks in the zero-shot setting
and report their unweighted mean as CS9.
We use character-length-normalized multiple-choice accuracy (\texttt{acc\_norm})
for HellaSwag~\citep{zellers2019hellaswag}, PIQA~\citep{bisk2020piqa},
ARC-Easy and ARC-Challenge~\citep{clark2018arc}, SciQ~\citep{welbl2017sciQ},
and OpenBookQA~\citep{mihaylov2018openbookqa}.
LAMBADA~\citep{paperno2016lambada}, WinoGrande~\citep{sakaguchi2021winogrande},
and COPA~\citep{roemmele2011commonsensecausalreasoning} use \texttt{acc}.
For \texttt{lambada\_openai}, accuracy requires every token of the
target last word to be a greedy prediction; it is not multiple choice.
SciQ includes the support passage, OpenBookQA uses only the question and choices,
and COPA uses the SuperGLUE release~\citep{wang2019copa}.
We use the full test splits for LAMBADA, ARC-Easy, ARC-Challenge, SciQ, and
OpenBookQA, and full validation splits for HellaSwag, PIQA, WinoGrande, and COPA.
No prompt reaches the 4,096-token input limit.
Individual commonsense benchmark accuracies appear in Appendix~\ref{app:commonsense_benchmarks}.

\paragraph{TriviaQA.}
\label{app:eval_triviaqa}
We report five-shot, closed-book exact match on all 17,944 validation questions
from \texttt{mandarjoshi/trivia\_qa}, configuration
\texttt{rc.nocontext}~\citep{joshi2017triviaqa}.
For each question, the harness samples five training examples with a fixed few-shot
seed, using the first answer alias in each demonstration.
Greedy generation produces at most 32 tokens and stops at a newline, period,
comma, or EOS. No prompt reaches the 4,064-token input limit.
The \texttt{remove\_whitespace} filter strips surrounding whitespace.
Exact match then ignores case and ASCII punctuation and accepts any supplied
answer alias, but does not remove articles.
This differs from the original TriviaQA normalization; we do not report F1.

\subsection{FP8 Attention Kernel Numerics}
\label{app:kernel_numerics}

We summarize the numerical choices in our causal FP8 attention kernels for
head dimensions 128 and 256. These are the implementations used for the
kernel diagnostics and throughput measurements, not a specification of every
earlier ablation kernel or vendor baseline. The FA3-style probability encoding
is discussed separately in Appendix~\ref{app:fa3_max_shift}.

\subsubsection{Operands, Scaling, and Accumulation}
\label{app:kernel_operands}

Let $D$ be the head dimension and $\tau=D^{-1/2}$.
The inputs $q,k,v,dO$ are BF16, with $q,k$ taken after QK normalization and
RoPE where used. Before quantization, we multiply $q$ by $\tau$ in FP32 and subtract
the FP32 mean key over sequence positions from $k$, separately for each
channel and KV head. The fused RoPE producer rounds rotated channels to BF16
before applying $\tau$. Thus the dequantized query operand
$\bar Q=s_QQ_8\approx\tau q$ already includes the attention scale.
Native GQA shares key-value operands without repeating them.

For an input block $X$, E4M3 codes and their FP32 scale are
\begin{equation}
\label{eq:app_kernel_quantizer}
s_X=\mathrm{fl}_{32}\!\left(\frac{\max(\max|X|,10^{-30})}{448}\right),
\qquad
X_8=\mathcal{C}_8(X/s_X),
\end{equation}
where $\mathcal{C}_8$ rounds to nearest-even and saturates at $\pm448$.
Here $\mathrm{fl}_{32}$ denotes FP32 rounding, and decoded values are $s_XX_8$.
Input producers use IEEE division for $X/s_X$.
The $dO$ scale uses multiplication by $\mathrm{fl}_{32}(1/448)$ in place
of division. Input blocks span all channels and are separate for each sequence
and head. Transposed layouts reuse the same codes and scales.

Table~\ref{tab:app_kernel_dimensions} summarizes block sizes and accumulation
windows. Matrix products use FP8 codes with FP32 accumulator registers,
with block scales applied to their results. The registers do not imply full
FP32 internal accumulation of FP8 products. Long products therefore periodically
promote partial results into a separate FP32 accumulator.
Score and $dP$ products contract over the head dimension; elementwise arithmetic
and approximate base-two exponentials use FP32 fast math.
The following arithmetic rules apply at both head dimensions, with the
geometry differences listed in the table.

\begin{table}[!htbp]
\centering
\caption{Quantization blocks and accumulation windows of our FP8 attention kernels.
Input blocks span all channels. Score-gradient tiles are query rows by keys.}
\label{tab:app_kernel_dimensions}
\small
\setlength{\tabcolsep}{4pt}
\begin{tabular*}{\linewidth}{@{\extracolsep{\fill}}lcc@{}}
\toprule
Setting & $D=128$ & $D=256$ \\
\midrule
$Q,dO$ quantization blocks & 128 query rows & 64 query rows \\
$K,V$ quantization blocks & 64 keys & 32 keys \\
Forward probability groups & 64 keys & 32 keys \\
$dS$ quantization tiles & $64\times128$ & $64\times64$ \\
\midrule
Forward $PV$ promotion window & 256 keys & 128 keys \\
$dV$ promotion window & 512 query rows & 512 query rows \\
$dK$ promotion window & 64 query rows & 64 query rows \\
$dQ$ promotion window & 512 keys & 512 keys \\
\bottomrule
\end{tabular*}
\end{table}

\subsubsection{Forward Probability Encoding}
\label{app:kernel_forward}

The forward pass visits key tiles from the causal diagonal toward key zero.
It initializes a running row maximum $m=-\infty$, normalizer $\ell=0$, and
FP32 output accumulator $O_{\mathrm{acc}}=0$.
Let $\mathrm{TC}$ denote a Tensor Core code product with FP32 accumulator registers.
Scores are $S_{ij}=\mathrm{TC}(Q_{8,i},K_{8,j})\,\mathrm{fl}_{32}(s_Qs_K(j))$,
with future keys masked.
Each tile updates $m_{\mathrm{new}}=\max(m,\max_j S_{ij})$, computes
$\alpha=\exp(m-m_{\mathrm{new}})$, and replaces $m$ by $m_{\mathrm{new}}$.
For each query row and key group $g$, it encodes the exponentials as
\begin{equation}
\label{eq:app_kernel_forward_basis}
\begin{aligned}
\nu_g&=\max\!\left(\max_{j\in g}S_j,\,m-12\ln2\right),
&w_g&=\frac{\exp(\nu_g-m)s_V(g)}{448},\\
C_{P,j}^{\mathrm{fwd}}&=\mathcal{C}_8\!\left(448\exp(S_j-\nu_g)\right).
\end{aligned}
\end{equation}
The group maximum maps to 448 unless the floor at $m-12\ln2$ is active.
Within a tile, initialize $T=C_{P,g_0}^{\mathrm{fwd}}V_{8,g_0}$ and update
$T\gets T\,\mathrm{clip}(w_{g-1}/w_g,10^{-30},10^{30})+
C_{P,g}^{\mathrm{fwd}}V_{8,g}$ for each remaining group.
Rescaling uses FP32, while code products accumulate through the Tensor Core.
Promote once per tile using
$O_{\mathrm{acc}}\gets\alpha O_{\mathrm{acc}}+Tw_{\mathrm{last}}$.
Crucially, $\ell\gets\alpha\ell+\sum_j\exp(S_{ij}-m)$ uses exponentials
before the FP8 cast. The normalized output $O=O_{\mathrm{acc}}/\ell$ is stored in BF16, while
$\mathrm{LSE}=m+\ln\ell$ is stored in FP32.
This output-storage rounding is excluded from the ideal operand-error
decomposition in Section~\ref{sec:violate_invariant}.

\subsubsection{Backward Correction and Gradient Computation}
\label{app:kernel_backward}

The backward pass reuses the input codes and scales and recomputes scores and
$\widehat{dP}$ from FP8 operands. Its scaled FP32 probabilities are
\begin{equation}
\label{eq:app_kernel_backward_probability}
\Pi_{ij}=2^{\min\!\left(
\mathrm{TC}(Q_{8,i},K_{8,j})\,\mathrm{fl}_{32}(s_Qs_K\log_2e)
-\mathrm{fl}_{32}(\mathrm{LSE}_i\log_2e)+8,\,12\right)}.
\end{equation}
Ideally, with the exponent clamp inactive, $\Pi=256\exp(S-\mathrm{LSE})$;
probabilities for future keys are zero.
These reconstructed probabilities are not explicitly renormalized.
They are used before FP8 casting to form both the correction and $dS$.

\paragraph{Matched correction.}
Delta-Matching first recomputes the products needed for
$\delta_i^{\mathrm{matched}}=\sum_j(\Pi_{ij}/256)\widehat{dP}_{ij}$.
Here $A_{ij}=\mathrm{TC}(dO_{8,i},V_{8,j})$ and
$\widehat{dP}_{ij}=\mathrm{fl}_{32}(A_{ij}\,\mathrm{fl}_{32}(s_{dO}s_V))$.
The pass traverses keys in ascending order, reduces $\Pi\widehat{dP}$ over
32-key groups in FP32, and multiplies each partial by $2^{-8}$ in FP32.
It combines these partials in an FP64 accumulator and stores the final
correction in FP32.
The attention matmuls themselves remain FP8.
Stale delta skips this pass and instead computes $\sum_c dO_{ic}O_{ic}$
in FP32 from BF16 $dO$ and the saved BF16 output.

\paragraph{Score-gradient quantization.}
The next pass computes $dK,dV$ and stores quantized $dS$ for the final $dQ$ pass.
Each key block visits 64-query-row blocks in ascending order from the causal
diagonal to the sequence end. Its pre-cast intermediate is
\begin{equation}
\label{eq:app_kernel_scaled_ds}
U_{ij}=\Pi_{ij}\left[
A_{ij}\,\mathrm{fl}_{32}(s_{dO}s_Vs_K(j))
-\mathrm{fl}_{32}(\delta_i s_K(j))\right].
\end{equation}
Ideally, $U_{ij}=256s_K(j)P_{ij}(\widehat{dP}_{ij}-\delta_i)$.
We cast each tile using one scale $\psi$,
\begin{equation}
\label{eq:app_kernel_ds_scale}
C_{S,ij}=\mathcal{C}_8\!\left(U_{ij}\,\mathrm{fl}_{32}(1/\psi)\right),
\qquad
dS'_{ij}=\frac{\psi C_{S,ij}}{256s_K(j)}.
\end{equation}
The base scale is $\mathrm{fl}_{32}(\max|U|\,\mathrm{fl}_{32}(1/448))$.
It is floored at $10^{-30}$ and, for nonempty tiles, at
$g_{\max}256\,10^{-8}/s_Q$, where $g_{\max}$ is the largest preceding
$dK$ tile weight $\psi2^{-8}s_Q$ for that key block, initialized to zero.
Tiles whose base scale is below $10^{-30}$ store $\psi=0$ and zero codes.
For $dK$, compute each $C_S^\top Q_8$ tile in a fresh Tensor Core accumulator,
promote with weight $\psi2^{-8}s_Q$, update $g_{\max}$, and finally multiply
the summed gradient by $\mathrm{fl}_{32}(1/s_K)$ per key-scale group.

For $dV$, the operand $C_P^{\mathrm{bwd}}=\mathcal{C}_8(\Pi)$ has fixed
decoding factor $2^{-8}$. Accumulate $(C_P^{\mathrm{bwd}})^\top dO_8$
in a Tensor Core accumulator $W$ with basis $f=2^{-8}s_{dO}$.
Initialize $W=0$ and $f_{\mathrm{prev}}=1$. Before each query block, rescale
$W$ by $\mathrm{clip}(f_{\mathrm{prev}}/f,10^{-8},10^8)$ and update
$f_{\mathrm{prev}}=f$ before adding the code product.
Promote $Wf$ into the FP32 result at globally aligned 512-query-row
boundaries and the final block, then reset $W$.

\paragraph{Query gradients and storage.}
The final pass owns 64 query rows and visits key tiles in ascending order
up to the causal diagonal, using stored $C_S$ and $\psi$.
Initialize the accumulators, window basis $\psi_w$, and running maximum
$\psi_{\max}$ to zero. Skip tiles with $\psi=0$ or $\psi<10^{-8}\psi_{\max}$.
For accepted tiles, rescale the current Tensor Core accumulator by
$\mathrm{clip}(\psi_w/\psi,10^{-8},10^8)$, then add $C_SK_8$ and set
$\psi_w=\psi$ and $\psi_{\max}=\max(\psi_{\max},\psi)$.
The first accepted tile initializes the accumulator directly.
At globally aligned 512-key boundaries and the final tile, promote the
accumulator times $\psi_w$ into FP32 $dQ$ if any tile was accepted.
Reset the window accumulator and $\psi_w$, retaining $\psi_{\max}$.
Finally scale $dQ$ by $\mathrm{fl}_{32}(\tau/256)$.
Both stale and matched modes share these casts and gradient passes.
The implementation stores FP8 score-gradient tiles between passes, so it does
not eliminate quadratic intermediate storage.
All three gradients are returned in BF16.
For GQA, per-query-head $dK,dV$ are summed within each KV-head group in FP32
and rounded to BF16.

\paragraph{Gradient treatment and correction variants.}
Quantizers and key centering use straight-through gradients.
The returned $dQ,dK$ correspond to rotated queries and uncentered rotated keys;
RoPE is differentiated afterward where used.
In particular, the backward pass does not subtract the position-wise mean of $dK$
as differentiation through key centering would require.
Centering preserves softmax before quantization in exact arithmetic, but can
change the quantized forward.
Consistent $dO$ replaces raw $dO$ in the saved-output correction with its
dequantized FP8 operand.
The protected-$dP$ variant retains stale delta and computes only $dP$ from
BF16 operands with FP32 accumulation, leaving only the $s_K$ scale fold in
\eqref{eq:app_kernel_scaled_ds}; other attention products retain their FP8 paths.

Delta-Matching enforces the pre-cast invariant for normalized probabilities
in exact arithmetic. Approximate probability reconstruction, finite-precision
accumulation, and the subsequent $dS$ cast remain distinct numerical effects.
BF16 output-storage rounding additionally affects saved-output corrections,
but not the matched correction, which does not consume $O$.
Appendix~\ref{app:ds_quantization} measures the resulting
pre- and post-cast residuals.

\section{Additional Results}
\label{app:results}

We measure kernel-level row-sum residuals and attention throughput, then extend the main experiments with
FP8 linear layers, alternative stale-delta
mitigations, and studies of training dynamics across model architectures,
parameter scales, tokens per optimizer step, and optimizer choices. We also report general
capabilities after context extension and provide individual commonsense
benchmark accuracies and per-task RULER results.

\newenvironment{benchmarktable}{%
  \par\addvspace{\dimexpr\intextsep\relax}\noindent
  \begin{minipage}{\linewidth}
  \captionsetup{type=table}
}{%
  \end{minipage}\filbreak\addvspace{\dimexpr\intextsep\relax}
}

\subsection{Row-Sum Residuals Before and After FP8 Quantization}
\label{app:ds_quantization}

\paragraph{Setup.}
We probe the final BF16/FP32 and Delta-Matching checkpoints of the 1.67B
GDN/GQA model after 28,610 optimizer steps, using one shared initialization
and the training configurations in Appendix~\ref{app:train_config}.
The model has 16 query heads, four KV heads, and head dimension 128, with
RoPE applied to 32 channels.
We randomly sample 32 training windows of 8,192 tokens and capture post-RoPE
$Q,K$, $V$, and $dO$ at all six attention layers, indexed 3, 7, 11, 15, 19, and 23.
Each checkpoint is rerun in BF16 to obtain these activations and gradients;
the probe is not a replay of its FP8 training trajectory.
Our FP8 kernel then performs input quantization and the forward and backward passes.
The three corrections use identical inputs, with keys and values expanded
to all 16 query heads; KV-head diagnostics sum each group's four contributions.

\paragraph{Measurement and common row set.}
The instrumented gradient pass records per-row sums of FP32 $dS$ and $|dS|$,
together with the FP8 codes and scales consumed by $dQ$ and $dK$.
We decode the cast values as $dS'_{ij}=C_{S,ij}\psi/(2^8s_K(j))$
(Appendix~\ref{app:kernel_backward}).
After excluding query positions $i<64$, we retain rows whose pre-cast
denominator $\ell_i=\sum_j|dS_{ij}|$ is nonzero under all three corrections.
This common set covers about 99.8\% of candidate rows overall for either checkpoint.
Writing $\ell'_i=\sum_j|dS'_{ij}|$, we measure
\begin{equation}
\label{eq:app_ds_quantization_residuals}
r_i^{\mathrm{pre}}=\frac{\left|\sum_jdS_{ij}\right|}{\ell_i},
\qquad
r_i^{\mathrm{post}}=\frac{\left|\sum_jdS'_{ij}\right|}{\ell_i},
\qquad
r_i^{\mathrm{post,own}}=\frac{\left|\sum_jdS'_{ij}\right|}{\ell'_i}.
\end{equation}
Our primary post-cast statistic fixes each correction's own denominator
across the cast, not across corrections.
It retains rows whose cast gradient vanishes, assigning $r_i^{\mathrm{post}}=0$.
Only the own-denominator statistic excludes such rows.

We take row medians within each window and layer, averaging layers 3--19
within each window. Table~\ref{tab:app_ds_quantization} reports the mean
$\pm$ one SE over 32 windows, not over seeds or individual rows.
For across-layer ratios, we first average all six layer summaries within each window,
then take the geometric mean of paired window-level ratios.
The fixed- and own-denominator summaries agree within about 1.1\% in every reported cell.

\begin{table}[!htbp]
\centering
\caption{Normalized score-gradient row-sum residuals on the common row set.
The post-cast columns use the pre-cast denominator.
Values are mean $\pm$ one SE over 32 windows of per-layer row medians,
with layers 3--19 averaged within each window.
Checkpoint headings identify the activation source, not the probe correction.
All values are dimensionless fractions.}
\label{tab:app_ds_quantization}
\footnotesize
\setlength{\tabcolsep}{3pt}
\begin{tabular*}{\linewidth}{@{\extracolsep{\fill}}lcccc@{}}
\toprule
& \multicolumn{2}{c}{Before FP8 cast} & \multicolumn{2}{c}{After FP8 cast} \\
\cmidrule(lr){2-3}\cmidrule(lr){4-5}
Correction & Layers 3--19 & Layer 23 & Layers 3--19 & Layer 23 \\
\midrule
\multicolumn{5}{l}{\textit{BF16/FP32 checkpoint activations}} \\
Stale delta & $(1.1045\pmstd{0.0075})10^{-2}$ & $(4.660\pmstd{0.057})10^{-2}$ & $(1.1705\pmstd{0.0085})10^{-2}$ & $(4.679\pmstd{0.056})10^{-2}$ \\
Consistent $dO$ & $(2.466\pmstd{0.031})10^{-3}$ & $(6.103\pmstd{0.047})10^{-3}$ & $(3.898\pmstd{0.052})10^{-3}$ & $(8.62\pmstd{0.11})10^{-3}$ \\
Delta-Matching & $(1.425\pmstd{0.010})10^{-7}$ & $(7.102\pmstd{0.049})10^{-7}$ & $(2.700\pmstd{0.041})10^{-3}$ & $(3.394\pmstd{0.058})10^{-3}$ \\
\midrule
\multicolumn{5}{l}{\textit{Delta-Matching checkpoint activations}} \\
Stale delta & $(1.0818\pmstd{0.0083})10^{-2}$ & $(3.212\pmstd{0.046})10^{-2}$ & $(1.1507\pmstd{0.0091})10^{-2}$ & $(3.249\pmstd{0.046})10^{-2}$ \\
Consistent $dO$ & $(2.408\pmstd{0.033})10^{-3}$ & $(4.346\pmstd{0.056})10^{-3}$ & $(3.883\pmstd{0.054})10^{-3}$ & $(6.72\pmstd{0.11})10^{-3}$ \\
Delta-Matching & $(1.466\pmstd{0.013})10^{-7}$ & $(4.401\pmstd{0.069})10^{-7}$ & $(2.740\pmstd{0.042})10^{-3}$ & $(3.348\pmstd{0.059})10^{-3}$ \\
\bottomrule
\end{tabular*}
\end{table}

\paragraph{Correction-pass consistency.}
Delta-Matching forms its correction in a separate pass.
We rerun the gradient pass with corrections zero and one.
Writing its pre-cast row sum as $R_i(c)$, we infer
$S_i^{PdP}=R_i(0)$ and $S_i^P=R_i(0)-R_i(1)$, giving
$\delta_i^*=S_i^{PdP}/S_i^P$ under affine arithmetic.
We separately measure the normalized discrepancy where its denominator is nonzero,
\begin{equation}
\label{eq:app_ds_correction_discrepancy}
e_i=\frac{|\delta_i-\delta_i^*|\,|S_i^P|}{\sum_j|dS_{ij}|}.
\end{equation}
Delta-Matching's window-averaged median discrepancy is
$1.4\times10^{-7}$--$6.9\times10^{-7}$, with median inferred probability-normalization
errors around $3\times10^{-7}$.
This checks scalar centering at FP32-level numerical error, not elementwise
or bitwise agreement of the two passes' probabilities and $dP$.

\paragraph{Residuals after quantization.}
Delta-Matching's pre-cast residuals are consistent with FP32 rounding, but
the FP8 cast raises the reported medians to $2.7\times10^{-3}$--$3.4\times10^{-3}$.
On BF16/FP32 and Delta-Matching checkpoint activations, respectively,
stale delta remains $6.25\times$ and $5.29\times$ larger across layers
(log-ratio SE $0.011$ in both cases), and $13.82\times$ and $9.72\times$
larger in layer 23.
Consistent $dO$ remains $1.67\times$ and $1.53\times$ larger across layers,
with log-ratio SEs $0.004$ and $0.003$.

\paragraph{Key-gradient common mode.}
After summing each GQA group's four query-head contributions, we measure
\begin{equation}
\label{eq:app_ds_key_gradient_sum}
c_K=\frac{\left\|\sum_jdK_j\right\|_2}
{\left(\sum_j\|dK_j\|_2^2\right)^{1/2}}.
\end{equation}
The sums include all 8,192 keys, and the gradients include all query rows,
including those excluded from row statistics.
For KV group $g$, exact arithmetic before inverse RoPE gives
\begin{equation}
\label{eq:app_ds_gqa_key_sum}
\sum_jdK_{g,j}=\sum_{h\in g}\sum_i\bar Q_{h,i}\sum_jdS'_{h,ij},
\end{equation}
where $\bar Q=s_QQ_8$ includes the attention scale
(Appendix~\ref{app:kernel_operands}).
Measured kernel outputs also include matmul and storage rounding.
We take the median over four KV heads before aggregating layers and windows.
Stale delta's measured common mode is $4.33\times$ and $4.26\times$ larger
than Delta-Matching across layers in the two checkpoint states, and
$7.21\times$ and $7.14\times$ larger in layer 23.
After inverse RoPE, the 96 unrotated content channels retain this ordering,
with across-layer ratios of $4.43\times$ and $4.20\times$.
The 32 rotary channels instead give $c_K\approx0.34$--$0.43$ under every
correction. Position-dependent inverse rotations remove their cancellation
guarantee, so nonzero common mode there alone does not indicate stale-delta error.

\paragraph{Signed casting offset.}
We isolate the cast's signed contribution using
\begin{equation}
\label{eq:app_ds_signed_increment}
\kappa_i=\frac{\sum_jdS'_{ij}-\sum_jdS_{ij}}{\ell_i}.
\end{equation}
This statistic includes the entire common row set, even when the cast gradient
vanishes. We average over rows within each window, then report the mean and SE
across windows.
In layers 3--19, the absolute mean increment is at most $8.0\times10^{-6}$
across corrections and checkpoints.
In layer 23 of the Delta-Matching checkpoint, it is
$(1.08\pm0.14)\times10^{-3}$ for the matched correction and
$(3.96\pm0.34)\times10^{-3}$ for stale delta.
Thus the cast error is neither necessarily unbiased nor identical across
corrections. These signed row-sum offsets do not directly measure bias
in parameter gradients or its training impact.

\paragraph{Takeaway.}
Delta-Matching reduces pre-cast row-sum residuals to FP32-level numerical error
and retains lower residuals and KV-head common modes after FP8 quantization.
The cast still introduces finite residuals and a checkpoint-dependent signed
offset, so invariant preservation does not imply an unbiased quantized gradient.
These measurements cover two final checkpoints from one initialization and
one head-dimension-128 kernel, not residual trajectories throughout training.

\subsection{Attention Kernel Throughput}
\label{app:kernel_throughput}

\paragraph{Setup.}
We benchmark our FP8 attention implementation against FA3 BF16,
cuDNN BF16, and cuDNN FP8 through Transformer Engine on one H100 SXM 80GB GPU
in an otherwise idle eight-GPU node.
Each call uses two 8,192-token sequences with causal attention and native GQA,
without repeating key-value heads.
Table~\ref{tab:app_kernel_throughput} covers three attention shapes used by the
1.67B GDN/GQA base model, its head-dimension-256 variant, and the 2.58B pure-GQA model.
The cuDNN BF16 baseline uses PyTorch scaled-dot-product attention forced to the
cuDNN backend; the FP8 baseline uses Transformer Engine's delayed-scaling
\texttt{DotProductAttention}.

\paragraph{Timing and accounting.}
Inputs are random normal tensors.
We run five interleaved rounds, warming GPU clocks with three seconds of matrix
products before timing each benchmark.
Each round uses the median of 20 CUDA-event measurements after five warm-up calls.
Forward timing enables gradients and saves softmax statistics; backward time is
the forward-plus-backward time minus the training-forward time.
Our FP8 kernel timings exclude input-quantizer time because quantization can be
fused into adjacent operations; this benchmark uses prequantized inputs.
Timings also exclude autograd-wrapper overhead but include Delta-Matching's
separate row-correction pass.
The BF16 baselines include autograd overhead, and cuDNN/TE FP8 also includes input
casting, so these measurements are not like-for-like training-step comparisons.
For batch size $B$, query-head count $H_q$, sequence length $N$, and head dimension
$d$, nominal causal forward work is $2BH_qN^2d$ FLOPs and backward work is
$5BH_qN^2d$ FLOPs.
All methods receive the same FLOP count; the correction pass adds time but no
credited work. We report mean throughput $\pm$ one SE across the five rounds.

\begin{table}[!htbp]
\centering
\caption{\textbf{Attention kernel throughput.}
Values are TFLOP/s, reported as mean $\pm$ SE over five timing rounds.
Parentheses give ratios to FA3 BF16 in the same column and shape.
cuDNN/TE FP8 includes input casting and is unsupported at head dimension 256
in the measured training configuration.}
\label{tab:app_kernel_throughput}
\small
\setlength{\tabcolsep}{3pt}
\begin{tabular*}{\linewidth}{@{\extracolsep{\fill}}lccc@{}}
\toprule
Kernel & Forward & Backward & Forward + backward \\
\midrule
\multicolumn{4}{l}{\textit{Head dimension 128, 16 query heads, 4 key-value heads}} \\
FA3 BF16 & $386.9\pmstd{2.9}$ ($1.00\times$) & $230.6\pmstd{0.5}$ ($1.00\times$) & $260.7\pmstd{0.6}$ ($1.00\times$) \\
cuDNN BF16 & $601.6\pmstd{0.7}$ ($1.55\times$) & $472.9\pmstd{6.5}$ ($2.05\times$) & $503.6\pmstd{5.2}$ ($1.93\times$) \\
cuDNN/TE FP8 & $525.4\pmstd{1.4}$ ($1.36\times$) & $423.9\pmstd{1.2}$ ($1.84\times$) & $448.7\pmstd{1.1}$ ($1.72\times$) \\
Ours, stale delta & $661.0\pmstd{1.7}$ ($1.71\times$) & $537.5\pmstd{1.7}$ ($2.33\times$) & $567.8\pmstd{1.2}$ ($2.18\times$) \\
Ours, Delta-Matching & $661.0\pmstd{1.7}$ ($1.71\times$) & $421.5\pmstd{1.0}$ ($1.83\times$) & $470.2\pmstd{0.8}$ ($1.80\times$) \\
\midrule
\multicolumn{4}{l}{\textit{Head dimension 256, 8 query heads, 2 key-value heads}} \\
FA3 BF16 & $703.6\pmstd{5.5}$ ($1.00\times$) & $391.6\pmstd{5.1}$ ($1.00\times$) & $448.3\pmstd{4.7}$ ($1.00\times$) \\
cuDNN BF16 & $613.9\pmstd{1.6}$ ($0.87\times$) & $372.3\pmstd{5.6}$ ($0.95\times$) & $419.4\pmstd{5.1}$ ($0.94\times$) \\
cuDNN/TE FP8 & \multicolumn{3}{c}{Not supported} \\
Ours, stale delta & $696.6\pmstd{1.2}$ ($0.99\times$) & $518.2\pmstd{1.3}$ ($1.32\times$) & $559.1\pmstd{0.9}$ ($1.25\times$) \\
Ours, Delta-Matching & $696.6\pmstd{1.2}$ ($0.99\times$) & $426.4\pmstd{1.0}$ ($1.09\times$) & $479.5\pmstd{0.8}$ ($1.07\times$) \\
\midrule
\multicolumn{4}{l}{\textit{Head dimension 256, 16 query heads, 4 key-value heads}} \\
FA3 BF16 & $643.5\pmstd{11.1}$ ($1.00\times$) & $419.5\pmstd{2.7}$ ($1.00\times$) & $465.7\pmstd{1.4}$ ($1.00\times$) \\
cuDNN BF16 & $658.9\pmstd{0.5}$ ($1.02\times$) & $354.4\pmstd{1.4}$ ($0.84\times$) & $408.3\pmstd{1.3}$ ($0.88\times$) \\
cuDNN/TE FP8 & \multicolumn{3}{c}{Not supported} \\
Ours, stale delta & $686.0\pmstd{2.2}$ ($1.07\times$) & $516.9\pmstd{2.5}$ ($1.23\times$) & $556.1\pmstd{2.0}$ ($1.19\times$) \\
Ours, Delta-Matching & $686.0\pmstd{2.2}$ ($1.07\times$) & $426.0\pmstd{1.1}$ ($1.02\times$) & $477.7\pmstd{1.0}$ ($1.03\times$) \\
\bottomrule
\end{tabular*}
\end{table}

\paragraph{Takeaway.}
At head dimension 128, Delta-Matching reaches $1.05\times$ the
forward-plus-backward throughput of cuDNN/TE FP8.
At head dimension 256, it reaches $1.07\times$ and $1.03\times$ FA3 BF16
throughput with 8 and 16 query heads, respectively.
The corresponding stale-delta kernels reach $1.27\times$, $1.25\times$, and
$1.19\times$ these baselines.
Delta-Matching's correction pass adds roughly 21--28\% to stale-delta backward time.

\subsection{Delta-Matching with FP8 Linear Layers}
\label{app:fp8_linear}

These runs use 8 query heads and 2 key-value heads of dimension 256, whereas
the default configuration in Section~\ref{sec:exp_cmp} uses 16 query heads and 4 key-value heads of dimension 128 at the
same total query width. Dimension 128 supports the evaluated cuDNN FP8 configuration;
the vendor kernel is not evaluated here.
Absolute results are not directly comparable across these head dimensions.
Individual commonsense benchmark accuracies are reported in Table~\ref{tab:exp_cs9_breakdown}.

Using the precision scopes in Table~\ref{tab:train_precision}, the BF16 baseline
is P0 and reuses the Head Dim 256 reference; MP with FP8 FFNs is P1, stale and
matched FP8 attention with FP8 FFNs are P3, and Delta-Matching with FP8 FFNs and
projections is P4.

\begin{table}[!htbp]
\centering
\caption{
\textbf{Delta-Matching composes with FP8 linear layers.}
Mean $\pm$ standard error over two initialization seeds. Extraction scores are fractions;
other downstream scores are percentages.
}
\label{tab:exp_fp8_linear}
\small
\setlength{\tabcolsep}{4pt}
\resizebox{\textwidth}{!}{%
\begin{tabular}{lcccccc}
\toprule
Method & Validation CE $\downarrow$ & RULER-4K $\uparrow$ & RULER-8K $\uparrow$ & Extraction $\uparrow$ & TriviaQA $\uparrow$ & MMLU $\uparrow$ \\
\midrule
BF16/FP32 MP & $1.4135\pmstd{0.0003}$ & $62.94\pmstd{1.56}$ & $52.84\pmstd{2.14}$ & $0.6625\pmstd{0.0240}$ & $16.41\pmstd{0.21}$ & $35.97\pmstd{0.22}$ \\
BF16/FP32 MP + FP8 FFN & $1.4156\pmstd{0.0008}$ & $58.22\pmstd{1.64}$ & $49.21\pmstd{1.95}$ & $0.6751\pmstd{0.0024}$ & $16.01\pmstd{0.21}$ & $33.90\pmstd{0.11}$ \\
\midrule
FP8 Attn (Stale Delta) + FP8 FFN & $2.0429\pmstd{0.0166}$ & $18.85\pmstd{5.67}$ & $14.09\pmstd{5.86}$ & $0.3401\pmstd{0.0402}$ & $\phantom{0}1.54\pmstd{0.15}$ & $26.43\pmstd{0.41}$ \\
Delta-Matching + FP8 FFN & $1.4150\pmstd{0.0001}$ & $64.11\pmstd{3.22}$ & $51.48\pmstd{2.79}$ & $0.6881\pmstd{0.0107}$ & $15.45\pmstd{0.21}$ & $34.78\pmstd{0.13}$ \\
Delta-Matching + FP8 FFN + FP8 QKVO & $1.4164\pmstd{0.0001}$ & $59.54\pmstd{2.12}$ & $51.42\pmstd{1.73}$ & $0.6781\pmstd{0.0011}$ & $15.36\pmstd{0.23}$ & $35.60\pmstd{0.03}$ \\
\bottomrule
\end{tabular}%
}
\end{table}

\paragraph{Takeaway.}
Delta-Matching enables nearly full FP8 training, with FP8 operands for all FFN
and attention-projection GEMMs and all seven attention-core matmuls, while
matching the BF16/FP32 reference in validation loss and downstream performance.
Recurrent layers, the output head, softmax, reductions, and accumulation retain higher precision.
With the same FP8 FFNs, stale-delta attention instead substantially degrades
validation loss and downstream performance.

\subsection{Alternative Methods for Stale-Delta Mitigation}
\label{app:alternative_mitigations}
We evaluate alternative stale-delta corrections and probability encodings.
Each has limitations in training dynamics or final model quality in the tested configurations.

\subsubsection{Auxiliary BF16-Output Correction}
\label{app:exact_output_delta}

\paragraph{Derivation of the output error and remaining backward mismatch.}
We use the notation of Section~\ref{sec:violate_invariant}, with normalized $P$
before probability quantization and $\widehat{dP}=\widehat{dO}\widehat{V}^{\top}$.
We ignore backward matmul and row-reduction rounding and consider $dS$ before its
subsequent quantization. For any row correction $c_i$,
\begin{equation}
\label{eq:app_exact_output_row_sum}
\sum_j P_{ij}(\widehat{dP}_{ij}-c_i)
=\sum_j P_{ij}\widehat{dP}_{ij}-c_i\sum_jP_{ij}
=\delta_i^{\mathrm{matched}}-c_i.
\end{equation}
The correction must therefore equal the backward pass's attention-weighted mean,
regardless of output accuracy.

Let $O^{\mathrm{ref}}$ be an auxiliary BF16 attention output from the original
$Q,K,V$. It retains BF16 rounding and is not mathematically exact.
The probe uses $\delta_i^{\mathrm{ref}}=\langle dO_i,O_i^{\mathrm{ref}}\rangle$
for the backward correction only, leaving the FP8 forward output unchanged.
Write $O^{\mathrm{FP8}}$ for that saved output, equal to $\widehat{O}$ under
Section~\ref{sec:violate_invariant}'s ideal-product assumptions.
Inserting the reference correction into the stale residual gives
\begin{equation}
\label{eq:app_exact_output_decomposition}
\begin{aligned}
\varepsilon_i^{\mathrm{stale}}
&=\delta_i^{\mathrm{matched}}-\langle dO_i,O_i^{\mathrm{FP8}}\rangle\\
&=\bigl(\delta_i^{\mathrm{matched}}-\delta_i^{\mathrm{ref}}\bigr)
  +\langle dO_i,O_i^{\mathrm{ref}}-O_i^{\mathrm{FP8}}\rangle\\
&=\varepsilon_i^{\mathrm{ref}}
  +\langle dO_i,O_i^{\mathrm{ref}}-O_i^{\mathrm{FP8}}\rangle.
\end{aligned}
\end{equation}
Replacing the output removes the second term, but $\varepsilon_i^{\mathrm{ref}}$ need not vanish.

Define $O_i^b=\sum_jP_{ij}\widehat{V}_j$,
the output implied by the backward operands.
By bilinearity, $\delta_i^{\mathrm{matched}}=\langle\widehat{dO}_i,O_i^b\rangle$.
Adding and subtracting $\langle dO_i,O_i^b\rangle$ yields
\begin{equation}
\label{eq:app_exact_output_remaining}
\begin{aligned}
\varepsilon_i^{\mathrm{ref}}
&=\langle\widehat{dO}_i,O_i^b\rangle-\langle dO_i,O_i^{\mathrm{ref}}\rangle\\
&=\langle\widehat{dO}_i-dO_i,O_i^b\rangle
  +\langle dO_i,O_i^b-O_i^{\mathrm{ref}}\rangle.
\end{aligned}
\end{equation}
The gradient-quantization mismatch remains, and $O^{\mathrm{ref}}$ need not equal $O^b$.
Unlike consistent $dO$ in Section~\ref{sec:delta_matching}, this intervention
does not isolate the probability-cast error.
These identities apply at fixed operands; they do not imply additive effects on
training loss across trajectories.

\paragraph{Experimental setup.}
We compare BF16/FP32, stale delta, the auxiliary BF16-output correction, and
Delta-Matching using one shared initialization and data order.
The 1.665B GDN/GQA model has attention head dimension 256, with FP8 restricted to the
attention core and BF16 projections and FFNs (P2 versus P0 in
Table~\ref{tab:train_precision}).
Each run uses eight H100 GPUs and 128 packed Nemotron-CC v2.1 sequences of 8,192 tokens
per step, totaling approximately 14.7B tokens over 14,000 steps.
AdamW uses $\beta_1=0.9$, $\beta_2=0.95$, weight decay 0.1, and gradient clipping at 1.0.
The learning rate warms up for 466 steps to $2.4\times10^{-3}$, stays constant
through step 11,200, and decays linearly to zero at step 14,000.
The auxiliary BF16 forward is used only for delta; a fixed-input check confirmed
unchanged forward output and $dV$, while $dQ$ and $dK$ changed.
Stale and auxiliary-output runs use PyTorch RMSNorm and SwiGLU implementations,
whereas Delta-Matching and BF16/FP32 use Liger implementations of the same operations.
The effect of this implementation difference was not measured separately.

\begin{table}[!htbp]
\centering
\caption{Auxiliary BF16-output correction after 14,000 training steps.
Validation CE is in nats per token on 32 shared 8,192-token windows, evaluated
through the same BF16 attention path. Each method uses one initialization seed.
Only means were recorded, so no standard errors are reported.}
\label{tab:app_exact_output_delta}
\small
\setlength{\tabcolsep}{4pt}
\begin{tabular*}{\linewidth}{@{\extracolsep{\fill}}lccc@{}}
\toprule
Method & Correction & Validation CE $\downarrow$ & $\Delta$CE vs MP \\
\midrule
BF16/FP32 MP & Reference & $1.4779$ & $0$ \\
\midrule
Naive FP8 (Stale Delta) & $\langle dO_i,O_i^{\mathrm{FP8}}\rangle$ & $1.6346$ & $+0.1567$ \\
Auxiliary BF16-output delta & $\langle dO_i,O_i^{\mathrm{ref}}\rangle$ & $1.5755$ & $+0.0976$ \\
Delta-Matching & $\delta_i^{\mathrm{matched}}$ & $1.4781$ & $+0.0002$ \\
\bottomrule
\end{tabular*}
\end{table}

\paragraph{Results.}
The auxiliary output lowers final validation CE by 0.0591 nats relative to
stale delta, but its CE remains 0.0974 above Delta-Matching
(Table~\ref{tab:app_exact_output_delta}).
These endpoint differences do not isolate the residual terms' costs;
the second comparison also changes the RMSNorm/SwiGLU implementation.
Delta-Matching finishes 0.0002 nats above BF16/FP32.
No downstream benchmarks were evaluated for these runs.

The auxiliary-output run's training-loss gap to BF16/FP32 exceeds stale delta's in every
250-step bin starting from step 2,750 through 9,000, with a one-step spike
of 1.83 nats above BF16/FP32 at step 5,420.
Stale delta has the larger binned gap from step 10,250 onward.
Delta-Matching stays within 0.0015 nats of BF16/FP32 in every 250-step bin
from step 1,000 onward.
The auxiliary output thus improves final loss without consistently improving training stability.

A higher-precision auxiliary output improves final loss but does not guarantee
a correction consistent with the FP8 backward pass.
Delta-Matching directly enforces that consistency and matches BF16/FP32 training
and validation loss in this experiment without an additional forward pass.

\subsubsection{Protecting the Fragile Operation}
\label{app:protect_fragile_operation}

\paragraph{Setup.}
Following the SageBwd precision strategy in SageAttention3~\citep{zhang2026sageattention3},
we compute $dP=dO\,V^\top$ with BF16 operands while retaining FP8 for the other
six attention-core matmuls and using the saved-output stale delta.
This SageBwd-style implementation uses E4M3/BF16 rather than INT8/FP16.
We use the base 1.67B GDN/GQA configuration from Section~\ref{sec:exp_setup},
with head dimension 128, two initialization seeds, and a shared data order.
Each run trains for 28,610 steps on 30B Nemotron-CC tokens at 8K context
(Appendix~\ref{app:train_recipes}).
All methods use FP8 FFNs; the FP8 attention variants also use FP8 projections.
Relative to naive FP8, only the $dP$ precision path changes.

\begin{table}[!htbp]
\centering
\caption{Protecting $dP$ in the base 1.67B GDN/GQA configuration.
Results are mean $\pm$ one SE over two initialization seeds.
Validation CE is in nats per token; downstream scores are percentages.
The reference, naive FP8, and Delta-Matching rows reproduce results for the
default configuration in Table~\ref{tab:exp_cmp_overall}.}
\label{tab:app_sagebwd_eval}
\scriptsize
\setlength{\tabcolsep}{3pt}
\begin{tabular*}{\linewidth}{@{\extracolsep{\fill}}lccccccc@{}}
\toprule
& & \multicolumn{3}{c}{In-Context Recall $\uparrow$} & \multicolumn{3}{c}{General Capabilities $\uparrow$} \\
\cmidrule(lr){3-5}\cmidrule(lr){6-8}
Method & Validation CE $\downarrow$ & RULER-4K & RULER-8K & Extraction & TriviaQA & MMLU & CS9 \\
\midrule
BF16/FP32 MP & $1.4178\pmstd{0.0004}$ & $61.5\pmstd{3.0}$ & $52.3\pmstd{2.3}$ & $65.7\pmstd{2.0}$ & $15.5\pmstd{0.2}$ & $35.4\pmstd{1.4}$ & $58.8\pmstd{0.2}$ \\
\midrule
Naive FP8 (Stale Delta) & $1.8970\pmstd{0.0296}$ & $24.8\pmstd{0.3}$ & $19.8\pmstd{2.1}$ & $44.7\pmstd{2.8}$ & $\phantom{0}2.7\pmstd{0.1}$ & $26.3\pmstd{0.1}$ & $47.0\pmstd{0.9}$ \\
SageBwd-style & $1.4175\pmstd{0.0005}$ & $57.8\pmstd{3.3}$ & $46.0\pmstd{1.1}$ & $66.8\pmstd{1.1}$ & $15.8\pmstd{0.0}$ & $35.0\pmstd{0.9}$ & $59.1\pmstd{0.1}$ \\
Delta-Matching & $1.4162\pmstd{0.0006}$ & $61.4\pmstd{1.2}$ & $47.8\pmstd{0.5}$ & $68.0\pmstd{0.5}$ & $15.6\pmstd{0.2}$ & $34.0\pmstd{0.7}$ & $59.0\pmstd{0.2}$ \\
\bottomrule
\end{tabular*}
\end{table}

\paragraph{Stable loss with different gain dynamics.}
The SageBwd-style runs track BF16/FP32 training loss within 0.0035 nats in
every 2,000-step window and retain near-reference validation loss
(Table~\ref{tab:app_sagebwd_eval}).
Downstream performance is broadly comparable, although RULER-8K remains lower
at 46.0 versus 52.3, as it does for Delta-Matching at 47.8.
Protecting $dP$ avoids FP8 rounding of $dO$ in that product, but the saved-output
correction still does not enforce the zero-row-sum invariant.

The gain trajectories nevertheless differ substantially.
In layer 19 of the first seed and layer 15 of the second, 64 and 29 of the
128 key gains exceed 4, respectively, with the largest reaching 27.1.
Corresponding query gains are smaller than in BF16/FP32, whose gains remain balanced.
This imbalance does not imply attention collapse.
The affected layers retain near-reference entropy, and removing their
attention outputs costs 0.90 and 1.01 times the reference's loss increase
(Table~\ref{tab:app_sagebwd_dynamics}).
For nonrotary channels, the exact-arithmetic logits depend on $g_{q,c}g_{k,c}$,
so key-gain growth need not sharpen attention when query gains decrease.

\begin{table}[!htbp]
\centering
\caption{Gain imbalance without attention collapse in the SageBwd-style runs.
Counts cover all 128 channels. Entropy is averaged over 16 held-out windows
and query positions 1,024 through 8,191.
$\Delta\mathrm{CE}_{\mathrm{mask}}$ is the loss increase from zeroing the layer's
attention output relative to each model's unmasked loss, with paired SE over
500 windows. Entropy is in nats and CE in nats per token.
Layer indices are zero-based.}
\label{tab:app_sagebwd_dynamics}
\small
\setlength{\tabcolsep}{3pt}
\begin{tabular*}{\linewidth}{@{\extracolsep{\fill}}cllccc@{}}
\toprule
Seed & Layer & Method & $\#(g_{k,c}>4)$ & Entropy & $\Delta\mathrm{CE}_{\mathrm{mask}}$ \\
\midrule
First & 19 & BF16/FP32 MP & 1 & $4.08$ & $0.1101\pmstd{0.0022}$ \\
 & & SageBwd-style & 64 & $4.21$ & $0.0987\pmstd{0.0021}$ \\
\midrule
Second & 15 & BF16/FP32 MP & 0 & $4.71$ & $0.0647\pmstd{0.0013}$ \\
 & & SageBwd-style & 29 & $4.67$ & $0.0652\pmstd{0.0014}$ \\
\bottomrule
\end{tabular*}
\end{table}

\paragraph{Interaction with key centering.}
The kernel subtracts the sequence-mean key $\bar K$ before quantization and
uses centered keys in the query-gradient product.
Holding operands fixed and ignoring additional rounding, with
$\tau=1/\sqrt{d}$ and $\varepsilon_i=\sum_jdS_{ij}$,
\begin{equation}
\label{eq:app_sagebwd_key_centering}
dQ_i^{\mathrm{centered}}
=\tau\sum_jdS_{ij}(K_j-\bar K)
=dQ_i^{\mathrm{uncentered}}-\tau\varepsilon_i\bar K.
\end{equation}
A nonzero row sum therefore removes a term from the query gradient, while
the key gradient retains its corresponding summed component
$\sum_jdK_j=\tau\sum_i\varepsilon_iQ_i$.
This asymmetry can contribute to the observed gain imbalance.
In an 8,000-step diagnostic run, a proxy for this imbalance correlates with
layer-19 log key gains at Spearman $-0.68$.
The proxy uses recorded summed key gradients and fixed checkpoint key offsets,
so it supports the mechanism without measuring the exact update at each step.
Delta-Matching removes the stale residual before $dS$ quantization, although
subsequent rounding can leave a smaller residual.
It has four layer-19 key gains above 4 in each seed, while all seven attention
matmuls remain in FP8.

Protecting a single attention matmul with BF16 enables stable training in the
tested setup, but leaves markedly different training dynamics.
Despite near-reference loss, selected layers develop much larger key gains
and smaller query gains than BF16/FP32.

\subsubsection{FA3-style Probability Encoding and Stale Delta}
\label{app:fa3_max_shift}

\paragraph{Setup and probability encoding.}
We compare FA3-style max-shift and Delta-Matching on the 1.67B GDN/GQA hybrid
(head dimension 256), trained on 30B Nemotron-CC tokens with eight H100 GPUs
(Appendix~\ref{app:train_recipes}).
Only the attention core uses FP8 (P2 versus P0 in Table~\ref{tab:train_precision});
projections and FFNs remain BF16. All evaluations use the same BF16 attention path.
The BF16/FP32 reference reuses the runs in the Head Dim 256 BF16/FP32
row of Table~\ref{tab:exp_attn_arch}, whereas the Delta-Matching results come from
separate runs using an earlier version of the Delta-Matching kernels with the
same model geometry and precision scope.

FlashAttention-3 (FA3) scales the exponentials relative to the running row maximum
by $2^8$ before casting to E4M3, reducing underflow~\citep{shah2024flashattention3}.
FA3-MS adapts this encoding to block-scaled values and FP8 backward;
it is not a stock FA3 training kernel. Folding value scales into probability codes
can round even the row-maximum code; our kernel maps each 32-key group's maximum
to the exact E4M3 code 448.
FA3-MS's externally matched correction reuses its backward operand codes,
but separate accumulation can leave a mismatch. Our Delta-Matching kernel
uses promoted accumulation in both passes and forms its correction from products
used inside the backward pass.

\begin{table}[!htbp]
\centering
\caption{FA3-style max-shift versus Delta-Matching at head dimension 256.
Validation CE is in nats per token; all downstream scores are percentages.
CS9 is the unweighted mean accuracy over nine individual commonsense benchmarks.
MP and Delta-Matching use the same two initialization seeds and report mean
$\pm$ one SE. Each FA3-MS row uses one of those seeds.
N/E denotes not evaluated.}
\label{tab:app_fa3_encoding}
\scriptsize
\setlength{\tabcolsep}{1pt}
\begin{tabular*}{\linewidth}{@{\extracolsep{\fill}}lcccccccc@{}}
\toprule
& & & \multicolumn{3}{c}{In-Context Recall $\uparrow$} & \multicolumn{3}{c}{General Capabilities $\uparrow$} \\
\cmidrule(lr){4-6}\cmidrule(lr){7-9}
Method & $n$ & Validation CE $\downarrow$ & RULER-4K & RULER-8K & Extraction & TriviaQA & MMLU & CS9 \\
\midrule
BF16/FP32 MP & 2 & $1.41354\pmstd{0.00026}$ & $62.94\pmstd{1.56}$ & $52.84\pmstd{2.14}$ & $66.25\pmstd{2.40}$ & $16.41\pmstd{0.21}$ & $35.97\pmstd{0.22}$ & $59.41\pmstd{0.81}$ \\
\midrule
FA3-MS (stale) & 1 & $1.4445$ & $58.97$ & $51.51$ & $67.91$ & $13.71$ & $32.15$ & $57.21$ \\
FA3-MS (externally matched) & 1 & $1.4349$ & $58.49$ & $45.05$ & $67.06$ & $15.09$ & \textit{N/E} & $58.10$ \\
Delta-Matching (\textit{ours}) & 2 & $1.41409\pmstd{0.00019}$ & $63.20\pmstd{0.86}$ & $52.28\pmstd{1.17}$ & $68.33\pmstd{0.92}$ & $16.30\pmstd{0.16}$ & $33.68\pmstd{0.19}$ & $59.35\pmstd{0.07}$ \\
\bottomrule
\end{tabular*}
\end{table}

\paragraph{Results.}
FA3-MS with stale delta completed training without non-finite or skipped updates,
but validation CE was $0.0307\pm0.0002$ above the same-initialization
BF16/FP32 reference. External matching reduced this gap to $0.0211\pm0.0002$.
Delta-Matching's two paired gaps were $0.0001\pm0.0002$ and $0.0010\pm0.0002$.
These uncertainties are SEs over 2,000 validation windows, not across seeds.
The FA3-MS variants share the RMSNorm and SwiGLU operations but differ in kernel
implementations, whose effect was not isolated.
Their single-initialization results do not establish a general downstream effect.

FA3-style max-shift avoided collapse in these runs but left a validation CE gap
with stale or externally matched backward corrections. Delta-Matching with naive forward
encoding stayed closer to the BF16/FP32 reference.

\paragraph{Takeaways.}
\begin{itemize}
\item An auxiliary BF16 output lowers final validation loss but does not resolve
the backward correction mismatch (Appendix~\ref{app:exact_output_delta}).
\item Keeping one attention matmul in BF16 enables stable training but leaves
markedly different gain dynamics in the tested setup
(Appendix~\ref{app:protect_fragile_operation}).
\item FA3-style max-shift FP8 forward leaves a validation-loss gap to BF16/FP32
with both stale-delta and externally matched backward corrections in the tested
runs (Appendix~\ref{app:fa3_max_shift}).
\end{itemize}

\subsection{Architectures and Scales Where Stale-Delta Training Still Converges}
\label{app:convergent_architectures}

Convergence can conceal stale-delta corruption.
We inspect QK-normalization gains and mask the last attention layer's output.
The 1.475B pure-GQA model retains near-reference gains, whereas the 2.58B pure-GQA
and Mamba-2/GQA models have abnormal gains despite small loss gaps.
Masking suggests compensation for the affected layers.

\begin{figure}[!htbp]
\centering
\includegraphics[width=0.50\textwidth]{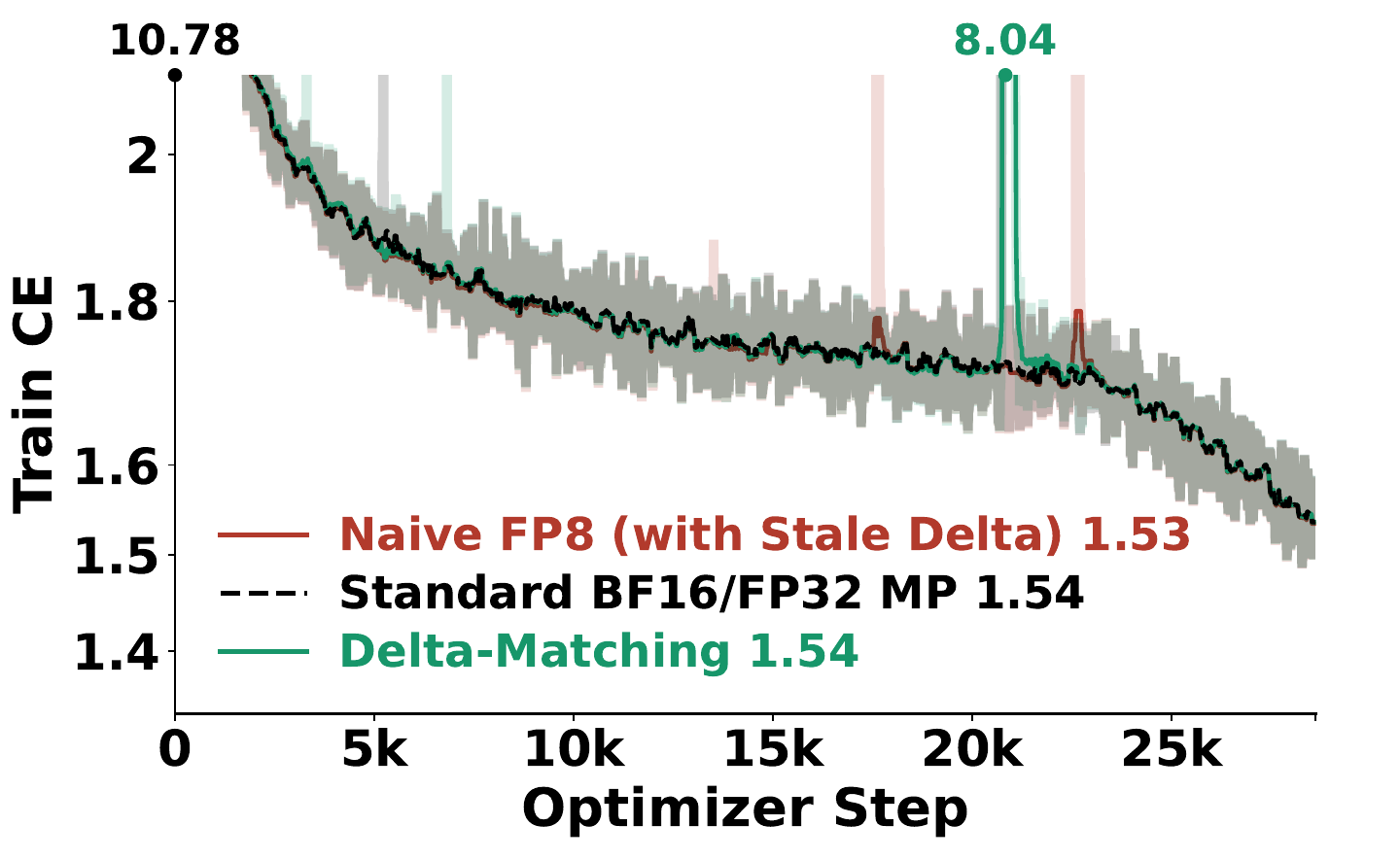}
\caption{Training cross-entropy for a 1.475B pure GQA LLM, comparing
BF16/FP32 mixed precision, naive FP8 with stale delta, and Delta-Matching.}
\label{fig:app_pure_gqa_loss}
\label{fig:app_convergent_architectures}
\end{figure}

\begin{table}[!htbp]
\centering
\caption{Last-attention-layer probes after 30B-token training, using one
initialization seed per method and configuration. Masking zeros the attention output while retaining
the residual and FFN. $\Delta\mathrm{CE}_{\mathrm{mask}}$ is the resulting CE increase,
with paired SEs over 2,000 shared 8,192-token windows evaluated in BF16.
SEs are rounded to four decimal places.
The maximum gain product is $\max_c |g_{q,c}g_{k,c}|$; the top-six share is the
percentage of $\sum_c(g_{q,c}g_{k,c})^2$ in the six largest channels.
The 1.475B pure-GQA rows use the same configuration as
Table~\ref{tab:app_pure_gqa_eval}.}
\label{tab:app_arch_layer_probe}
\small
\setlength{\tabcolsep}{3pt}
\begin{tabular*}{\linewidth}{@{\extracolsep{\fill}}lcccc@{}}
\toprule
Method & Validation CE & Max.\ gain product & Top-six share (\%) & $\Delta\mathrm{CE}_{\mathrm{mask}}$ \\
\midrule
\multicolumn{5}{l}{\textbf{1.67B GDN/GQA}} \\
BF16/FP32 MP & $1.4174$ & $3.1$ & $6.9$ & $0.0585\pmstd{0.0003}$ \\
Naive FP8 (Stale Delta) & $1.8674$ & $2244.4$ & $99.9$ & $0.1269\pmstd{0.0010}$ \\
Delta-Matching & $1.4169$ & $3.7$ & $7.8$ & $0.0151\pmstd{0.0001}$ \\
\midrule
\multicolumn{5}{l}{\textbf{1.67B Mamba-2/GQA}} \\
BF16/FP32 MP & $1.4144$ & $4.4$ & $9.6$ & $0.0288\pmstd{0.0002}$ \\
Naive FP8 (Stale Delta) & $1.4215$ & $404.1$ & $87.3$ & $0.0119\pmstd{0.0001}$ \\
Delta-Matching & $1.4147$ & $4.2$ & $9.0$ & $0.0165\pmstd{0.0002}$ \\
\midrule
\multicolumn{5}{l}{\textbf{1.475B pure GQA}} \\
BF16/FP32 MP & $1.5242$ & $4.1$ & $9.7$ & $0.2585\pmstd{0.0008}$ \\
Naive FP8 (Stale Delta) & $1.5263$ & $4.0$ & $9.4$ & $0.2784\pmstd{0.0011}$ \\
Delta-Matching & $1.5254$ & $3.9$ & $9.1$ & $0.2180\pmstd{0.0008}$ \\
\midrule
\multicolumn{5}{l}{\textbf{2.58B pure GQA}} \\
BF16/FP32 MP & $1.4388$ & $2.2$ & $5.3$ & $0.0302\pmstd{0.0002}$ \\
Naive FP8 (Stale Delta) & $1.4452$ & $878.9$ & $89.1$ & $0.0008\pmstd{0.0000}$ \\
Delta-Matching & $1.4376$ & $2.5$ & $4.9$ & $0.0095\pmstd{0.0001}$ \\
\bottomrule
\end{tabular*}
\end{table}

\subsubsection{Pure Grouped Query Attention LLM}
\label{app:pure_gqa_layer_probe}

We probe one of the two initializations of the 1.475B pure-GQA model shown in
Figure~\ref{fig:app_pure_gqa_loss} and Table~\ref{tab:app_pure_gqa_eval}.
Its 28 attention layers use 16 query and 8 KV heads of dimension 128,
full-head RoPE with $\theta=10^6$, and no attention output gate.
Only the attention core uses FP8 through the Triton implementation;
projections and FFNs remain BF16, corresponding to P2 against the P0 reference
in Table~\ref{tab:train_precision}.
Stale delta raises validation CE by $0.0020\pm0.0003$ nats relative to MP,
but the final layer's maximum gain product remains near the reference
(4.0 versus 4.1). Masking this layer costs 0.2784 nats, compared with 0.2585
for MP, so the model still depends substantially on its final attention layer.

We also evaluate one seed per method at 2.58B parameters under the same
30B-token training recipe (Table~\ref{tab:app_arch_layer_probe}).
This model retains 28 attention layers, full-head RoPE, and no output gate,
but uses hidden width 2560, FFN width 8192, and 16 query and 4 KV heads of dimension 256.
Its FP8 methods use our FP8 attention kernels with P4 precision, including FP8 FFNs and attention
projections, against a P1 MP reference.
The larger stale-delta model's maximum final-layer gain product reaches 878.9, versus
2.2 for MP and 2.5 for Delta-Matching, with 89.1\% of squared gain products
concentrated in six channels.
Pure GQA thus exhibits gain corruption at the larger scale, although head geometry,
precision scope, and implementation also change.

Despite this corruption, the larger stale model's paired validation CE gap
is only $0.0064\pm0.0003$ nats.
Masking its last attention output costs just 0.0008 nats, versus 0.0302 for MP
and 0.0095 for Delta-Matching. The stale model's penalty is about 3\% of MP's.
With that output masked in both models, stale-delta CE is 1.4460, below MP's 1.4690.
Near-reference performance despite little dependence on the affected layer is
consistent with compensation by the remaining 27 attention layers rather than
gain repair, but the probe does not identify which components compensate.

\begin{table}[!htbp]
\centering
\caption{
\textbf{Evaluation of the 1.475B pure grouped-query attention LLM.}
Mean $\pm$ standard error over two initialization seeds. All scores except
validation cross-entropy are percentages.
}
\label{tab:app_pure_gqa_eval}
\small
\setlength{\tabcolsep}{3pt}
\resizebox{\textwidth}{!}{%
\begin{tabular}{lccccccc}
\toprule
Method & Validation CE $\downarrow$ & RULER-4K $\uparrow$ & RULER-8K $\uparrow$ & Extraction $\uparrow$ & TriviaQA $\uparrow$ & MMLU $\uparrow$ & CS9 $\uparrow$ \\
\midrule
BF16/FP32 MP & $1.5283\pmstd{0.0041}$ & $49.0\pmstd{0.6}$ & $37.2\pmstd{3.1}$ & $48.8\pmstd{0.3}$ & $10.7\pmstd{0.0}$ & $26.1\pmstd{0.1}$ & $54.6\pmstd{0.3}$ \\
\midrule
Naive FP8 (Stale Delta) & $1.5249\pmstd{0.0014}$ & $52.1\pmstd{2.7}$ & $37.9\pmstd{0.8}$ & $48.2\pmstd{0.2}$ & $11.1\pmstd{0.0}$ & $26.3\pmstd{0.4}$ & $54.6\pmstd{0.2}$ \\
Delta-Matching (\textit{ours}) & $1.5283\pmstd{0.0029}$ & $45.4\pmstd{3.1}$ & $36.0\pmstd{1.2}$ & $49.4\pmstd{1.6}$ & $10.7\pmstd{0.1}$ & $27.4\pmstd{1.4}$ & $54.8\pmstd{0.2}$ \\
\bottomrule
\end{tabular}%
}
\end{table}

\subsubsection{Mamba-2 and Grouped Query Attention Hybrid LLM}
\label{app:mamba_layer_probe}

The Mamba-2/GQA hybrid in Section~\ref{sec:exp_model_arch} retains near-reference
validation loss and downstream performance under stale-delta training
(Figure~\ref{fig:app_mamba_gqa_loss} and Table~\ref{tab:exp_model_arch}).
It shares the GDN/GQA hybrid's six-attention-layer layout, head geometry,
and training recipe, but uses 18 Mamba-2 layers and a wider FFN.
Table~\ref{tab:app_arch_layer_probe} probes one final checkpoint per method,
separately from the two-seed benchmark averages.

Stale delta still inflates the last layer's gain products, with a maximum of
404.1 and a top-six share of 87.3\%, versus 4.4 and 9.6\% for MP.
This maximum is below the GDN/GQA stale model's 2244.4 but far above
Delta-Matching's 4.2. Yet the paired validation CE gap to MP is only
$0.0070\pm0.0002$ nats, compared with $0.4500\pm0.0016$ for GDN/GQA.

Masking the affected Mamba-2/GQA layer costs 0.0119 nats, about 41\% of MP's
0.0288-nat masking penalty. The layer still contributes, but less than in the
GDN/GQA stale model, whose masking penalty is
0.1269 nats, more than twice its MP reference's 0.0585.
With the last attention output masked, Mamba-2/GQA stale CE is 1.4334 versus
1.4432 for MP. These results are consistent with milder disruption and
compensation elsewhere, despite having the same number of attention layers as GDN/GQA.

\paragraph{Takeaway.}
Near-reference loss conceals severe gain corruption in the 2.58B pure-GQA and
Mamba-2/GQA models, unlike the smaller pure-GQA configuration.
Masking suggests compensation elsewhere in the network, with attention-layer
redundancy a plausible explanation for pure GQA.
Delta-Matching maintains near-reference performance without the observed gain corruption.

\subsection{Additional Study on Training Dynamics}
\label{app:token_per_step}
We test whether a larger global batch size reduces stale-delta degradation at a fixed token budget.

\paragraph{Setup.}
We train the 1.67B GDN/GQA hybrid with head dimension 256 on 30B Nemotron-CC
tokens using 8,192-token windows and eight H100 GPUs.
Only the attention core uses FP8 through the Triton implementation;
projections and FFNs remain BF16 (P2 versus P0 in Table~\ref{tab:train_precision}).
We compare batches of 128 and 256 sequences, giving 1.05M and 2.10M tokens
per step and 28,610 and 14,305 updates, respectively.
Both use the recipe in Appendix~\ref{app:train_recipes} and a peak learning rate
of $2.4\times10^{-3}$, with warmup and decay occupying the same fractions of training.
At each batch size, stale-delta and Delta-Matching share the implementation,
initialization, and data order, differing only in the backward correction.

\begin{table}[!htbp]
\centering
\caption{Effect of tokens per optimizer step at a fixed 30B-token budget.
The 1.05M-token rows report mean $\pm$ one SE across three initialization seeds.
The 2.10M-token rows use one of those seeds and have no across-seed error bars.
Validation CE is in nats per token; other scores are percentages.}
\label{tab:app_token_per_step}
\small
\setlength{\tabcolsep}{1.5pt}
\begin{tabular*}{\linewidth}{@{\extracolsep{\fill}}lcccccc@{}}
\toprule
Method & Validation CE $\downarrow$ & RULER-4K $\uparrow$ & RULER-8K $\uparrow$ & Extraction $\uparrow$ & MMLU $\uparrow$ & CS9 $\uparrow$ \\
\midrule
\multicolumn{7}{l}{1.05M tokens per step, 28,610 updates} \\
BF16/FP32 MP & $1.41387\pmstd{0.00036}$ & $62.38\pmstd{1.06}$ & $52.16\pmstd{1.41}$ & $67.21\pmstd{1.69}$ & $35.86\pmstd{0.17}$ & $59.16\pmstd{0.53}$ \\
Naive FP8 (Stale Delta) & $2.04837\pmstd{0.02806}$ & $13.47\pmstd{2.02}$ & $10.12\pmstd{2.45}$ & $31.52\pmstd{2.72}$ & $26.04\pmstd{0.14}$ & $44.07\pmstd{0.22}$ \\
Delta-Matching & $1.41498\pmstd{0.00088}$ & $63.80\pmstd{1.69}$ & $55.14\pmstd{2.10}$ & $67.59\pmstd{0.86}$ & $34.69\pmstd{1.00}$ & $58.91\pmstd{0.12}$ \\
\midrule
\multicolumn{7}{l}{2.10M tokens per step, 14,305 updates} \\
Naive FP8 (Stale Delta) & $1.6197$ & $46.64$ & $36.80$ & $59.45$ & $26.93$ & $52.69$ \\
Delta-Matching & $1.4119$ & $57.25$ & $48.54$ & $68.29$ & $33.38$ & $59.18$ \\
\bottomrule
\end{tabular*}
\end{table}

\paragraph{A larger batch reduces the final loss gap.}
For the initialization shared across batch sizes, doubling the batch reduces
the final validation CE gap between stale delta and Delta-Matching from
$0.6866\pm0.0022$ to $0.2078\pm0.0009$ nats.
Delta-Matching's own validation CE changes by only $-0.0014\pm0.0002$ nats.
These paired uncertainties are SEs over 2,000 held-out windows, not across seeds.
Despite the smaller gap, the larger-batch stale run still trails Delta-Matching
on every downstream metric in Table~\ref{tab:app_token_per_step}.

\paragraph{Earlier degradation but a smaller final gap.}
For the shared initialization, the larger-batch run deteriorates after fewer
updates (7,590 versus 8,820), but has half as many total updates and starts
learning-rate decay earlier in update count.
Its smaller final gap therefore need not imply less error per update.

At matched updates 9,000--9,500, the gaps are similar, averaging 0.029 with the standard batch and
0.034 with the doubled batch, before either learning-rate schedule decays.

\paragraph{Takeaway.} Increasing tokens per step reduces the observed stale-delta penalty at a fixed
token budget but does not eliminate it.
The results are consistent with error accumulating over optimizer updates,
while Delta-Matching maintains similar validation loss and downstream task performance at both batch sizes.

\subsection{Delta-Matching with Muon-Based Training}
\label{app:muon}

We test whether the stale-delta dynamics and the benefit of Delta-Matching
extend beyond AdamW by training with Muon~\citep{jordan2024muon,liu2025muonscalable}.

\paragraph{Setup.}
We use the 1.665B GDN/GQA hybrid with 18 GatedDeltaNet and 6 GQA layers,
16 query heads, 4 KV heads, head dimension 128, QK normalization, and RoPE
on 32 channels per head.
Each run uses eight H100 GPUs and a batch of 128 sequences of 8,192 tokens
from Nemotron-CC.
The BF16/FP32 reference uses FP8 FFNs and BF16 attention and projections (P1).
Naive FP8 and Delta-Matching use FP8 attention, QKVO projections, and FFNs (P4),
differing only in the backward correction.
All three share one initialization and data order with their stage-1 AdamW counterparts.

\paragraph{Optimizer and schedule.}
Muon updates the 222 two-dimensional hidden weight matrices in attention,
GatedDeltaNet projections, and FFNs.
It uses Nesterov momentum of 0.95, five Newton--Schulz iterations, and
decoupled weight decay of 0.1.
For an $m\times n$ matrix, the orthogonalized update is scaled by
$0.2\sqrt{\max(m,n)}$, following the Moonlight recipe~\citep{liu2025muonscalable}
to reuse the AdamW learning rate.
AdamW with $(\beta_1,\beta_2)=(0.9,0.95)$ updates the remaining parameters,
including tied embeddings, normalization and QK gains, GDN convolutions,
and state-space scalars.
We retain the 28,610-step warmup-stable-decay schedule, with 952 warmup steps,
peak learning rate $2.4\times10^{-3}$, and gradient-norm clipping at 1.0.

\subsubsection{Mechanism Probe}
\label{app:muon_mechanism}

We first inspect the initial 4,500 updates, approximately 4.7B tokens,
while the learning rate is still at its peak.
The AdamW curves cover the same initial interval with the same model,
initialization, data order, and learning-rate schedule.

\paragraph{Measurements.}
The carrier gain is $G_{23}=\frac{1}{3}\sum_{c\in\{7,8,24\}}g_{q,c}g_{k,c}$,
averaged over three channels in the last attention layer.
Its query and key RMSNorm gain vectors each have 128 entries, are shared
across heads, and are initialized to one.
The selected channels belong to the slow rotary pairs $(7,23)$ and $(8,24)$.
Gains are logged every 25 updates and plotted without smoothing.
The AdamW gain traces come from instrumented reruns with the same
initialization, data order, and schedule as the corresponding stage-1 runs.
Training cross-entropy is logged every 10 updates, with a rolling median
over 51 logged points used for the plotted loss traces.

\begin{figure}[!htbp]
\centering
\includegraphics[width=\linewidth]{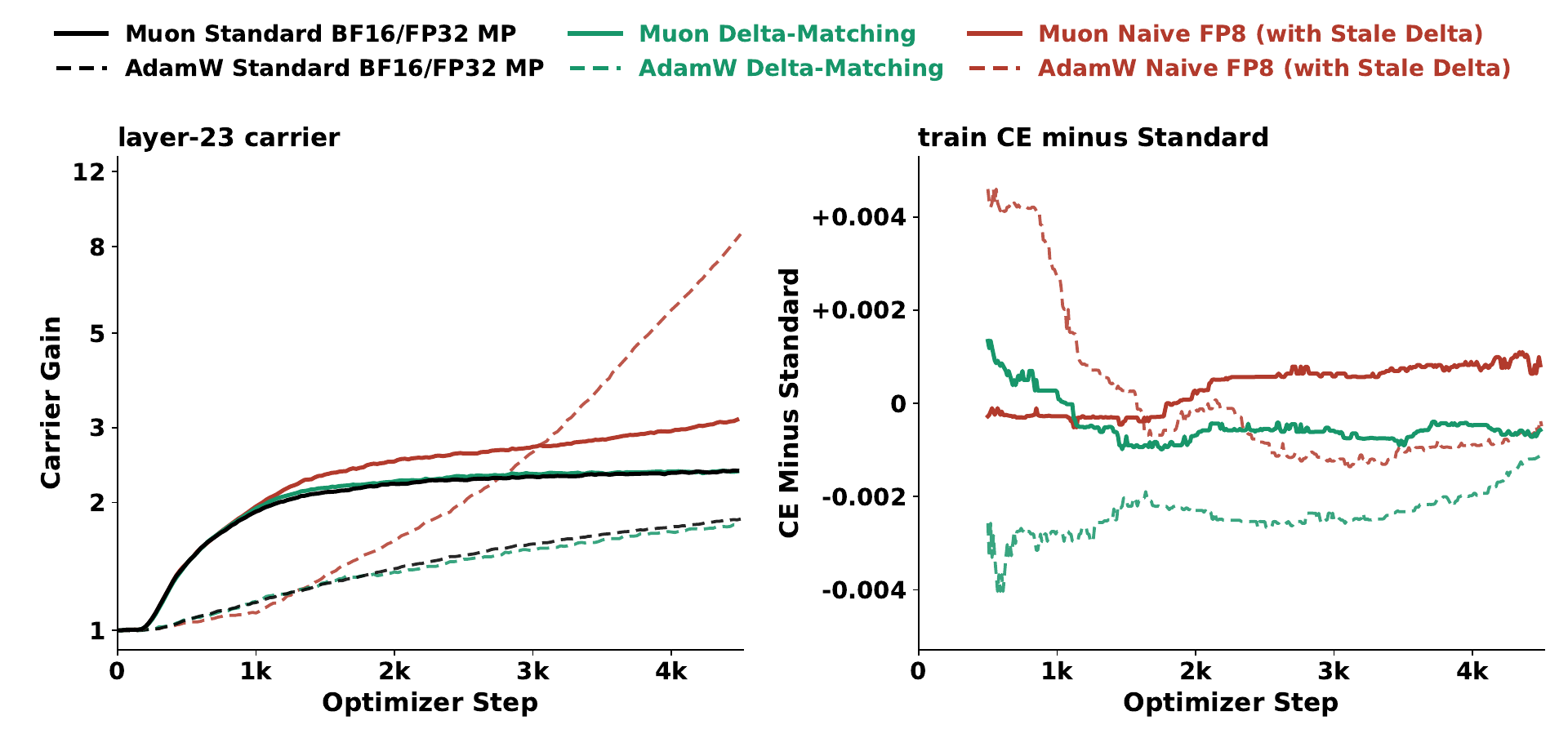}
\caption{Training dynamics with Muon-based updates (solid) and AdamW (dashed)
over the first 4,500 optimizer steps.
The left panel shows the layer-23 mean query--key gain product $G_{23}$ on a logarithmic axis;
the right shows training cross-entropy relative to each optimizer's
BF16/FP32 reference on a linear axis.
Stale delta produces excess gain growth under both optimizer recipes,
while Delta-Matching closely tracks the corresponding reference.}
\label{fig:app_muon}
\end{figure}

\paragraph{Results.}
Figure~\ref{fig:app_muon} shows that stale-delta gain growth persists under
Muon-based training, although it is less pronounced than under AdamW
during this interval.
The Muon stale-delta run also develops a positive training-loss gap,
whereas Delta-Matching closely follows the Muon BF16/FP32 gain trajectory
without the rising loss gap.
The AdamW stale-delta run exhibits stronger gain distortion even though its
loss gap is not consistently positive, illustrating why short loss traces
alone can miss the altered dynamics.

\subsubsection{Full Training and Downstream Evaluation}
\label{app:muon_full_training}

We train the three Muon methods for the full 28,610 updates, approximately
30B tokens, and evaluate their final checkpoints with the same protocol as
Table~\ref{tab:exp_cmp_overall}.
Tables~\ref{tab:app_muon_full} and~\ref{tab:app_muon_cs9} compare them with
the corresponding individual AdamW runs sharing the same initialization
and batches.
All rows use one shared initialization; the AdamW values are individual-run
results rather than the two-seed means for the default configuration in
Table~\ref{tab:exp_cmp_overall}.
Per-task RULER scores are reported in Table~\ref{tab:app_ruler_muon}.

\begin{table}[!htbp]
\centering
\caption{
\textbf{Validation loss and downstream performance across optimizers.}
Results after 28,610 updates with one shared initialization and data order.
Validation CE is in nats per token; downstream scores are percentages.
CS9 is the unweighted mean of the nine benchmark accuracies in
Table~\ref{tab:app_muon_cs9}. Single-initialization results carry no standard errors.
}
\label{tab:app_muon_full}
\scriptsize
\setlength{\tabcolsep}{1.5pt}
\begin{tabular*}{\linewidth}{@{\extracolsep{\fill}}llccccccc@{}}
\toprule
& & & \multicolumn{3}{c}{In-Context Recall $\uparrow$} & \multicolumn{3}{c}{General Capabilities $\uparrow$} \\
\cmidrule(lr){4-6}\cmidrule(lr){7-9}
Optimizer & Method & Validation CE $\downarrow$ & RULER-4K & RULER-8K & Extraction & TriviaQA & MMLU & CS9 \\
\midrule
Muon & BF16/FP32 MP & $1.3988$ & $62.1$ & $53.0$ & $68.2$ & $16.5$ & $36.2$ & $59.2$ \\
& Naive FP8 (Stale Delta) & $1.8077$ & $32.5$ & $26.2$ & $47.5$ & $\phantom{0}3.3$ & $28.2$ & $47.2$ \\
& Delta-Matching (\textit{ours}) & $1.3985$ & $61.2$ & $58.7$ & $70.6$ & $16.8$ & $38.1$ & $60.4$ \\
\midrule
AdamW & BF16/FP32 MP & $1.4174$ & $64.5$ & $54.5$ & $67.7$ & $15.3$ & $34.0$ & $59.0$ \\
& Naive FP8 (Stale Delta) & $1.8674$ & $25.2$ & $21.9$ & $47.6$ & $\phantom{0}2.9$ & $26.4$ & $47.8$ \\
& Delta-Matching (\textit{ours}) & $1.4169$ & $60.2$ & $48.3$ & $68.4$ & $15.4$ & $33.4$ & $59.2$ \\
\bottomrule
\end{tabular*}
\end{table}

\begin{table}[!htbp]
\centering
\caption{
\textbf{Commonsense individual benchmark accuracy across optimizers.}
The same single-initialization runs as in Table~\ref{tab:app_muon_full}.
Scores are percentages; Avg is the unweighted mean of the nine benchmarks.
}
\label{tab:app_muon_cs9}
\small
\setlength{\tabcolsep}{3pt}
\resizebox{\textwidth}{!}{%
\begin{tabular}{llccccccccc|c}
\toprule
Optimizer & Method & LAMBADA & HellaSwag & PIQA & ARC-Easy & SciQ & OpenBookQA & WinoGrande & COPA & ARC-Challenge & Avg \\
\midrule
Muon & BF16/FP32 MP & $47.5$ & $58.1$ & $71.7$ & $62.2$ & $90.8$ & $37.2$ & $58.5$ & $70.0$ & $36.9$ & $59.2$ \\
& Naive FP8 (Stale Delta) & $25.1$ & $37.8$ & $64.6$ & $47.7$ & $77.7$ & $29.8$ & $50.0$ & $64.0$ & $27.8$ & $47.2$ \\
& Delta-Matching (\textit{ours}) & $48.1$ & $58.1$ & $73.2$ & $65.0$ & $90.5$ & $37.8$ & $60.5$ & $72.0$ & $38.7$ & $60.4$ \\
\midrule
AdamW & BF16/FP32 MP & $46.1$ & $57.3$ & $72.1$ & $64.6$ & $90.1$ & $36.4$ & $56.7$ & $71.0$ & $36.8$ & $59.0$ \\
& Naive FP8 (Stale Delta) & $25.1$ & $36.4$ & $64.3$ & $46.5$ & $78.6$ & $31.6$ & $52.7$ & $68.0$ & $27.4$ & $47.8$ \\
& Delta-Matching (\textit{ours}) & $46.5$ & $57.0$ & $72.4$ & $62.5$ & $89.9$ & $38.0$ & $56.8$ & $72.0$ & $37.2$ & $59.2$ \\
\bottomrule
\end{tabular}%
}
\end{table}

\paragraph{Results.}
The Muon reference reaches lower validation CE than the paired AdamW
reference, $1.3988$ versus $1.4174$, but stale delta still causes a large
loss increase and downstream degradation.
Relative to Muon's reference, stale delta raises validation CE by
$0.40888\pm0.00145$ nats, while Delta-Matching differs by
$-0.00024\pm0.00016$ nats, with paired standard errors over 2,000 shared
validation windows.
Stale delta lowers RULER-8K from $53.0$ to $26.2$, TriviaQA from $16.5$ to
$3.3$, and CS9 from $59.2$ to $47.2$.
Delta-Matching retains near-reference validation loss and downstream
performance, with higher RULER-8K, extraction, MMLU, and CS9 scores in
this initialization.

\paragraph{Takeaway.}
For this Muon-based recipe and shared initialization, the stale-delta
distortion persists through full training, and Delta-Matching preserves
reference-level model quality.
This comparison changes the optimizer for hidden weight matrices;
normalization gains, including the QK gains, remain on AdamW.

\newpage
\subsection{General Capabilities after Context Extension}
\label{app:stage_capabilities}

Table~\ref{tab:app_stage_capabilities} reports TriviaQA, MMLU, and CS9 results
after context extension, complementing the in-context recall evaluation in
Section~\ref{sec:exp_stage}. Delta-Matching matches the BF16/FP32 mean MMLU score
and achieves a higher mean CS9 score, while TriviaQA remains variable across runs.

\begin{table}[!htbp]
\centering
\caption{
\textbf{General capabilities after context extension from 8K to 64K tokens.}
Results for the 1.67B GDN/GQA hybrid in Section~\ref{sec:exp_stage}.
All rows report mean $\pm$ standard error over two initialization seeds.
The cuDNN/TE FP8 first seed was rerun on the same batches after the original run
diverged, as in Table~\ref{tab:exp_training_stages}.
All scores are percentages. Individual commonsense benchmark accuracies appear
in Table~\ref{tab:app_stage_cs9}.
}
\label{tab:app_stage_capabilities}
\small
\setlength{\tabcolsep}{8pt}
\begin{tabular}{lccc}
\toprule
Method & TriviaQA $\uparrow$ & MMLU $\uparrow$ & CS9 $\uparrow$ \\
\midrule
BF16/FP32 MP & $\phantom{0}4.7\pmstd{4.6}$ & $25.5\pmstd{0.1}$ & $58.9\pmstd{0.4}$ \\
\midrule
cuDNN/TE FP8 & $\phantom{0}5.0\pmstd{3.8}$ & $24.9\pmstd{0.9}$ & $58.5\pmstd{0.3}$ \\
Naive FP8 (Stale Delta) & $13.0\pmstd{1.0}$ & $25.7\pmstd{0.3}$ & $57.4\pmstd{0.3}$ \\
Delta-Matching (\textit{ours}) & $\phantom{0}2.9\pmstd{1.6}$ & $25.5\pmstd{0.1}$ & $59.9\pmstd{0.0}$ \\
\bottomrule
\end{tabular}
\end{table}

\newpage
\subsection{Commonsense Individual Benchmark Accuracy}
\label{app:commonsense_benchmarks}

We report accuracies on nine individual commonsense benchmarks.
CS9 (Avg) is their unweighted mean.
Tables~\ref{tab:app_cs9_breakdown} and~\ref{tab:app_attn_cs9} report individual
accuracies for the default and variant attention rows, respectively, in
Table~\ref{tab:exp_cmp_overall} (Section~\ref{sec:exp_cmp}).
Table~\ref{tab:app_model_cs9} covers the GDN/MLA and KDA/GQA hybrids in
Table~\ref{tab:exp_model_arch} (Section~\ref{sec:exp_model_arch}), and
Table~\ref{tab:app_scale_cs9} covers the parameter scales in
Table~\ref{tab:exp_param_scale} (Section~\ref{sec:exp_param_scale}).
Table~\ref{tab:app_stage_cs9} covers the context-extension runs in
Table~\ref{tab:app_stage_capabilities} (Section~\ref{sec:exp_stage}).
Table~\ref{tab:exp_cs9_breakdown} covers the
FP8-linear runs in Table~\ref{tab:exp_fp8_linear} (Appendix~\ref{app:fp8_linear}).

\begin{benchmarktable}
\centering
\caption{
\textbf{Commonsense individual benchmark accuracy for the default configuration
in Table~\ref{tab:exp_cmp_overall}.}
Mean $\pm$ SE over two seeds; scores are percentages.
}
\label{tab:app_cs9_breakdown}
\scriptsize
\setlength{\tabcolsep}{1pt}
\begin{tabular}{@{}lccccccccc@{}}
\toprule
Method & LAMBADA & HellaSwag & PIQA & ARC-E & SciQ & OBQA & WinoGrande & COPA & ARC-C \\
\midrule
BF16/FP32 MP & $45.7\pmstd{0.4}$ & $57.0\pmstd{0.3}$ & $72.2\pmstd{0.1}$ & $63.1\pmstd{1.5}$ & $89.4\pmstd{0.7}$ & $37.0\pmstd{0.6}$ & $57.7\pmstd{0.9}$ & $70.5\pmstd{0.5}$ & $36.7\pmstd{0.1}$ \\
\midrule
cuDNN/TE FP8 & $33.1\pmstd{2.9}$ & $47.0\pmstd{2.7}$ & $68.9\pmstd{1.1}$ & $56.7\pmstd{0.8}$ & $83.0\pmstd{1.3}$ & $34.8\pmstd{2.2}$ & $53.8\pmstd{0.9}$ & $69.5\pmstd{0.5}$ & $32.1\pmstd{0.9}$ \\
Naive FP8 (Stale Delta) & $24.2\pmstd{0.9}$ & $35.6\pmstd{0.8}$ & $63.5\pmstd{0.8}$ & $46.7\pmstd{0.1}$ & $77.3\pmstd{1.3}$ & $30.8\pmstd{0.8}$ & $52.7\pmstd{0.0}$ & $65.0\pmstd{3.0}$ & $26.8\pmstd{0.6}$ \\
Naive FP8 (Consistent dO) & $42.8\pmstd{1.0}$ & $54.3\pmstd{0.8}$ & $70.9\pmstd{1.0}$ & $60.4\pmstd{0.4}$ & $88.8\pmstd{0.6}$ & $35.6\pmstd{2.2}$ & $57.7\pmstd{0.4}$ & $70.5\pmstd{1.5}$ & $34.6\pmstd{0.4}$ \\
Delta-Matching (\textit{ours}) & $\mathbf{46.7}\pmstd{0.2}$ & $\mathbf{57.2}\pmstd{0.1}$ & $\mathbf{72.4}\pmstd{0.0}$ & $\mathbf{62.4}\pmstd{0.1}$ & $\mathbf{89.8}\pmstd{0.1}$ & $\mathbf{37.1}\pmstd{0.9}$ & $\mathbf{57.9}\pmstd{1.1}$ & $\mathbf{71.0}\pmstd{1.0}$ & $\mathbf{36.3}\pmstd{0.9}$ \\
\bottomrule
\end{tabular}
\end{benchmarktable}

\begin{benchmarktable}
\centering
\caption{
\textbf{Commonsense individual benchmark accuracy across attention configurations.}
Mean $\pm$ standard error over two initialization seeds. Scores are percentages;
Avg is the unweighted mean of the nine benchmarks. Results correspond to
Table~\ref{tab:exp_attn_arch}.
}
\label{tab:app_attn_cs9}
\small
\setlength{\tabcolsep}{3pt}
\resizebox{\textwidth}{!}{%
\begin{tabular}{lccccccccc|c}
\toprule
Method & LAMBADA & HellaSwag & PIQA & ARC-Easy & SciQ & OpenBookQA & WinoGrande & COPA & ARC-Challenge & Avg \\
\midrule
\multicolumn{11}{l}{\textbf{NoPE}} \\
BF16/FP32 MP & $45.3\pmstd{0.5}$ & $57.1\pmstd{0.3}$ & $72.8\pmstd{0.1}$ & $63.2\pmstd{0.2}$ & $88.8\pmstd{0.2}$ & $37.2\pmstd{0.4}$ & $56.8\pmstd{0.2}$ & $68.5\pmstd{1.5}$ & $36.4\pmstd{1.2}$ & $58.5\pmstd{0.1}$ \\
Naive FP8 (Stale Delta) & $31.6\pmstd{4.9}$ & $44.7\pmstd{4.4}$ & $67.1\pmstd{1.6}$ & $53.8\pmstd{3.6}$ & $82.3\pmstd{4.6}$ & $32.6\pmstd{0.8}$ & $54.9\pmstd{0.6}$ & $68.5\pmstd{2.5}$ & $31.0\pmstd{1.8}$ & $51.8\pmstd{2.7}$ \\
Delta-Matching (\textit{ours}) & $44.9\pmstd{0.4}$ & $56.7\pmstd{0.1}$ & $72.3\pmstd{0.1}$ & $63.8\pmstd{0.6}$ & $90.0\pmstd{0.1}$ & $37.1\pmstd{0.1}$ & $57.7\pmstd{0.8}$ & $71.5\pmstd{1.5}$ & $36.0\pmstd{0.6}$ & $58.9\pmstd{0.2}$ \\
\midrule
\multicolumn{11}{l}{\textbf{No QK-norm}} \\
BF16/FP32 MP & $46.9\pmstd{0.5}$ & $57.1\pmstd{0.1}$ & $72.4\pmstd{0.1}$ & $65.3\pmstd{0.8}$ & $89.2\pmstd{0.2}$ & $36.3\pmstd{0.1}$ & $57.2\pmstd{0.5}$ & $70.5\pmstd{1.5}$ & $36.8\pmstd{0.4}$ & $59.1\pmstd{0.2}$ \\
Naive FP8 (Stale Delta) & \multicolumn{10}{c}{\textit{Training diverged}} \\
Delta-Matching (\textit{ours}) & $45.7\pmstd{0.5}$ & $56.9\pmstd{0.2}$ & $72.8\pmstd{0.7}$ & $64.1\pmstd{2.3}$ & $89.6\pmstd{0.1}$ & $36.7\pmstd{0.5}$ & $58.9\pmstd{0.5}$ & $69.0\pmstd{0.0}$ & $36.8\pmstd{0.1}$ & $58.9\pmstd{0.5}$ \\
\midrule
\multicolumn{11}{l}{\textbf{Head dimension $=256$}} \\
BF16/FP32 MP & $45.9\pmstd{0.0}$ & $57.1\pmstd{0.1}$ & $72.3\pmstd{0.1}$ & $64.2\pmstd{1.7}$ & $89.7\pmstd{0.5}$ & $38.2\pmstd{1.8}$ & $58.5\pmstd{0.1}$ & $71.5\pmstd{2.5}$ & $37.2\pmstd{0.6}$ & $59.4\pmstd{0.8}$ \\
Naive FP8 (Stale Delta) & $18.1\pmstd{1.2}$ & $32.6\pmstd{0.2}$ & $62.7\pmstd{0.1}$ & $42.8\pmstd{0.4}$ & $72.4\pmstd{0.4}$ & $29.7\pmstd{0.1}$ & $50.5\pmstd{0.2}$ & $62.0\pmstd{1.0}$ & $25.4\pmstd{0.8}$ & $44.0\pmstd{0.4}$ \\
Delta-Matching (\textit{ours}) & $46.6\pmstd{0.7}$ & $57.3\pmstd{0.2}$ & $72.4\pmstd{0.1}$ & $63.8\pmstd{0.9}$ & $89.5\pmstd{1.2}$ & $37.3\pmstd{0.9}$ & $57.8\pmstd{0.4}$ & $68.5\pmstd{0.5}$ & $36.9\pmstd{0.1}$ & $58.9\pmstd{0.2}$ \\
\bottomrule
\end{tabular}%
}
\end{benchmarktable}

\begin{benchmarktable}
\centering
\caption{
\textbf{Commonsense individual benchmark accuracy across model architectures.}
Results for the 1.67B GDN/MLA and KDA/GQA hybrids in Table~\ref{tab:exp_model_arch}.
Mean $\pm$ standard error over two initialization seeds. Scores are percentages;
Avg is the unweighted mean of the nine benchmarks.
}
\label{tab:app_model_cs9}
\small
\setlength{\tabcolsep}{3pt}
\resizebox{\textwidth}{!}{%
\begin{tabular}{lccccccccc|c}
\toprule
Method & LAMBADA & HellaSwag & PIQA & ARC-Easy & SciQ & OpenBookQA & WinoGrande & COPA & ARC-Challenge & Avg \\
\midrule
\multicolumn{11}{l}{\textbf{GDN/MLA}} \\
BF16/FP32 MP & $46.1\pmstd{0.8}$ & $57.1\pmstd{0.1}$ & $72.5\pmstd{0.2}$ & $64.6\pmstd{1.3}$ & $89.2\pmstd{0.4}$ & $37.3\pmstd{1.1}$ & $58.8\pmstd{1.0}$ & $70.0\pmstd{1.0}$ & $36.9\pmstd{0.1}$ & $59.2\pmstd{0.0}$ \\
Naive FP8 (Stale Delta) & $22.1\pmstd{3.2}$ & $34.9\pmstd{1.6}$ & $63.8\pmstd{1.4}$ & $45.9\pmstd{2.0}$ & $74.9\pmstd{0.4}$ & $31.4\pmstd{0.2}$ & $51.3\pmstd{0.1}$ & $63.0\pmstd{2.0}$ & $26.9\pmstd{1.2}$ & $46.0\pmstd{1.3}$ \\
Delta-Matching (\textit{ours}) & $46.1\pmstd{0.3}$ & $56.9\pmstd{0.4}$ & $72.8\pmstd{0.3}$ & $63.7\pmstd{0.3}$ & $89.5\pmstd{0.1}$ & $38.0\pmstd{1.8}$ & $58.4\pmstd{0.3}$ & $70.5\pmstd{0.5}$ & $37.2\pmstd{0.6}$ & $59.2\pmstd{0.3}$ \\
\midrule
\multicolumn{11}{l}{\textbf{KDA/GQA}} \\
BF16/FP32 MP & $47.2\pmstd{0.7}$ & $58.5\pmstd{0.2}$ & $73.0\pmstd{0.0}$ & $66.1\pmstd{1.0}$ & $90.3\pmstd{0.1}$ & $37.8\pmstd{0.4}$ & $58.4\pmstd{0.2}$ & $72.5\pmstd{3.5}$ & $38.4\pmstd{0.2}$ & $60.2\pmstd{0.3}$ \\
Naive FP8 (Stale Delta) & $26.5\pmstd{3.8}$ & $38.6\pmstd{3.7}$ & $65.1\pmstd{1.0}$ & $49.1\pmstd{3.5}$ & $79.5\pmstd{4.0}$ & $32.0\pmstd{0.4}$ & $51.9\pmstd{0.9}$ & $66.0\pmstd{2.0}$ & $28.0\pmstd{1.4}$ & $48.5\pmstd{2.2}$ \\
Delta-Matching (\textit{ours}) & $47.3\pmstd{0.3}$ & $58.4\pmstd{0.1}$ & $73.0\pmstd{0.9}$ & $65.2\pmstd{0.9}$ & $90.2\pmstd{0.2}$ & $37.8\pmstd{0.4}$ & $59.7\pmstd{0.8}$ & $70.5\pmstd{0.5}$ & $37.8\pmstd{0.1}$ & $60.0\pmstd{0.1}$ \\
\bottomrule
\end{tabular}%
}
\end{benchmarktable}

\begin{benchmarktable}
\centering
\caption{
\textbf{Commonsense individual benchmark accuracy across parameter scales.}
Mean $\pm$ standard error over two initialization seeds. Scores are percentages;
Avg is the unweighted mean of the nine benchmarks. Results correspond to
Table~\ref{tab:exp_param_scale}.
}
\label{tab:app_scale_cs9}
\small
\setlength{\tabcolsep}{3pt}
\resizebox{\textwidth}{!}{%
\begin{tabular}{lccccccccc|c}
\toprule
Method & LAMBADA & HellaSwag & PIQA & ARC-Easy & SciQ & OpenBookQA & WinoGrande & COPA & ARC-Challenge & Avg \\
\midrule
\multicolumn{11}{l}{\textbf{0.569B parameters}} \\
BF16/FP32 MP & $39.9\pmstd{0.0}$ & $49.8\pmstd{0.1}$ & $69.7\pmstd{0.4}$ & $58.1\pmstd{0.5}$ & $87.2\pmstd{0.9}$ & $33.3\pmstd{0.5}$ & $54.0\pmstd{1.1}$ & $69.0\pmstd{1.0}$ & $31.5\pmstd{0.5}$ & $54.7\pmstd{0.1}$ \\
Naive FP8 (Stale Delta) & $39.8\pmstd{0.7}$ & $49.1\pmstd{0.9}$ & $70.0\pmstd{0.4}$ & $58.1\pmstd{1.3}$ & $84.0\pmstd{0.6}$ & $33.7\pmstd{0.1}$ & $52.3\pmstd{0.1}$ & $66.0\pmstd{1.0}$ & $31.4\pmstd{0.6}$ & $53.8\pmstd{0.6}$ \\
Delta-Matching (\textit{ours}) & $39.4\pmstd{0.1}$ & $50.0\pmstd{0.0}$ & $70.3\pmstd{0.2}$ & $56.3\pmstd{1.5}$ & $86.4\pmstd{0.6}$ & $34.5\pmstd{0.5}$ & $54.7\pmstd{0.4}$ & $68.5\pmstd{2.5}$ & $31.6\pmstd{0.5}$ & $54.6\pmstd{0.0}$ \\
\midrule
\multicolumn{11}{l}{\textbf{5.289B parameters}} \\
BF16/FP32 MP & $51.6\pmstd{0.5}$ & $62.7\pmstd{0.3}$ & $74.5\pmstd{0.4}$ & $68.4\pmstd{0.1}$ & $91.0\pmstd{0.2}$ & $39.1\pmstd{0.7}$ & $61.8\pmstd{0.9}$ & $77.0\pmstd{0.0}$ & $41.5\pmstd{0.1}$ & $63.1\pmstd{0.1}$ \\
Naive FP8 (Stale Delta) & $20.1\pmstd{0.3}$ & $35.4\pmstd{0.8}$ & $64.3\pmstd{1.2}$ & $44.3\pmstd{0.9}$ & $73.4\pmstd{1.1}$ & $32.0\pmstd{1.2}$ & $50.9\pmstd{0.2}$ & $61.5\pmstd{2.5}$ & $26.8\pmstd{0.1}$ & $45.4\pmstd{0.3}$ \\
Delta-Matching (\textit{ours}) & $51.2\pmstd{0.4}$ & $62.5\pmstd{0.1}$ & $74.4\pmstd{0.5}$ & $68.0\pmstd{0.5}$ & $90.3\pmstd{0.1}$ & $39.5\pmstd{0.7}$ & $60.7\pmstd{0.6}$ & $71.0\pmstd{3.0}$ & $40.6\pmstd{0.2}$ & $62.0\pmstd{0.4}$ \\
\bottomrule
\end{tabular}%
}
\end{benchmarktable}

\begin{benchmarktable}
\centering
\caption{
\textbf{Commonsense individual benchmark accuracy after context extension from 8K to 64K tokens.}
Results correspond to Table~\ref{tab:app_stage_capabilities},
using the benchmark metrics in Table~\ref{tab:app_cs9_breakdown}.
All rows report mean $\pm$ standard error over two initialization seeds.
The cuDNN/TE FP8 first seed was rerun on the same batches after the original run
diverged, as in Table~\ref{tab:exp_training_stages}.
Scores are percentages; Avg is the unweighted mean of the nine benchmarks.
}
\label{tab:app_stage_cs9}
\small
\setlength{\tabcolsep}{3pt}
\resizebox{\textwidth}{!}{%
\begin{tabular}{lccccccccc|c}
\toprule
Method & LAMBADA & HellaSwag & PIQA & ARC-Easy & SciQ & OpenBookQA & WinoGrande & COPA & ARC-Challenge & Avg \\
\midrule
BF16/FP32 MP & $57.1\pmstd{0.4}$ & $58.2\pmstd{0.1}$ & $73.3\pmstd{0.3}$ & $54.2\pmstd{3.5}$ & $86.7\pmstd{1.3}$ & $36.7\pmstd{0.3}$ & $58.5\pmstd{0.6}$ & $72.5\pmstd{2.5}$ & $32.7\pmstd{0.9}$ & $58.9\pmstd{0.4}$ \\
\midrule
cuDNN/TE FP8 & $53.6\pmstd{1.8}$ & $55.5\pmstd{1.3}$ & $72.0\pmstd{0.9}$ & $57.0\pmstd{0.4}$ & $86.0\pmstd{0.1}$ & $37.6\pmstd{0.4}$ & $58.1\pmstd{1.3}$ & $73.5\pmstd{3.5}$ & $32.8\pmstd{0.6}$ & $58.5\pmstd{0.3}$ \\
Naive FP8 (Stale Delta) & $53.7\pmstd{0.1}$ & $54.2\pmstd{0.0}$ & $72.4\pmstd{0.6}$ & $56.0\pmstd{1.3}$ & $83.8\pmstd{0.1}$ & $36.2\pmstd{1.0}$ & $57.2\pmstd{0.4}$ & $71.0\pmstd{2.0}$ & $32.5\pmstd{0.4}$ & $57.4\pmstd{0.3}$ \\
Delta-Matching (\textit{ours}) & $56.6\pmstd{0.1}$ & $58.0\pmstd{0.1}$ & $72.9\pmstd{0.2}$ & $58.2\pmstd{0.5}$ & $88.3\pmstd{0.5}$ & $36.3\pmstd{0.1}$ & $58.8\pmstd{0.7}$ & $75.0\pmstd{0.0}$ & $34.5\pmstd{0.0}$ & $59.9\pmstd{0.0}$ \\
\bottomrule
\end{tabular}%
}
\end{benchmarktable}

\begin{benchmarktable}
\centering
\caption{
\textbf{Commonsense individual benchmark accuracy with FP8 linear layers.}
Configurations follow Appendix~\ref{app:fp8_linear}.
Mean $\pm$ standard error over two initialization seeds; scores are percentages.
Avg is the unweighted mean of the nine benchmarks.
}
\label{tab:exp_cs9_breakdown}
\small
\setlength{\tabcolsep}{4pt}
\setlength{\aboverulesep}{0pt}\setlength{\belowrulesep}{0pt}
\resizebox{\textwidth}{!}{%
\begin{tabular}{lccccccccc|c}
\toprule
Method & LAMBADA $\uparrow$ & Hella $\uparrow$ & PIQA $\uparrow$ & ARC-e $\uparrow$ & SciQ $\uparrow$ & OBQA $\uparrow$ & Wino $\uparrow$ & COPA $\uparrow$ & ARC-c $\uparrow$ & Avg $\uparrow$ \\
\midrule
BF16/FP32 MP & $45.9\pmstd{0.0}$ & $57.1\pmstd{0.1}$ & $72.3\pmstd{0.1}$ & $64.2\pmstd{1.7}$ & $89.7\pmstd{0.5}$ & $38.2\pmstd{1.8}$ & $58.5\pmstd{0.1}$ & $71.5\pmstd{2.5}$ & $37.2\pmstd{0.6}$ & $59.41\pmstd{0.81}$ \\
BF16/FP32 MP + FP8 FFN & $46.2\pmstd{0.3}$ & $57.2\pmstd{0.0}$ & $73.0\pmstd{0.6}$ & $64.0\pmstd{0.4}$ & $89.2\pmstd{0.8}$ & $37.3\pmstd{0.7}$ & $57.5\pmstd{0.7}$ & $73.0\pmstd{2.0}$ & $36.7\pmstd{0.3}$ & $59.36\pmstd{0.02}$ \\
\midrule
FP8 Attn (Stale Delta) + FP8 FFN & $18.9\pmstd{0.8}$ & $33.2\pmstd{0.3}$ & $63.0\pmstd{1.4}$ & $43.0\pmstd{1.0}$ & $73.7\pmstd{0.4}$ & $29.9\pmstd{0.5}$ & $50.0\pmstd{0.5}$ & $59.0\pmstd{3.0}$ & $25.7\pmstd{0.9}$ & $44.04\pmstd{0.55}$ \\
Delta-Matching + FP8 FFN & $46.7\pmstd{0.1}$ & $56.9\pmstd{0.2}$ & $72.4\pmstd{0.3}$ & $63.2\pmstd{1.3}$ & $89.8\pmstd{0.6}$ & $36.3\pmstd{0.3}$ & $57.7\pmstd{0.6}$ & $65.0\pmstd{2.0}$ & $36.6\pmstd{0.0}$ & $58.29\pmstd{0.51}$ \\
Delta-Matching + FP8 FFN + FP8 QKVO & $46.3\pmstd{0.3}$ & $57.0\pmstd{0.1}$ & $72.3\pmstd{0.1}$ & $62.5\pmstd{0.6}$ & $90.6\pmstd{0.4}$ & $37.8\pmstd{0.4}$ & $57.8\pmstd{0.9}$ & $70.0\pmstd{0.0}$ & $36.2\pmstd{0.5}$ & $58.94\pmstd{0.02}$ \\
\bottomrule
\end{tabular}%
}
\end{benchmarktable}

\newpage
\subsection{RULER Scores by Task}
\label{app:ruler_results}

We provide 13-task RULER breakdowns using the
evaluation protocol in Appendix~\ref{app:eval_recall}.
Tables~\ref{tab:app_ruler_cmp} and~\ref{tab:app_ruler_attn} break down the
default configuration and the NoPE, no-QK-normalization, and
head-dimension-256 variants, respectively, in Table~\ref{tab:exp_cmp_overall}
(Section~\ref{sec:exp_cmp}).
Tables~\ref{tab:app_ruler_model} and~\ref{tab:app_ruler_mamba} break down the
GDN/MLA and KDA/GQA rows and the Mamba-2/GQA rows, respectively, in
Table~\ref{tab:exp_model_arch} (Section~\ref{sec:exp_model_arch}).
Table~\ref{tab:app_ruler_scale} covers the parameter scales in
Table~\ref{tab:exp_param_scale} (Section~\ref{sec:exp_param_scale}).
Table~\ref{tab:app_ruler_stage} breaks down the context-extension results in
Table~\ref{tab:exp_training_stages} (Section~\ref{sec:exp_stage}).
Table~\ref{tab:app_ruler_linear} covers the FP8-linear runs in
Table~\ref{tab:exp_fp8_linear} (Appendix~\ref{app:fp8_linear}).
Table~\ref{tab:app_ruler_pure} provides the pure-GQA breakdown for
Table~\ref{tab:app_pure_gqa_eval} in Appendix~\ref{app:convergent_architectures}.
Table~\ref{tab:app_ruler_muon} covers the Muon and AdamW runs in
Table~\ref{tab:app_muon_full} (Appendix~\ref{app:muon_full_training}).

All scores are percentages, with mean $\pm$ one standard error across two
initialization seeds unless a single seed is shown.
Avg is the unweighted mean of the 13 tasks at each context length, computed
within each seed before aggregation across seeds.
The reported uncertainty reflects variation across seeds, not benchmark sampling.
Naive FP8 uses stale delta; the consistent-$dO$ variant is labeled separately.
Diverged runs retain the \textit{Training diverged} designation.

S1, S2, and S3 denote the three single-needle retrieval tasks; MK1, MK2, and MK3
denote the three multi-key tasks; MV and MQ denote multi-value and multi-query retrieval.
VT is variable tracking, CWE and FWE are common- and frequent-word extraction,
and QA1 and QA2 use SQuAD and HotpotQA, respectively.
Each context length is split into two panels covering all 13 tasks and Avg.

\begin{benchmarktable}
\centering
\caption{
\textbf{Per-task RULER scores for FP8 attention methods.}
4K context.
Results correspond to the default configuration in Table~\ref{tab:exp_cmp_overall}.
Mean $\pm$ standard error over two initialization seeds; scores are percentages.
}
\label{tab:app_ruler_cmp}
\scriptsize
\setlength{\tabcolsep}{2pt}
\begin{tabular*}{\linewidth}{@{\extracolsep{\fill}}lccccccc@{}}
\toprule
\multicolumn{8}{c}{\textbf{4K context}} \\
Method & S1 & S2 & S3 & MK1 & MK2 & MK3 & MV \\
\midrule
BF16/FP32 MP & $99.6\pmstd{0.4}$ & $100.0\pmstd{0.0}$ & $79.5\pmstd{18.7}$ & $95.3\pmstd{2.9}$ & $88.2\pmstd{6.0}$ & $19.6\pmstd{5.6}$ & $80.1\pmstd{9.8}$ \\
cuDNN/TE FP8 & $100.0\pmstd{0.0}$ & $99.7\pmstd{0.3}$ & $48.9\pmstd{12.9}$ & $50.7\pmstd{0.1}$ & $41.9\pmstd{16.1}$ & $3.5\pmstd{0.5}$ & $31.9\pmstd{9.8}$ \\
Naive FP8 & $97.9\pmstd{0.1}$ & $96.3\pmstd{2.1}$ & $19.0\pmstd{2.8}$ & $28.3\pmstd{0.9}$ & $3.5\pmstd{2.5}$ & $0.3\pmstd{0.3}$ & $15.6\pmstd{0.7}$ \\
FP8 (Consistent $dO$) & $100.0\pmstd{0.0}$ & $100.0\pmstd{0.0}$ & $83.9\pmstd{8.9}$ & $90.8\pmstd{6.0}$ & $49.0\pmstd{33.2}$ & $10.3\pmstd{7.9}$ & $66.5\pmstd{2.4}$ \\
Delta-Matching (\textit{ours}) & $100.0\pmstd{0.0}$ & $100.0\pmstd{0.0}$ & $85.6\pmstd{2.0}$ & $93.9\pmstd{1.3}$ & $72.8\pmstd{9.0}$ & $24.3\pmstd{4.1}$ & $83.6\pmstd{4.3}$ \\
\bottomrule
\end{tabular*}
\par\vspace{0.6em}
\begin{tabular*}{\linewidth}{@{\extracolsep{\fill}}lccccccc@{}}
\toprule
\multicolumn{8}{c}{\textbf{4K context (continued)}} \\
Method & MQ & VT & CWE & FWE & QA1 & QA2 & Avg \\
\midrule
BF16/FP32 MP & $72.4\pmstd{3.1}$ & $2.2\pmstd{0.8}$ & $52.8\pmstd{12.8}$ & $34.7\pmstd{10.1}$ & $45.5\pmstd{4.1}$ & $29.8\pmstd{2.2}$ & $61.5\pmstd{3.0}$ \\
cuDNN/TE FP8 & $22.3\pmstd{4.2}$ & $0.7\pmstd{0.0}$ & $32.6\pmstd{7.2}$ & $24.1\pmstd{3.4}$ & $27.5\pmstd{3.9}$ & $20.1\pmstd{0.1}$ & $38.8\pmstd{3.9}$ \\
Naive FP8 & $15.7\pmstd{1.0}$ & $0.4\pmstd{0.2}$ & $15.6\pmstd{5.2}$ & $1.5\pmstd{0.3}$ & $16.4\pmstd{0.2}$ & $12.0\pmstd{0.0}$ & $24.8\pmstd{0.3}$ \\
FP8 (Consistent $dO$) & $61.9\pmstd{0.4}$ & $5.6\pmstd{3.5}$ & $47.8\pmstd{3.6}$ & $23.1\pmstd{13.6}$ & $41.3\pmstd{5.1}$ & $26.8\pmstd{0.2}$ & $54.4\pmstd{3.0}$ \\
Delta-Matching (\textit{ours}) & $75.2\pmstd{5.1}$ & $11.8\pmstd{0.1}$ & $34.8\pmstd{11.1}$ & $42.3\pmstd{13.4}$ & $43.8\pmstd{8.6}$ & $29.9\pmstd{3.1}$ & $61.4\pmstd{1.2}$ \\
\bottomrule
\end{tabular*}
\end{benchmarktable}
\begin{benchmarktable}
\centering
\ContinuedFloat
\caption{\textbf{Per-task RULER scores for FP8 attention methods (continued).} 8K context.}
\scriptsize
\setlength{\tabcolsep}{2pt}
\begin{tabular*}{\linewidth}{@{\extracolsep{\fill}}lccccccc@{}}
\toprule
\multicolumn{8}{c}{\textbf{8K context}} \\
Method & S1 & S2 & S3 & MK1 & MK2 & MK3 & MV \\
\midrule
BF16/FP32 MP & $98.9\pmstd{1.1}$ & $100.0\pmstd{0.0}$ & $82.8\pmstd{16.4}$ & $88.4\pmstd{0.0}$ & $68.7\pmstd{12.1}$ & $4.8\pmstd{2.4}$ & $54.5\pmstd{7.1}$ \\
cuDNN/TE FP8 & $92.4\pmstd{7.2}$ & $98.3\pmstd{0.7}$ & $38.8\pmstd{31.2}$ & $38.3\pmstd{3.5}$ & $22.1\pmstd{9.7}$ & $1.3\pmstd{0.3}$ & $20.3\pmstd{7.9}$ \\
Naive FP8 & $75.0\pmstd{20.4}$ & $88.5\pmstd{2.3}$ & $7.8\pmstd{0.4}$ & $26.4\pmstd{0.4}$ & $1.1\pmstd{0.7}$ & $0.0\pmstd{0.0}$ & $15.9\pmstd{1.2}$ \\
FP8 (Consistent $dO$) & $100.0\pmstd{0.0}$ & $99.7\pmstd{0.1}$ & $80.3\pmstd{8.9}$ & $73.0\pmstd{12.4}$ & $37.9\pmstd{31.3}$ & $2.0\pmstd{1.6}$ & $40.0\pmstd{1.7}$ \\
Delta-Matching (\textit{ours}) & $100.0\pmstd{0.0}$ & $99.8\pmstd{0.2}$ & $63.7\pmstd{3.7}$ & $82.9\pmstd{0.9}$ & $45.8\pmstd{22.4}$ & $4.3\pmstd{2.1}$ & $56.8\pmstd{3.4}$ \\
\bottomrule
\end{tabular*}
\par\vspace{0.6em}
\begin{tabular*}{\linewidth}{@{\extracolsep{\fill}}lccccccc@{}}
\toprule
\multicolumn{8}{c}{\textbf{8K context (continued)}} \\
Method & MQ & VT & CWE & FWE & QA1 & QA2 & Avg \\
\midrule
BF16/FP32 MP & $56.9\pmstd{2.6}$ & $4.2\pmstd{1.1}$ & $27.6\pmstd{4.9}$ & $31.1\pmstd{11.5}$ & $35.6\pmstd{2.4}$ & $25.8\pmstd{1.0}$ & $52.3\pmstd{2.3}$ \\
cuDNN/TE FP8 & $13.1\pmstd{4.3}$ & $8.9\pmstd{1.7}$ & $16.1\pmstd{3.1}$ & $18.7\pmstd{0.6}$ & $18.6\pmstd{2.6}$ & $16.7\pmstd{0.9}$ & $31.0\pmstd{3.9}$ \\
Naive FP8 & $12.2\pmstd{1.1}$ & $0.2\pmstd{0.2}$ & $6.7\pmstd{2.7}$ & $2.1\pmstd{0.1}$ & $9.0\pmstd{2.6}$ & $12.2\pmstd{1.2}$ & $19.8\pmstd{2.1}$ \\
FP8 (Consistent $dO$) & $39.5\pmstd{0.7}$ & $8.8\pmstd{7.1}$ & $12.0\pmstd{3.1}$ & $13.8\pmstd{12.6}$ & $31.6\pmstd{1.2}$ & $23.5\pmstd{0.5}$ & $43.2\pmstd{2.3}$ \\
Delta-Matching (\textit{ours}) & $54.3\pmstd{10.0}$ & $6.3\pmstd{1.0}$ & $13.0\pmstd{3.5}$ & $37.6\pmstd{7.7}$ & $30.8\pmstd{7.6}$ & $26.2\pmstd{2.4}$ & $47.8\pmstd{0.5}$ \\
\bottomrule
\end{tabular*}
\end{benchmarktable}

\begin{benchmarktable}
\centering
\caption{
\textbf{Per-task RULER scores across attention configurations.}
4K context.
Results correspond to the NoPE, no-QK-normalization, and head-dimension-256
variants in Table~\ref{tab:exp_attn_arch}.
Mean $\pm$ standard error over two initialization seeds; scores are percentages.
}
\label{tab:app_ruler_attn}
\scriptsize
\setlength{\tabcolsep}{2pt}
\begin{tabular*}{\linewidth}{@{\extracolsep{\fill}}lccccccc@{}}
\toprule
\multicolumn{8}{c}{\textbf{4K context}} \\
Method & S1 & S2 & S3 & MK1 & MK2 & MK3 & MV \\
\midrule
\multicolumn{8}{l}{\textbf{NoPE}} \\
BF16/FP32 MP & $100.0\pmstd{0.0}$ & $100.0\pmstd{0.0}$ & $79.3\pmstd{17.9}$ & $92.9\pmstd{3.3}$ & $63.8\pmstd{7.4}$ & $12.8\pmstd{2.2}$ & $74.5\pmstd{8.1}$ \\
Naive FP8 & $100.0\pmstd{0.0}$ & $97.3\pmstd{2.5}$ & $59.1\pmstd{26.3}$ & $56.9\pmstd{3.3}$ & $9.1\pmstd{0.9}$ & $2.6\pmstd{0.0}$ & $40.2\pmstd{14.3}$ \\
Delta-Matching (\textit{ours}) & $100.0\pmstd{0.0}$ & $100.0\pmstd{0.0}$ & $76.9\pmstd{7.5}$ & $88.9\pmstd{7.7}$ & $49.5\pmstd{10.7}$ & $15.7\pmstd{9.3}$ & $70.3\pmstd{8.3}$ \\
\midrule
\multicolumn{8}{l}{\textbf{No QK-norm}} \\
BF16/FP32 MP & $100.0\pmstd{0.0}$ & $100.0\pmstd{0.0}$ & $89.0\pmstd{4.2}$ & $91.9\pmstd{1.1}$ & $73.1\pmstd{21.9}$ & $14.2\pmstd{4.4}$ & $72.5\pmstd{0.3}$ \\
Naive FP8 & \multicolumn{7}{c}{\textit{Training diverged}} \\
Delta-Matching (\textit{ours}) & $100.0\pmstd{0.0}$ & $99.9\pmstd{0.1}$ & $92.3\pmstd{5.5}$ & $81.7\pmstd{3.5}$ & $91.9\pmstd{3.1}$ & $35.2\pmstd{15.2}$ & $76.3\pmstd{4.1}$ \\
\midrule
\multicolumn{8}{l}{\textbf{Head Dim 256}} \\
BF16/FP32 MP & $100.0\pmstd{0.0}$ & $100.0\pmstd{0.0}$ & $91.8\pmstd{3.2}$ & $89.6\pmstd{8.4}$ & $93.8\pmstd{3.0}$ & $21.1\pmstd{5.5}$ & $71.3\pmstd{4.9}$ \\
Naive FP8 & $74.3\pmstd{18.5}$ & $10.3\pmstd{0.9}$ & $5.0\pmstd{4.6}$ & $18.9\pmstd{9.3}$ & $0.9\pmstd{0.5}$ & $0.6\pmstd{0.6}$ & $13.6\pmstd{4.9}$ \\
Delta-Matching (\textit{ours}) & $100.0\pmstd{0.0}$ & $100.0\pmstd{0.0}$ & $93.1\pmstd{0.3}$ & $97.2\pmstd{0.0}$ & $87.7\pmstd{6.1}$ & $13.8\pmstd{8.8}$ & $85.1\pmstd{1.3}$ \\
\bottomrule
\end{tabular*}
\par\vspace{0.6em}
\begin{tabular*}{\linewidth}{@{\extracolsep{\fill}}lccccccc@{}}
\toprule
\multicolumn{8}{c}{\textbf{4K context (continued)}} \\
Method & MQ & VT & CWE & FWE & QA1 & QA2 & Avg \\
\midrule
\multicolumn{8}{l}{\textbf{NoPE}} \\
BF16/FP32 MP & $73.3\pmstd{1.9}$ & $16.9\pmstd{6.3}$ & $36.5\pmstd{5.7}$ & $39.9\pmstd{4.1}$ & $38.3\pmstd{7.5}$ & $26.4\pmstd{1.0}$ & $58.0\pmstd{3.0}$ \\
Naive FP8 & $33.7\pmstd{9.5}$ & $0.6\pmstd{0.1}$ & $37.4\pmstd{6.1}$ & $17.3\pmstd{10.1}$ & $23.0\pmstd{11.4}$ & $19.2\pmstd{3.6}$ & $38.2\pmstd{2.7}$ \\
Delta-Matching (\textit{ours}) & $67.9\pmstd{7.6}$ & $24.2\pmstd{20.9}$ & $41.7\pmstd{2.6}$ & $51.8\pmstd{5.2}$ & $45.6\pmstd{1.6}$ & $29.0\pmstd{1.2}$ & $58.6\pmstd{1.1}$ \\
\midrule
\multicolumn{8}{l}{\textbf{No QK-norm}} \\
BF16/FP32 MP & $66.4\pmstd{0.4}$ & $3.6\pmstd{3.2}$ & $43.6\pmstd{1.3}$ & $37.7\pmstd{1.0}$ & $41.4\pmstd{1.8}$ & $26.8\pmstd{0.2}$ & $58.5\pmstd{1.9}$ \\
Naive FP8 & \multicolumn{7}{c}{\textit{Training diverged}} \\
Delta-Matching (\textit{ours}) & $71.7\pmstd{3.1}$ & $5.5\pmstd{2.4}$ & $37.7\pmstd{10.4}$ & $19.7\pmstd{17.8}$ & $45.2\pmstd{0.4}$ & $29.7\pmstd{0.5}$ & $60.5\pmstd{0.6}$ \\
\midrule
\multicolumn{8}{l}{\textbf{Head Dim 256}} \\
BF16/FP32 MP & $77.8\pmstd{4.0}$ & $11.5\pmstd{3.7}$ & $36.9\pmstd{3.1}$ & $37.2\pmstd{4.1}$ & $54.9\pmstd{2.5}$ & $32.3\pmstd{0.3}$ & $62.9\pmstd{1.6}$ \\
Naive FP8 & $0.2\pmstd{0.2}$ & $0.0\pmstd{0.0}$ & $7.7\pmstd{5.3}$ & $0.1\pmstd{0.0}$ & $7.6\pmstd{3.2}$ & $9.7\pmstd{1.1}$ & $11.5\pmstd{0.2}$ \\
Delta-Matching (\textit{ours}) & $77.6\pmstd{0.1}$ & $14.8\pmstd{9.8}$ & $32.3\pmstd{2.8}$ & $31.6\pmstd{11.0}$ & $45.6\pmstd{4.8}$ & $30.6\pmstd{1.2}$ & $62.3\pmstd{1.2}$ \\
\bottomrule
\end{tabular*}
\end{benchmarktable}
\begin{benchmarktable}
\centering
\ContinuedFloat
\caption{\textbf{Per-task RULER scores across attention configurations (continued).} 8K context.}
\scriptsize
\setlength{\tabcolsep}{2pt}
\begin{tabular*}{\linewidth}{@{\extracolsep{\fill}}lccccccc@{}}
\toprule
\multicolumn{8}{c}{\textbf{8K context}} \\
Method & S1 & S2 & S3 & MK1 & MK2 & MK3 & MV \\
\midrule
\multicolumn{8}{l}{\textbf{NoPE}} \\
BF16/FP32 MP & $100.0\pmstd{0.0}$ & $100.0\pmstd{0.0}$ & $82.2\pmstd{2.4}$ & $88.2\pmstd{2.6}$ & $52.5\pmstd{6.7}$ & $4.5\pmstd{0.3}$ & $54.5\pmstd{9.8}$ \\
Naive FP8 & $98.9\pmstd{1.1}$ & $78.5\pmstd{20.5}$ & $36.5\pmstd{17.9}$ & $44.4\pmstd{8.6}$ & $6.8\pmstd{1.8}$ & $0.3\pmstd{0.1}$ & $24.2\pmstd{0.9}$ \\
Delta-Matching (\textit{ours}) & $100.0\pmstd{0.0}$ & $100.0\pmstd{0.0}$ & $59.9\pmstd{17.1}$ & $85.7\pmstd{6.5}$ & $35.5\pmstd{9.3}$ & $6.9\pmstd{4.5}$ & $59.5\pmstd{5.3}$ \\
\midrule
\multicolumn{8}{l}{\textbf{No QK-norm}} \\
BF16/FP32 MP & $98.9\pmstd{1.1}$ & $100.0\pmstd{0.0}$ & $80.9\pmstd{8.1}$ & $69.6\pmstd{9.4}$ & $44.0\pmstd{26.0}$ & $2.7\pmstd{1.1}$ & $53.1\pmstd{10.8}$ \\
Naive FP8 & \multicolumn{7}{c}{\textit{Training diverged}} \\
Delta-Matching (\textit{ours}) & $100.0\pmstd{0.0}$ & $100.0\pmstd{0.0}$ & $94.2\pmstd{3.8}$ & $73.8\pmstd{4.2}$ & $77.8\pmstd{1.0}$ & $20.8\pmstd{11.6}$ & $47.2\pmstd{9.0}$ \\
\midrule
\multicolumn{8}{l}{\textbf{Head Dim 256}} \\
BF16/FP32 MP & $100.0\pmstd{0.0}$ & $100.0\pmstd{0.0}$ & $83.8\pmstd{9.8}$ & $83.4\pmstd{3.2}$ & $77.3\pmstd{13.5}$ & $5.4\pmstd{3.6}$ & $58.7\pmstd{1.1}$ \\
Naive FP8 & $42.1\pmstd{12.3}$ & $10.4\pmstd{1.8}$ & $6.3\pmstd{1.7}$ & $14.7\pmstd{2.9}$ & $0.0\pmstd{0.0}$ & $0.2\pmstd{0.0}$ & $11.0\pmstd{2.5}$ \\
Delta-Matching (\textit{ours}) & $99.9\pmstd{0.1}$ & $99.9\pmstd{0.1}$ & $84.0\pmstd{7.8}$ & $91.7\pmstd{3.1}$ & $78.6\pmstd{2.6}$ & $6.7\pmstd{4.9}$ & $56.5\pmstd{2.9}$ \\
\bottomrule
\end{tabular*}
\par\vspace{0.6em}
\begin{tabular*}{\linewidth}{@{\extracolsep{\fill}}lccccccc@{}}
\toprule
\multicolumn{8}{c}{\textbf{8K context (continued)}} \\
Method & MQ & VT & CWE & FWE & QA1 & QA2 & Avg \\
\midrule
\multicolumn{8}{l}{\textbf{NoPE}} \\
BF16/FP32 MP & $51.1\pmstd{4.4}$ & $17.3\pmstd{7.3}$ & $19.1\pmstd{0.6}$ & $40.2\pmstd{2.7}$ & $30.0\pmstd{7.4}$ & $23.3\pmstd{0.7}$ & $51.0\pmstd{2.6}$ \\
Naive FP8 & $22.5\pmstd{0.9}$ & $1.0\pmstd{0.8}$ & $11.7\pmstd{1.9}$ & $22.7\pmstd{8.4}$ & $17.8\pmstd{9.0}$ & $16.0\pmstd{3.2}$ & $29.3\pmstd{2.7}$ \\
Delta-Matching (\textit{ours}) & $57.9\pmstd{0.0}$ & $17.0\pmstd{14.8}$ & $12.4\pmstd{4.9}$ & $50.1\pmstd{5.1}$ & $36.6\pmstd{0.6}$ & $25.2\pmstd{0.0}$ & $49.7\pmstd{1.4}$ \\
\midrule
\multicolumn{8}{l}{\textbf{No QK-norm}} \\
BF16/FP32 MP & $50.2\pmstd{5.8}$ & $2.3\pmstd{1.1}$ & $7.0\pmstd{0.8}$ & $40.7\pmstd{12.4}$ & $31.5\pmstd{3.5}$ & $23.3\pmstd{0.1}$ & $46.5\pmstd{5.2}$ \\
Naive FP8 & \multicolumn{7}{c}{\textit{Training diverged}} \\
Delta-Matching (\textit{ours}) & $49.4\pmstd{0.4}$ & $2.6\pmstd{1.0}$ & $5.2\pmstd{1.5}$ & $22.7\pmstd{14.4}$ & $33.7\pmstd{0.7}$ & $26.2\pmstd{0.4}$ & $50.3\pmstd{0.4}$ \\
\midrule
\multicolumn{8}{l}{\textbf{Head Dim 256}} \\
BF16/FP32 MP & $58.6\pmstd{0.9}$ & $2.5\pmstd{0.3}$ & $8.8\pmstd{4.2}$ & $37.7\pmstd{3.7}$ & $44.0\pmstd{3.8}$ & $26.7\pmstd{0.5}$ & $52.8\pmstd{2.1}$ \\
Naive FP8 & $1.1\pmstd{1.1}$ & $0.0\pmstd{0.0}$ & $2.9\pmstd{2.1}$ & $0.6\pmstd{0.0}$ & $5.4\pmstd{0.2}$ & $5.4\pmstd{0.0}$ & $7.7\pmstd{0.7}$ \\
Delta-Matching (\textit{ours}) & $51.5\pmstd{3.2}$ & $13.7\pmstd{0.5}$ & $12.0\pmstd{5.9}$ & $42.2\pmstd{4.9}$ & $37.0\pmstd{6.4}$ & $26.9\pmstd{2.3}$ & $53.9\pmstd{2.9}$ \\
\bottomrule
\end{tabular*}
\end{benchmarktable}

\begin{benchmarktable}
\centering
\caption{
\textbf{Per-task RULER scores for GDN/MLA and KDA/GQA hybrids.}
4K context.
Results correspond to the GDN/MLA and KDA/GQA rows in Table~\ref{tab:exp_model_arch}.
Mean $\pm$ standard error over two initialization seeds; scores are percentages.
}
\label{tab:app_ruler_model}
\scriptsize
\setlength{\tabcolsep}{2pt}
\begin{tabular*}{\linewidth}{@{\extracolsep{\fill}}lccccccc@{}}
\toprule
\multicolumn{8}{c}{\textbf{4K context}} \\
Method & S1 & S2 & S3 & MK1 & MK2 & MK3 & MV \\
\midrule
\multicolumn{8}{l}{\textbf{GDN/MLA}} \\
BF16/FP32 MP & $100.0\pmstd{0.0}$ & $100.0\pmstd{0.0}$ & $72.8\pmstd{11.8}$ & $92.0\pmstd{2.6}$ & $63.3\pmstd{14.3}$ & $24.0\pmstd{9.2}$ & $63.9\pmstd{11.1}$ \\
Naive FP8 & $99.9\pmstd{0.1}$ & $97.7\pmstd{2.1}$ & $9.5\pmstd{1.3}$ & $40.5\pmstd{7.9}$ & $5.3\pmstd{4.9}$ & $0.6\pmstd{0.2}$ & $17.2\pmstd{1.7}$ \\
Delta-Matching (\textit{ours}) & $100.0\pmstd{0.0}$ & $100.0\pmstd{0.0}$ & $94.5\pmstd{1.7}$ & $92.9\pmstd{0.5}$ & $59.8\pmstd{6.0}$ & $19.7\pmstd{6.9}$ & $79.1\pmstd{7.1}$ \\
\midrule
\multicolumn{8}{l}{\textbf{KDA/GQA}} \\
BF16/FP32 MP & $100.0\pmstd{0.0}$ & $100.0\pmstd{0.0}$ & $86.0\pmstd{1.8}$ & $93.2\pmstd{4.6}$ & $91.5\pmstd{4.7}$ & $30.9\pmstd{7.7}$ & $83.0\pmstd{2.0}$ \\
Naive FP8 & $92.7\pmstd{7.3}$ & $63.7\pmstd{36.3}$ & $20.3\pmstd{20.3}$ & $45.0\pmstd{20.0}$ & $8.6\pmstd{8.2}$ & $5.2\pmstd{5.2}$ & $28.0\pmstd{10.1}$ \\
Delta-Matching (\textit{ours}) & $100.0\pmstd{0.0}$ & $100.0\pmstd{0.0}$ & $80.7\pmstd{10.7}$ & $87.0\pmstd{5.4}$ & $55.7\pmstd{0.7}$ & $25.0\pmstd{17.4}$ & $75.6\pmstd{2.3}$ \\
\bottomrule
\end{tabular*}
\par\vspace{0.6em}
\begin{tabular*}{\linewidth}{@{\extracolsep{\fill}}lccccccc@{}}
\toprule
\multicolumn{8}{c}{\textbf{4K context (continued)}} \\
Method & MQ & VT & CWE & FWE & QA1 & QA2 & Avg \\
\midrule
\multicolumn{8}{l}{\textbf{GDN/MLA}} \\
BF16/FP32 MP & $69.7\pmstd{0.2}$ & $2.0\pmstd{0.1}$ & $36.5\pmstd{0.7}$ & $12.0\pmstd{6.7}$ & $47.8\pmstd{5.2}$ & $29.5\pmstd{3.3}$ & $54.9\pmstd{2.2}$ \\
Naive FP8 & $13.3\pmstd{3.4}$ & $0.5\pmstd{0.3}$ & $19.1\pmstd{0.5}$ & $6.1\pmstd{5.9}$ & $15.9\pmstd{1.9}$ & $11.0\pmstd{2.4}$ & $25.9\pmstd{2.0}$ \\
Delta-Matching (\textit{ours}) & $77.0\pmstd{7.7}$ & $14.6\pmstd{6.6}$ & $44.3\pmstd{0.5}$ & $25.8\pmstd{3.3}$ & $45.7\pmstd{0.9}$ & $31.4\pmstd{0.0}$ & $60.4\pmstd{0.6}$ \\
\midrule
\multicolumn{8}{l}{\textbf{KDA/GQA}} \\
BF16/FP32 MP & $80.9\pmstd{1.3}$ & $24.6\pmstd{1.7}$ & $50.9\pmstd{7.8}$ & $30.0\pmstd{8.7}$ & $52.1\pmstd{0.3}$ & $30.1\pmstd{0.1}$ & $65.6\pmstd{0.5}$ \\
Naive FP8 & $19.3\pmstd{15.8}$ & $0.6\pmstd{0.6}$ & $19.5\pmstd{10.0}$ & $0.9\pmstd{0.8}$ & $19.6\pmstd{5.2}$ & $17.2\pmstd{2.2}$ & $26.2\pmstd{10.9}$ \\
Delta-Matching (\textit{ours}) & $71.7\pmstd{2.3}$ & $17.4\pmstd{3.9}$ & $44.7\pmstd{2.4}$ & $44.3\pmstd{2.8}$ & $54.5\pmstd{0.1}$ & $33.1\pmstd{1.1}$ & $60.7\pmstd{2.4}$ \\
\bottomrule
\end{tabular*}
\end{benchmarktable}
\begin{benchmarktable}
\centering
\ContinuedFloat
\caption{\textbf{Per-task RULER scores for GDN/MLA and KDA/GQA hybrids (continued).} 8K context.}
\scriptsize
\setlength{\tabcolsep}{2pt}
\begin{tabular*}{\linewidth}{@{\extracolsep{\fill}}lccccccc@{}}
\toprule
\multicolumn{8}{c}{\textbf{8K context}} \\
Method & S1 & S2 & S3 & MK1 & MK2 & MK3 & MV \\
\midrule
\multicolumn{8}{l}{\textbf{GDN/MLA}} \\
BF16/FP32 MP & $100.0\pmstd{0.0}$ & $99.9\pmstd{0.1}$ & $85.1\pmstd{3.1}$ & $77.0\pmstd{1.8}$ & $52.4\pmstd{10.8}$ & $6.6\pmstd{3.4}$ & $47.6\pmstd{2.4}$ \\
Naive FP8 & $84.4\pmstd{13.0}$ & $68.7\pmstd{30.7}$ & $2.8\pmstd{1.8}$ & $32.4\pmstd{10.0}$ & $1.8\pmstd{1.6}$ & $0.2\pmstd{0.2}$ & $14.8\pmstd{2.1}$ \\
Delta-Matching (\textit{ours}) & $99.8\pmstd{0.2}$ & $99.7\pmstd{0.1}$ & $87.3\pmstd{1.5}$ & $76.8\pmstd{0.2}$ & $47.9\pmstd{9.7}$ & $8.0\pmstd{4.6}$ & $56.8\pmstd{8.2}$ \\
\midrule
\multicolumn{8}{l}{\textbf{KDA/GQA}} \\
BF16/FP32 MP & $100.0\pmstd{0.0}$ & $99.7\pmstd{0.3}$ & $70.1\pmstd{16.3}$ & $86.4\pmstd{5.0}$ & $58.0\pmstd{6.0}$ & $14.8\pmstd{8.2}$ & $46.1\pmstd{6.1}$ \\
Naive FP8 & $92.4\pmstd{7.6}$ & $64.5\pmstd{34.9}$ & $27.8\pmstd{27.8}$ & $30.9\pmstd{13.5}$ & $3.1\pmstd{2.7}$ & $1.1\pmstd{1.1}$ & $22.8\pmstd{8.0}$ \\
Delta-Matching (\textit{ours}) & $98.0\pmstd{2.0}$ & $100.0\pmstd{0.0}$ & $76.1\pmstd{5.9}$ & $82.4\pmstd{4.2}$ & $41.5\pmstd{10.9}$ & $8.4\pmstd{5.0}$ & $57.8\pmstd{6.3}$ \\
\bottomrule
\end{tabular*}
\par\vspace{0.6em}
\begin{tabular*}{\linewidth}{@{\extracolsep{\fill}}lccccccc@{}}
\toprule
\multicolumn{8}{c}{\textbf{8K context (continued)}} \\
Method & MQ & VT & CWE & FWE & QA1 & QA2 & Avg \\
\midrule
\multicolumn{8}{l}{\textbf{GDN/MLA}} \\
BF16/FP32 MP & $44.1\pmstd{2.1}$ & $1.9\pmstd{1.1}$ & $12.7\pmstd{6.4}$ & $23.2\pmstd{4.9}$ & $38.1\pmstd{2.5}$ & $27.2\pmstd{2.4}$ & $47.4\pmstd{0.1}$ \\
Naive FP8 & $13.0\pmstd{3.3}$ & $0.0\pmstd{0.0}$ & $5.1\pmstd{3.1}$ & $0.9\pmstd{0.9}$ & $10.5\pmstd{3.1}$ & $10.1\pmstd{2.1}$ & $18.9\pmstd{5.3}$ \\
Delta-Matching (\textit{ours}) & $52.9\pmstd{6.4}$ & $23.9\pmstd{3.9}$ & $17.2\pmstd{2.1}$ & $13.9\pmstd{7.9}$ & $38.7\pmstd{0.5}$ & $26.3\pmstd{0.9}$ & $50.0\pmstd{1.7}$ \\
\midrule
\multicolumn{8}{l}{\textbf{KDA/GQA}} \\
BF16/FP32 MP & $51.3\pmstd{1.3}$ & $13.1\pmstd{8.4}$ & $14.0\pmstd{5.0}$ & $27.4\pmstd{5.5}$ & $41.3\pmstd{0.5}$ & $27.3\pmstd{2.5}$ & $50.0\pmstd{2.3}$ \\
Naive FP8 & $17.3\pmstd{12.1}$ & $1.5\pmstd{1.5}$ & $5.0\pmstd{0.5}$ & $1.6\pmstd{1.5}$ & $14.8\pmstd{5.0}$ & $14.9\pmstd{3.1}$ & $22.9\pmstd{9.2}$ \\
Delta-Matching (\textit{ours}) & $54.6\pmstd{6.8}$ & $12.1\pmstd{3.7}$ & $12.8\pmstd{6.9}$ & $44.9\pmstd{2.8}$ & $43.4\pmstd{2.8}$ & $27.0\pmstd{0.4}$ & $50.7\pmstd{2.2}$ \\
\bottomrule
\end{tabular*}
\end{benchmarktable}

\begin{benchmarktable}
\centering
\caption{
\textbf{Per-task RULER scores across parameter scales.}
4K context.
Results correspond to Table~\ref{tab:exp_param_scale}.
Mean $\pm$ standard error over two initialization seeds; scores are percentages.
}
\label{tab:app_ruler_scale}
\scriptsize
\setlength{\tabcolsep}{2pt}
\begin{tabular*}{\linewidth}{@{\extracolsep{\fill}}lccccccc@{}}
\toprule
\multicolumn{8}{c}{\textbf{4K context}} \\
Method & S1 & S2 & S3 & MK1 & MK2 & MK3 & MV \\
\midrule
\multicolumn{8}{l}{\textbf{0.569B}} \\
BF16/FP32 MP & $100.0\pmstd{0.0}$ & $100.0\pmstd{0.0}$ & $19.0\pmstd{6.6}$ & $82.8\pmstd{1.2}$ & $54.7\pmstd{1.3}$ & $4.4\pmstd{0.4}$ & $54.0\pmstd{0.0}$ \\
Naive FP8 & $100.0\pmstd{0.0}$ & $93.9\pmstd{6.1}$ & $56.7\pmstd{23.3}$ & $92.1\pmstd{1.9}$ & $40.0\pmstd{19.4}$ & $6.3\pmstd{2.5}$ & $58.5\pmstd{14.2}$ \\
Delta-Matching (\textit{ours}) & $100.0\pmstd{0.0}$ & $100.0\pmstd{0.0}$ & $41.0\pmstd{8.8}$ & $81.7\pmstd{3.3}$ & $45.0\pmstd{0.0}$ & $3.1\pmstd{0.3}$ & $53.4\pmstd{3.5}$ \\
\midrule
\multicolumn{8}{l}{\textbf{5.289B}} \\
BF16/FP32 MP & $100.0\pmstd{0.0}$ & $100.0\pmstd{0.0}$ & $83.9\pmstd{4.9}$ & $94.2\pmstd{5.0}$ & $81.9\pmstd{0.5}$ & $38.3\pmstd{10.9}$ & $56.6\pmstd{0.1}$ \\
Naive FP8 & $90.6\pmstd{9.2}$ & $82.6\pmstd{17.4}$ & $8.9\pmstd{8.9}$ & $39.9\pmstd{11.9}$ & $1.4\pmstd{0.2}$ & $1.3\pmstd{0.9}$ & $19.5\pmstd{5.0}$ \\
Delta-Matching (\textit{ours}) & $100.0\pmstd{0.0}$ & $100.0\pmstd{0.0}$ & $77.5\pmstd{19.5}$ & $98.2\pmstd{1.4}$ & $89.1\pmstd{1.7}$ & $36.5\pmstd{0.9}$ & $84.2\pmstd{9.0}$ \\
\bottomrule
\end{tabular*}
\par\vspace{0.6em}
\begin{tabular*}{\linewidth}{@{\extracolsep{\fill}}lccccccc@{}}
\toprule
\multicolumn{8}{c}{\textbf{4K context (continued)}} \\
Method & MQ & VT & CWE & FWE & QA1 & QA2 & Avg \\
\midrule
\multicolumn{8}{l}{\textbf{0.569B}} \\
BF16/FP32 MP & $58.3\pmstd{4.1}$ & $1.2\pmstd{0.7}$ & $34.2\pmstd{0.6}$ & $15.6\pmstd{3.4}$ & $41.1\pmstd{1.7}$ & $25.1\pmstd{0.5}$ & $45.4\pmstd{0.5}$ \\
Naive FP8 & $57.2\pmstd{8.1}$ & $0.5\pmstd{0.2}$ & $36.0\pmstd{8.5}$ & $26.8\pmstd{5.7}$ & $40.3\pmstd{7.1}$ & $24.9\pmstd{2.3}$ & $48.7\pmstd{1.2}$ \\
Delta-Matching (\textit{ours}) & $52.5\pmstd{6.0}$ & $1.3\pmstd{0.1}$ & $34.9\pmstd{8.1}$ & $26.2\pmstd{3.5}$ & $36.3\pmstd{4.3}$ & $26.0\pmstd{1.2}$ & $46.3\pmstd{1.0}$ \\
\midrule
\multicolumn{8}{l}{\textbf{5.289B}} \\
BF16/FP32 MP & $67.2\pmstd{0.1}$ & $36.9\pmstd{20.1}$ & $49.9\pmstd{7.7}$ & $52.2\pmstd{4.7}$ & $57.2\pmstd{0.4}$ & $34.6\pmstd{0.8}$ & $65.6\pmstd{0.2}$ \\
Naive FP8 & $6.3\pmstd{5.1}$ & $0.1\pmstd{0.1}$ & $11.3\pmstd{0.5}$ & $9.2\pmstd{8.3}$ & $20.4\pmstd{2.2}$ & $12.4\pmstd{0.8}$ & $23.4\pmstd{2.4}$ \\
Delta-Matching (\textit{ours}) & $84.5\pmstd{0.8}$ & $29.3\pmstd{4.3}$ & $55.3\pmstd{1.3}$ & $62.8\pmstd{6.6}$ & $49.4\pmstd{6.4}$ & $33.2\pmstd{2.6}$ & $69.2\pmstd{2.0}$ \\
\bottomrule
\end{tabular*}
\end{benchmarktable}
\begin{benchmarktable}
\centering
\ContinuedFloat
\caption{\textbf{Per-task RULER scores across parameter scales (continued).} 8K context.}
\scriptsize
\setlength{\tabcolsep}{2pt}
\begin{tabular*}{\linewidth}{@{\extracolsep{\fill}}lccccccc@{}}
\toprule
\multicolumn{8}{c}{\textbf{8K context}} \\
Method & S1 & S2 & S3 & MK1 & MK2 & MK3 & MV \\
\midrule
\multicolumn{8}{l}{\textbf{0.569B}} \\
BF16/FP32 MP & $99.6\pmstd{0.4}$ & $99.2\pmstd{0.6}$ & $10.4\pmstd{0.2}$ & $66.5\pmstd{3.5}$ & $36.2\pmstd{1.8}$ & $1.1\pmstd{1.1}$ & $43.1\pmstd{0.5}$ \\
Naive FP8 & $99.9\pmstd{0.1}$ & $99.1\pmstd{0.9}$ & $50.6\pmstd{17.6}$ & $79.2\pmstd{5.2}$ & $24.9\pmstd{9.5}$ & $0.9\pmstd{0.3}$ & $42.1\pmstd{6.1}$ \\
Delta-Matching (\textit{ours}) & $98.7\pmstd{1.1}$ & $99.9\pmstd{0.1}$ & $29.5\pmstd{18.3}$ & $69.8\pmstd{1.6}$ & $26.7\pmstd{0.9}$ & $0.3\pmstd{0.3}$ & $40.3\pmstd{3.6}$ \\
\midrule
\multicolumn{8}{l}{\textbf{5.289B}} \\
BF16/FP32 MP & $99.8\pmstd{0.2}$ & $99.1\pmstd{0.7}$ & $79.7\pmstd{1.3}$ & $88.5\pmstd{5.7}$ & $68.7\pmstd{1.7}$ & $16.6\pmstd{9.8}$ & $47.1\pmstd{6.5}$ \\
Naive FP8 & $75.2\pmstd{21.0}$ & $57.5\pmstd{40.7}$ & $1.1\pmstd{1.1}$ & $25.5\pmstd{8.7}$ & $0.4\pmstd{0.2}$ & $0.0\pmstd{0.0}$ & $17.6\pmstd{6.4}$ \\
Delta-Matching (\textit{ours}) & $100.0\pmstd{0.0}$ & $100.0\pmstd{0.0}$ & $90.2\pmstd{1.0}$ & $95.5\pmstd{1.1}$ & $73.1\pmstd{11.1}$ & $19.8\pmstd{1.0}$ & $63.0\pmstd{6.0}$ \\
\bottomrule
\end{tabular*}
\par\vspace{0.6em}
\begin{tabular*}{\linewidth}{@{\extracolsep{\fill}}lccccccc@{}}
\toprule
\multicolumn{8}{c}{\textbf{8K context (continued)}} \\
Method & MQ & VT & CWE & FWE & QA1 & QA2 & Avg \\
\midrule
\multicolumn{8}{l}{\textbf{0.569B}} \\
BF16/FP32 MP & $47.5\pmstd{5.1}$ & $2.1\pmstd{1.5}$ & $15.5\pmstd{0.2}$ & $16.2\pmstd{7.5}$ & $30.8\pmstd{1.6}$ & $23.4\pmstd{1.8}$ & $37.8\pmstd{0.4}$ \\
Naive FP8 & $40.4\pmstd{4.2}$ & $0.3\pmstd{0.0}$ & $10.7\pmstd{2.9}$ & $25.4\pmstd{7.5}$ & $31.7\pmstd{1.5}$ & $22.6\pmstd{1.6}$ & $40.6\pmstd{1.1}$ \\
Delta-Matching (\textit{ours}) & $37.9\pmstd{5.2}$ & $2.3\pmstd{0.7}$ & $13.0\pmstd{2.1}$ & $24.4\pmstd{10.9}$ & $30.6\pmstd{3.4}$ & $22.7\pmstd{0.3}$ & $38.2\pmstd{0.8}$ \\
\midrule
\multicolumn{8}{l}{\textbf{5.289B}} \\
BF16/FP32 MP & $54.1\pmstd{0.8}$ & $20.9\pmstd{18.8}$ & $4.5\pmstd{1.9}$ & $37.2\pmstd{5.2}$ & $48.3\pmstd{2.1}$ & $30.4\pmstd{0.2}$ & $53.5\pmstd{1.8}$ \\
Naive FP8 & $4.8\pmstd{1.4}$ & $0.0\pmstd{0.0}$ & $4.9\pmstd{1.9}$ & $1.9\pmstd{0.9}$ & $10.1\pmstd{0.7}$ & $13.0\pmstd{1.0}$ & $16.3\pmstd{5.8}$ \\
Delta-Matching (\textit{ours}) & $58.8\pmstd{1.8}$ & $16.2\pmstd{13.1}$ & $15.3\pmstd{0.7}$ & $58.1\pmstd{0.2}$ & $41.6\pmstd{2.2}$ & $29.6\pmstd{1.4}$ & $58.6\pmstd{0.1}$ \\
\bottomrule
\end{tabular*}
\end{benchmarktable}

\begin{benchmarktable}
\centering
\caption{
\textbf{Per-task RULER scores after context extension.}
4K context.
Results correspond to Table~\ref{tab:exp_training_stages}.
Mean $\pm$ standard error over two initialization seeds; scores are percentages.
cuDNN/TE FP8 includes the same-batch first-seed rerun documented in
Table~\ref{tab:exp_training_stages}; its mean across the five context lengths is
$48.4\pm5.0$.
}
\label{tab:app_ruler_stage}
\scriptsize
\setlength{\tabcolsep}{2pt}
\begin{tabular*}{\linewidth}{@{\extracolsep{\fill}}lccccccc@{}}
\toprule
\multicolumn{8}{c}{\textbf{4K context}} \\
Method & S1 & S2 & S3 & MK1 & MK2 & MK3 & MV \\
\midrule
BF16/FP32 MP & $100.0\pmstd{0.0}$ & $97.2\pmstd{2.4}$ & $75.0\pmstd{7.8}$ & $81.9\pmstd{0.7}$ & $60.8\pmstd{0.0}$ & $65.6\pmstd{4.0}$ & $90.8\pmstd{0.9}$ \\
cuDNN/TE FP8 & $99.7\pmstd{0.3}$ & $94.4\pmstd{5.6}$ & $86.3\pmstd{12.1}$ & $85.2\pmstd{2.0}$ & $53.7\pmstd{7.7}$ & $58.2\pmstd{27.4}$ & $60.3\pmstd{24.9}$ \\
Naive FP8 & $100.0\pmstd{0.0}$ & $92.8\pmstd{7.2}$ & $82.6\pmstd{2.4}$ & $80.1\pmstd{11.5}$ & $77.1\pmstd{14.7}$ & $52.8\pmstd{0.2}$ & $76.5\pmstd{1.0}$ \\
Delta-Matching (\textit{ours}) & $100.0\pmstd{0.0}$ & $99.9\pmstd{0.1}$ & $88.3\pmstd{7.5}$ & $90.8\pmstd{0.4}$ & $56.8\pmstd{19.0}$ & $69.7\pmstd{1.7}$ & $82.3\pmstd{6.4}$ \\
\bottomrule
\end{tabular*}
\par\vspace{0.6em}
\begin{tabular*}{\linewidth}{@{\extracolsep{\fill}}lccccccc@{}}
\toprule
\multicolumn{8}{c}{\textbf{4K context (continued)}} \\
Method & MQ & VT & CWE & FWE & QA1 & QA2 & Avg \\
\midrule
BF16/FP32 MP & $80.0\pmstd{5.5}$ & $53.5\pmstd{2.9}$ & $31.2\pmstd{5.0}$ & $56.1\pmstd{7.0}$ & $37.7\pmstd{4.7}$ & $32.6\pmstd{1.2}$ & $66.3\pmstd{0.9}$ \\
cuDNN/TE FP8 & $52.0\pmstd{21.0}$ & $27.6\pmstd{4.5}$ & $37.6\pmstd{1.8}$ & $51.2\pmstd{4.6}$ & $31.0\pmstd{10.4}$ & $21.6\pmstd{2.8}$ & $58.4\pmstd{8.7}$ \\
Naive FP8 & $74.1\pmstd{5.6}$ & $15.8\pmstd{7.2}$ & $54.6\pmstd{12.8}$ & $45.7\pmstd{0.9}$ & $34.4\pmstd{2.8}$ & $29.7\pmstd{1.5}$ & $62.8\pmstd{1.7}$ \\
Delta-Matching (\textit{ours}) & $73.3\pmstd{6.3}$ & $55.7\pmstd{5.4}$ & $37.1\pmstd{1.4}$ & $56.8\pmstd{1.9}$ & $39.1\pmstd{7.9}$ & $29.7\pmstd{3.9}$ & $67.6\pmstd{4.4}$ \\
\bottomrule
\end{tabular*}
\end{benchmarktable}
\begin{benchmarktable}
\centering
\ContinuedFloat
\caption{\textbf{Per-task RULER scores after context extension (continued).} 8K context.}
\scriptsize
\setlength{\tabcolsep}{2pt}
\begin{tabular*}{\linewidth}{@{\extracolsep{\fill}}lccccccc@{}}
\toprule
\multicolumn{8}{c}{\textbf{8K context}} \\
Method & S1 & S2 & S3 & MK1 & MK2 & MK3 & MV \\
\midrule
BF16/FP32 MP & $100.0\pmstd{0.0}$ & $99.1\pmstd{0.9}$ & $75.7\pmstd{4.1}$ & $74.8\pmstd{3.6}$ & $45.8\pmstd{4.8}$ & $58.5\pmstd{3.7}$ & $65.0\pmstd{12.3}$ \\
cuDNN/TE FP8 & $100.0\pmstd{0.0}$ & $99.5\pmstd{0.5}$ & $67.6\pmstd{13.2}$ & $78.0\pmstd{2.8}$ & $53.7\pmstd{7.9}$ & $47.2\pmstd{29.2}$ & $49.6\pmstd{18.4}$ \\
Naive FP8 & $100.0\pmstd{0.0}$ & $98.7\pmstd{0.7}$ & $85.0\pmstd{2.2}$ & $76.8\pmstd{4.4}$ & $65.7\pmstd{17.1}$ & $37.4\pmstd{4.4}$ & $58.4\pmstd{3.8}$ \\
Delta-Matching (\textit{ours}) & $100.0\pmstd{0.0}$ & $99.9\pmstd{0.1}$ & $86.1\pmstd{6.3}$ & $87.6\pmstd{1.6}$ & $44.3\pmstd{23.7}$ & $55.5\pmstd{4.1}$ & $73.1\pmstd{4.4}$ \\
\bottomrule
\end{tabular*}
\par\vspace{0.6em}
\begin{tabular*}{\linewidth}{@{\extracolsep{\fill}}lccccccc@{}}
\toprule
\multicolumn{8}{c}{\textbf{8K context (continued)}} \\
Method & MQ & VT & CWE & FWE & QA1 & QA2 & Avg \\
\midrule
BF16/FP32 MP & $69.2\pmstd{9.2}$ & $45.5\pmstd{6.5}$ & $11.5\pmstd{4.5}$ & $57.2\pmstd{1.1}$ & $35.6\pmstd{4.4}$ & $31.7\pmstd{0.5}$ & $59.2\pmstd{3.3}$ \\
cuDNN/TE FP8 & $46.2\pmstd{17.1}$ & $12.0\pmstd{2.6}$ & $16.8\pmstd{3.2}$ & $53.8\pmstd{0.1}$ & $27.2\pmstd{8.0}$ & $19.4\pmstd{4.2}$ & $51.6\pmstd{5.7}$ \\
Naive FP8 & $59.6\pmstd{2.3}$ & $13.5\pmstd{8.3}$ & $25.5\pmstd{7.3}$ & $51.7\pmstd{0.8}$ & $31.2\pmstd{0.6}$ & $27.4\pmstd{1.6}$ & $56.2\pmstd{1.5}$ \\
Delta-Matching (\textit{ours}) & $71.0\pmstd{0.8}$ & $42.7\pmstd{1.4}$ & $17.7\pmstd{0.6}$ & $62.8\pmstd{5.4}$ & $37.2\pmstd{2.4}$ & $30.3\pmstd{0.5}$ & $62.2\pmstd{2.8}$ \\
\bottomrule
\end{tabular*}
\end{benchmarktable}
\begin{benchmarktable}
\centering
\ContinuedFloat
\caption{\textbf{Per-task RULER scores after context extension (continued).} 16K context.}
\scriptsize
\setlength{\tabcolsep}{2pt}
\begin{tabular*}{\linewidth}{@{\extracolsep{\fill}}lccccccc@{}}
\toprule
\multicolumn{8}{c}{\textbf{16K context}} \\
Method & S1 & S2 & S3 & MK1 & MK2 & MK3 & MV \\
\midrule
BF16/FP32 MP & $100.0\pmstd{0.0}$ & $99.9\pmstd{0.1}$ & $75.3\pmstd{5.5}$ & $70.5\pmstd{3.9}$ & $44.9\pmstd{1.5}$ & $49.2\pmstd{0.8}$ & $55.3\pmstd{13.8}$ \\
cuDNN/TE FP8 & $100.0\pmstd{0.0}$ & $99.9\pmstd{0.1}$ & $91.3\pmstd{3.3}$ & $76.5\pmstd{3.9}$ & $49.8\pmstd{5.2}$ & $36.0\pmstd{20.8}$ & $43.7\pmstd{16.8}$ \\
Naive FP8 & $100.0\pmstd{0.0}$ & $99.9\pmstd{0.1}$ & $82.4\pmstd{4.0}$ & $74.7\pmstd{2.7}$ & $45.6\pmstd{35.0}$ & $27.9\pmstd{2.9}$ & $49.2\pmstd{0.7}$ \\
Delta-Matching (\textit{ours}) & $100.0\pmstd{0.0}$ & $100.0\pmstd{0.0}$ & $91.0\pmstd{5.0}$ & $81.2\pmstd{3.0}$ & $46.3\pmstd{31.9}$ & $36.0\pmstd{9.4}$ & $60.4\pmstd{0.6}$ \\
\bottomrule
\end{tabular*}
\par\vspace{0.6em}
\begin{tabular*}{\linewidth}{@{\extracolsep{\fill}}lccccccc@{}}
\toprule
\multicolumn{8}{c}{\textbf{16K context (continued)}} \\
Method & MQ & VT & CWE & FWE & QA1 & QA2 & Avg \\
\midrule
BF16/FP32 MP & $57.2\pmstd{15.0}$ & $34.9\pmstd{7.4}$ & $16.0\pmstd{1.5}$ & $50.3\pmstd{3.3}$ & $32.8\pmstd{6.4}$ & $30.2\pmstd{1.2}$ & $55.1\pmstd{3.7}$ \\
cuDNN/TE FP8 & $43.3\pmstd{16.3}$ & $6.6\pmstd{3.6}$ & $10.6\pmstd{2.1}$ & $37.7\pmstd{4.8}$ & $24.3\pmstd{4.5}$ & $18.9\pmstd{3.3}$ & $49.1\pmstd{5.4}$ \\
Naive FP8 & $54.9\pmstd{3.3}$ & $3.0\pmstd{2.1}$ & $21.7\pmstd{5.6}$ & $37.1\pmstd{6.0}$ & $26.9\pmstd{3.7}$ & $27.1\pmstd{0.9}$ & $50.0\pmstd{2.2}$ \\
Delta-Matching (\textit{ours}) & $60.6\pmstd{2.6}$ & $17.1\pmstd{12.3}$ & $14.9\pmstd{3.9}$ & $49.0\pmstd{2.4}$ & $32.6\pmstd{3.4}$ & $31.3\pmstd{2.1}$ & $55.4\pmstd{1.1}$ \\
\bottomrule
\end{tabular*}
\end{benchmarktable}
\begin{benchmarktable}
\centering
\ContinuedFloat
\caption{\textbf{Per-task RULER scores after context extension (continued).} 32K context.}
\scriptsize
\setlength{\tabcolsep}{2pt}
\begin{tabular*}{\linewidth}{@{\extracolsep{\fill}}lccccccc@{}}
\toprule
\multicolumn{8}{c}{\textbf{32K context}} \\
Method & S1 & S2 & S3 & MK1 & MK2 & MK3 & MV \\
\midrule
BF16/FP32 MP & $100.0\pmstd{0.0}$ & $99.4\pmstd{0.2}$ & $77.6\pmstd{0.4}$ & $64.2\pmstd{11.8}$ & $34.6\pmstd{1.4}$ & $43.8\pmstd{7.0}$ & $43.9\pmstd{16.3}$ \\
cuDNN/TE FP8 & $100.0\pmstd{0.0}$ & $99.9\pmstd{0.1}$ & $79.6\pmstd{0.0}$ & $70.7\pmstd{3.5}$ & $40.0\pmstd{1.2}$ & $21.7\pmstd{4.9}$ & $41.7\pmstd{15.3}$ \\
Naive FP8 & $100.0\pmstd{0.0}$ & $99.8\pmstd{0.2}$ & $84.9\pmstd{2.5}$ & $66.2\pmstd{1.4}$ & $35.9\pmstd{24.9}$ & $18.5\pmstd{1.7}$ & $42.9\pmstd{0.6}$ \\
Delta-Matching (\textit{ours}) & $100.0\pmstd{0.0}$ & $99.9\pmstd{0.1}$ & $90.1\pmstd{4.7}$ & $75.6\pmstd{2.0}$ & $44.5\pmstd{33.5}$ & $24.0\pmstd{17.4}$ & $49.6\pmstd{3.8}$ \\
\bottomrule
\end{tabular*}
\par\vspace{0.6em}
\begin{tabular*}{\linewidth}{@{\extracolsep{\fill}}lccccccc@{}}
\toprule
\multicolumn{8}{c}{\textbf{32K context (continued)}} \\
Method & MQ & VT & CWE & FWE & QA1 & QA2 & Avg \\
\midrule
BF16/FP32 MP & $51.5\pmstd{16.7}$ & $23.2\pmstd{2.8}$ & $8.3\pmstd{0.5}$ & $43.1\pmstd{1.9}$ & $30.8\pmstd{7.8}$ & $29.2\pmstd{1.0}$ & $50.0\pmstd{5.1}$ \\
cuDNN/TE FP8 & $39.0\pmstd{13.0}$ & $5.8\pmstd{2.1}$ & $3.0\pmstd{0.4}$ & $39.8\pmstd{8.5}$ & $21.3\pmstd{2.5}$ & $18.3\pmstd{2.9}$ & $44.7\pmstd{3.4}$ \\
Naive FP8 & $48.1\pmstd{2.1}$ & $4.5\pmstd{2.8}$ & $7.9\pmstd{1.0}$ & $36.0\pmstd{3.4}$ & $24.9\pmstd{3.9}$ & $25.2\pmstd{2.6}$ & $45.8\pmstd{0.8}$ \\
Delta-Matching (\textit{ours}) & $53.5\pmstd{5.4}$ & $19.3\pmstd{12.6}$ & $6.3\pmstd{3.1}$ & $49.7\pmstd{3.9}$ & $31.8\pmstd{0.8}$ & $27.1\pmstd{1.7}$ & $51.6\pmstd{0.1}$ \\
\bottomrule
\end{tabular*}
\end{benchmarktable}
\begin{benchmarktable}
\centering
\ContinuedFloat
\caption{\textbf{Per-task RULER scores after context extension (continued).} 64K context.}
\scriptsize
\setlength{\tabcolsep}{2pt}
\begin{tabular*}{\linewidth}{@{\extracolsep{\fill}}lccccccc@{}}
\toprule
\multicolumn{8}{c}{\textbf{64K context}} \\
Method & S1 & S2 & S3 & MK1 & MK2 & MK3 & MV \\
\midrule
BF16/FP32 MP & $100.0\pmstd{0.0}$ & $99.2\pmstd{0.6}$ & $80.6\pmstd{7.4}$ & $57.8\pmstd{8.0}$ & $40.8\pmstd{1.8}$ & $28.7\pmstd{4.9}$ & $37.1\pmstd{9.9}$ \\
cuDNN/TE FP8 & $98.0\pmstd{2.0}$ & $85.6\pmstd{2.4}$ & $76.0\pmstd{4.6}$ & $69.7\pmstd{0.9}$ & $26.6\pmstd{10.2}$ & $9.1\pmstd{2.9}$ & $31.5\pmstd{6.1}$ \\
Naive FP8 & $99.4\pmstd{0.6}$ & $83.9\pmstd{5.3}$ & $83.1\pmstd{6.1}$ & $62.5\pmstd{4.7}$ & $19.6\pmstd{19.6}$ & $7.3\pmstd{0.1}$ & $41.9\pmstd{0.8}$ \\
Delta-Matching (\textit{ours}) & $98.6\pmstd{0.4}$ & $85.0\pmstd{5.6}$ & $79.9\pmstd{7.9}$ & $73.3\pmstd{6.1}$ & $42.1\pmstd{34.3}$ & $10.5\pmstd{9.3}$ & $37.0\pmstd{3.2}$ \\
\bottomrule
\end{tabular*}
\par\vspace{0.6em}
\begin{tabular*}{\linewidth}{@{\extracolsep{\fill}}lccccccc@{}}
\toprule
\multicolumn{8}{c}{\textbf{64K context (continued)}} \\
Method & MQ & VT & CWE & FWE & QA1 & QA2 & Avg \\
\midrule
BF16/FP32 MP & $45.1\pmstd{11.0}$ & $14.3\pmstd{3.9}$ & $2.0\pmstd{0.9}$ & $42.5\pmstd{8.1}$ & $26.9\pmstd{7.7}$ & $23.5\pmstd{2.3}$ & $46.0\pmstd{4.5}$ \\
cuDNN/TE FP8 & $32.1\pmstd{8.5}$ & $7.5\pmstd{2.6}$ & $1.0\pmstd{0.6}$ & $28.1\pmstd{16.1}$ & $19.9\pmstd{2.9}$ & $15.0\pmstd{2.4}$ & $38.5\pmstd{1.8}$ \\
Naive FP8 & $45.7\pmstd{1.2}$ & $8.7\pmstd{1.8}$ & $1.2\pmstd{0.5}$ & $22.1\pmstd{15.8}$ & $25.4\pmstd{3.6}$ & $20.0\pmstd{1.2}$ & $40.1\pmstd{0.6}$ \\
Delta-Matching (\textit{ours}) & $44.0\pmstd{3.8}$ & $17.9\pmstd{0.3}$ & $1.1\pmstd{0.5}$ & $48.7\pmstd{4.1}$ & $29.8\pmstd{1.4}$ & $24.5\pmstd{0.7}$ & $45.6\pmstd{0.6}$ \\
\bottomrule
\end{tabular*}
\end{benchmarktable}

\begin{benchmarktable}
\centering
\caption{
\textbf{Per-task RULER scores with FP8 linear layers.}
4K context.
Results correspond to Table~\ref{tab:exp_fp8_linear}.
Mean $\pm$ standard error over two initialization seeds; scores are percentages.
FFN8 and QKVO8 denote FP8 FFNs and FP8 attention projections, respectively.
}
\label{tab:app_ruler_linear}
\scriptsize
\setlength{\tabcolsep}{2pt}
\begin{tabular*}{\linewidth}{@{\extracolsep{\fill}}lccccccc@{}}
\toprule
\multicolumn{8}{c}{\textbf{4K context}} \\
Method & S1 & S2 & S3 & MK1 & MK2 & MK3 & MV \\
\midrule
BF16/FP32 MP & $100.0\pmstd{0.0}$ & $100.0\pmstd{0.0}$ & $91.8\pmstd{3.2}$ & $89.6\pmstd{8.4}$ & $93.8\pmstd{3.0}$ & $21.1\pmstd{5.5}$ & $71.3\pmstd{4.9}$ \\
BF16/FP32 MP + FFN8 & $99.9\pmstd{0.1}$ & $100.0\pmstd{0.0}$ & $72.9\pmstd{1.3}$ & $91.9\pmstd{3.3}$ & $89.1\pmstd{5.1}$ & $5.1\pmstd{1.9}$ & $77.6\pmstd{11.9}$ \\
Naive FP8 + FFN8 & $86.7\pmstd{13.3}$ & $70.7\pmstd{28.3}$ & $6.4\pmstd{6.2}$ & $24.7\pmstd{9.1}$ & $0.8\pmstd{0.4}$ & $0.3\pmstd{0.3}$ & $14.5\pmstd{5.1}$ \\
Delta-Matching + FFN8 & $100.0\pmstd{0.0}$ & $100.0\pmstd{0.0}$ & $97.1\pmstd{1.3}$ & $94.9\pmstd{0.5}$ & $77.7\pmstd{7.1}$ & $24.8\pmstd{4.6}$ & $78.9\pmstd{15.1}$ \\
Delta-Matching + FFN8 + QKVO8 & $98.8\pmstd{1.2}$ & $100.0\pmstd{0.0}$ & $85.1\pmstd{1.1}$ & $97.5\pmstd{1.1}$ & $81.0\pmstd{6.6}$ & $11.1\pmstd{0.7}$ & $79.2\pmstd{1.3}$ \\
\bottomrule
\end{tabular*}
\par\vspace{0.6em}
\begin{tabular*}{\linewidth}{@{\extracolsep{\fill}}lccccccc@{}}
\toprule
\multicolumn{8}{c}{\textbf{4K context (continued)}} \\
Method & MQ & VT & CWE & FWE & QA1 & QA2 & Avg \\
\midrule
BF16/FP32 MP & $77.8\pmstd{4.0}$ & $11.5\pmstd{3.7}$ & $36.9\pmstd{3.1}$ & $37.2\pmstd{4.1}$ & $54.9\pmstd{2.5}$ & $32.3\pmstd{0.3}$ & $62.94\pmstd{1.56}$ \\
BF16/FP32 MP + FFN8 & $72.5\pmstd{7.1}$ & $15.6\pmstd{3.8}$ & $25.9\pmstd{7.0}$ & $27.3\pmstd{16.8}$ & $48.7\pmstd{1.9}$ & $30.5\pmstd{1.5}$ & $58.22\pmstd{1.64}$ \\
Naive FP8 + FFN8 & $7.3\pmstd{7.3}$ & $1.0\pmstd{1.0}$ & $7.9\pmstd{0.1}$ & $0.1\pmstd{0.0}$ & $12.4\pmstd{0.8}$ & $12.3\pmstd{1.7}$ & $18.85\pmstd{5.67}$ \\
Delta-Matching + FFN8 & $77.8\pmstd{3.3}$ & $11.9\pmstd{11.5}$ & $41.3\pmstd{4.7}$ & $47.6\pmstd{3.5}$ & $50.1\pmstd{1.1}$ & $31.4\pmstd{0.6}$ & $64.11\pmstd{3.22}$ \\
Delta-Matching + FFN8 + QKVO8 & $75.0\pmstd{7.2}$ & $7.5\pmstd{0.3}$ & $33.4\pmstd{0.2}$ & $28.9\pmstd{20.1}$ & $45.8\pmstd{4.0}$ & $30.7\pmstd{0.7}$ & $59.54\pmstd{2.12}$ \\
\bottomrule
\end{tabular*}
\end{benchmarktable}
\begin{benchmarktable}
\centering
\ContinuedFloat
\caption{\textbf{Per-task RULER scores with FP8 linear layers (continued).} 8K context.}
\scriptsize
\setlength{\tabcolsep}{2pt}
\begin{tabular*}{\linewidth}{@{\extracolsep{\fill}}lccccccc@{}}
\toprule
\multicolumn{8}{c}{\textbf{8K context}} \\
Method & S1 & S2 & S3 & MK1 & MK2 & MK3 & MV \\
\midrule
BF16/FP32 MP & $100.0\pmstd{0.0}$ & $100.0\pmstd{0.0}$ & $83.8\pmstd{9.8}$ & $83.4\pmstd{3.2}$ & $77.3\pmstd{13.5}$ & $5.4\pmstd{3.6}$ & $58.7\pmstd{1.1}$ \\
BF16/FP32 MP + FFN8 & $98.8\pmstd{1.2}$ & $99.7\pmstd{0.3}$ & $61.3\pmstd{7.3}$ & $81.9\pmstd{3.7}$ & $70.6\pmstd{5.0}$ & $1.1\pmstd{0.9}$ & $55.0\pmstd{13.9}$ \\
Naive FP8 + FFN8 & $66.0\pmstd{30.0}$ & $54.4\pmstd{29.6}$ & $2.1\pmstd{0.3}$ & $23.6\pmstd{4.0}$ & $0.4\pmstd{0.2}$ & $0.2\pmstd{0.2}$ & $10.9\pmstd{7.8}$ \\
Delta-Matching + FFN8 & $100.0\pmstd{0.0}$ & $100.0\pmstd{0.0}$ & $91.5\pmstd{1.7}$ & $86.7\pmstd{1.3}$ & $46.9\pmstd{18.1}$ & $5.4\pmstd{0.2}$ & $52.2\pmstd{5.7}$ \\
Delta-Matching + FFN8 + QKVO8 & $100.0\pmstd{0.0}$ & $98.9\pmstd{1.1}$ & $74.5\pmstd{6.9}$ & $89.8\pmstd{4.8}$ & $70.1\pmstd{1.9}$ & $2.7\pmstd{0.3}$ & $54.0\pmstd{3.0}$ \\
\bottomrule
\end{tabular*}
\par\vspace{0.6em}
\begin{tabular*}{\linewidth}{@{\extracolsep{\fill}}lccccccc@{}}
\toprule
\multicolumn{8}{c}{\textbf{8K context (continued)}} \\
Method & MQ & VT & CWE & FWE & QA1 & QA2 & Avg \\
\midrule
BF16/FP32 MP & $58.6\pmstd{0.9}$ & $2.5\pmstd{0.3}$ & $8.8\pmstd{4.2}$ & $37.7\pmstd{3.7}$ & $44.0\pmstd{3.8}$ & $26.7\pmstd{0.5}$ & $52.84\pmstd{2.14}$ \\
BF16/FP32 MP + FFN8 & $50.2\pmstd{10.2}$ & $7.8\pmstd{4.5}$ & $12.2\pmstd{0.0}$ & $36.2\pmstd{4.6}$ & $38.9\pmstd{1.3}$ & $26.0\pmstd{1.6}$ & $49.21\pmstd{1.95}$ \\
Naive FP8 + FFN8 & $3.5\pmstd{3.5}$ & $0.0\pmstd{0.0}$ & $3.9\pmstd{1.3}$ & $0.2\pmstd{0.2}$ & $7.8\pmstd{0.2}$ & $10.2\pmstd{0.6}$ & $14.09\pmstd{5.86}$ \\
Delta-Matching + FFN8 & $55.3\pmstd{7.2}$ & $9.0\pmstd{8.6}$ & $16.3\pmstd{7.6}$ & $39.8\pmstd{7.6}$ & $39.8\pmstd{0.8}$ & $26.2\pmstd{0.6}$ & $51.48\pmstd{2.79}$ \\
Delta-Matching + FFN8 + QKVO8 & $56.3\pmstd{3.8}$ & $9.7\pmstd{0.1}$ & $17.4\pmstd{0.0}$ & $30.3\pmstd{9.8}$ & $37.3\pmstd{4.3}$ & $27.5\pmstd{0.9}$ & $51.42\pmstd{1.73}$ \\
\bottomrule
\end{tabular*}
\end{benchmarktable}

\begin{benchmarktable}
\centering
\caption{
\textbf{Per-task RULER scores for the pure GQA model.}
4K context.
Results correspond to Table~\ref{tab:app_pure_gqa_eval}.
Mean $\pm$ standard error over two initialization seeds; scores are percentages.
}
\label{tab:app_ruler_pure}
\scriptsize
\setlength{\tabcolsep}{2pt}
\begin{tabular*}{\linewidth}{@{\extracolsep{\fill}}lccccccc@{}}
\toprule
\multicolumn{8}{c}{\textbf{4K context}} \\
Method & S1 & S2 & S3 & MK1 & MK2 & MK3 & MV \\
\midrule
BF16/FP32 MP & $99.9\pmstd{0.1}$ & $98.0\pmstd{1.0}$ & $52.3\pmstd{0.3}$ & $82.6\pmstd{5.4}$ & $58.2\pmstd{14.8}$ & $11.7\pmstd{5.9}$ & $45.4\pmstd{15.5}$ \\
Naive FP8 & $100.0\pmstd{0.0}$ & $98.9\pmstd{0.9}$ & $63.2\pmstd{5.4}$ & $78.2\pmstd{5.0}$ & $73.8\pmstd{16.6}$ & $19.1\pmstd{13.9}$ & $54.4\pmstd{10.8}$ \\
Delta-Matching (\textit{ours}) & $99.9\pmstd{0.1}$ & $62.6\pmstd{37.4}$ & $74.4\pmstd{18.8}$ & $68.7\pmstd{12.1}$ & $39.0\pmstd{28.0}$ & $7.8\pmstd{1.2}$ & $52.0\pmstd{6.2}$ \\
\bottomrule
\end{tabular*}
\par\vspace{0.6em}
\begin{tabular*}{\linewidth}{@{\extracolsep{\fill}}lccccccc@{}}
\toprule
\multicolumn{8}{c}{\textbf{4K context (continued)}} \\
Method & MQ & VT & CWE & FWE & QA1 & QA2 & Avg \\
\midrule
BF16/FP32 MP & $43.5\pmstd{10.3}$ & $2.5\pmstd{0.8}$ & $45.8\pmstd{4.9}$ & $12.0\pmstd{9.2}$ & $54.5\pmstd{1.5}$ & $30.8\pmstd{0.4}$ & $49.0\pmstd{0.6}$ \\
Naive FP8 & $55.9\pmstd{0.1}$ & $0.7\pmstd{0.1}$ & $41.6\pmstd{4.7}$ & $13.7\pmstd{9.1}$ & $49.5\pmstd{3.1}$ & $28.4\pmstd{0.0}$ & $52.1\pmstd{2.7}$ \\
Delta-Matching (\textit{ours}) & $50.8\pmstd{0.1}$ & $5.4\pmstd{4.2}$ & $36.2\pmstd{1.8}$ & $12.8\pmstd{0.4}$ & $50.9\pmstd{2.5}$ & $29.4\pmstd{0.0}$ & $45.4\pmstd{3.1}$ \\
\bottomrule
\end{tabular*}
\end{benchmarktable}
\begin{benchmarktable}
\centering
\ContinuedFloat
\caption{\textbf{Per-task RULER scores for the pure GQA model (continued).} 8K context.}
\scriptsize
\setlength{\tabcolsep}{2pt}
\begin{tabular*}{\linewidth}{@{\extracolsep{\fill}}lccccccc@{}}
\toprule
\multicolumn{8}{c}{\textbf{8K context}} \\
Method & S1 & S2 & S3 & MK1 & MK2 & MK3 & MV \\
\midrule
BF16/FP32 MP & $99.2\pmstd{0.8}$ & $97.0\pmstd{0.8}$ & $39.7\pmstd{11.1}$ & $61.2\pmstd{13.6}$ & $23.8\pmstd{11.0}$ & $3.5\pmstd{2.7}$ & $33.7\pmstd{12.0}$ \\
Naive FP8 & $99.7\pmstd{0.1}$ & $66.9\pmstd{14.7}$ & $45.4\pmstd{10.8}$ & $64.2\pmstd{0.8}$ & $61.4\pmstd{15.2}$ & $6.5\pmstd{5.5}$ & $28.0\pmstd{4.3}$ \\
Delta-Matching (\textit{ours}) & $99.5\pmstd{0.3}$ & $73.5\pmstd{24.5}$ & $55.7\pmstd{13.5}$ & $59.1\pmstd{0.1}$ & $31.3\pmstd{19.7}$ & $5.7\pmstd{1.7}$ & $31.3\pmstd{5.6}$ \\
\bottomrule
\end{tabular*}
\par\vspace{0.6em}
\begin{tabular*}{\linewidth}{@{\extracolsep{\fill}}lccccccc@{}}
\toprule
\multicolumn{8}{c}{\textbf{8K context (continued)}} \\
Method & MQ & VT & CWE & FWE & QA1 & QA2 & Avg \\
\midrule
BF16/FP32 MP & $29.7\pmstd{11.7}$ & $1.4\pmstd{0.2}$ & $19.9\pmstd{3.0}$ & $9.8\pmstd{7.1}$ & $37.8\pmstd{5.0}$ & $26.8\pmstd{0.6}$ & $37.2\pmstd{3.1}$ \\
Naive FP8 & $34.0\pmstd{4.2}$ & $1.3\pmstd{0.7}$ & $18.4\pmstd{1.5}$ & $8.1\pmstd{2.8}$ & $33.4\pmstd{0.4}$ & $25.1\pmstd{0.1}$ & $37.9\pmstd{0.8}$ \\
Delta-Matching (\textit{ours}) & $33.6\pmstd{2.5}$ & $2.1\pmstd{1.3}$ & $8.0\pmstd{1.5}$ & $10.1\pmstd{4.7}$ & $33.9\pmstd{1.5}$ & $24.9\pmstd{0.3}$ & $36.0\pmstd{1.2}$ \\
\bottomrule
\end{tabular*}
\end{benchmarktable}

\begin{benchmarktable}
\centering
\caption{
\textbf{Per-task RULER scores for the Mamba-2/GQA hybrid.}
4K context.
Results correspond to the Mamba-2/GQA rows in Table~\ref{tab:exp_model_arch}.
Mean $\pm$ standard error over two initialization seeds; scores are percentages.
}
\label{tab:app_ruler_mamba}
\scriptsize
\setlength{\tabcolsep}{2pt}
\begin{tabular*}{\linewidth}{@{\extracolsep{\fill}}lccccccc@{}}
\toprule
\multicolumn{8}{c}{\textbf{4K context}} \\
Method & S1 & S2 & S3 & MK1 & MK2 & MK3 & MV \\
\midrule
BF16/FP32 MP & $100.0\pmstd{0.0}$ & $100.0\pmstd{0.0}$ & $76.0\pmstd{3.0}$ & $94.1\pmstd{0.1}$ & $60.9\pmstd{10.7}$ & $8.3\pmstd{2.7}$ & $81.6\pmstd{8.6}$ \\
Naive FP8 & $100.0\pmstd{0.0}$ & $99.9\pmstd{0.1}$ & $70.4\pmstd{21.2}$ & $94.7\pmstd{2.1}$ & $79.5\pmstd{17.7}$ & $17.5\pmstd{5.5}$ & $77.9\pmstd{18.8}$ \\
Delta-Matching (\textit{ours}) & $100.0\pmstd{0.0}$ & $99.9\pmstd{0.1}$ & $74.0\pmstd{16.0}$ & $93.4\pmstd{1.2}$ & $76.6\pmstd{13.4}$ & $16.1\pmstd{7.9}$ & $72.1\pmstd{12.0}$ \\
\bottomrule
\end{tabular*}
\par\vspace{0.6em}
\begin{tabular*}{\linewidth}{@{\extracolsep{\fill}}lccccccc@{}}
\toprule
\multicolumn{8}{c}{\textbf{4K context (continued)}} \\
Method & MQ & VT & CWE & FWE & QA1 & QA2 & Avg \\
\midrule
BF16/FP32 MP & $70.3\pmstd{8.5}$ & $2.6\pmstd{0.5}$ & $32.3\pmstd{7.6}$ & $19.3\pmstd{1.0}$ & $50.6\pmstd{3.4}$ & $31.4\pmstd{1.8}$ & $56.0\pmstd{1.6}$ \\
Naive FP8 & $78.6\pmstd{11.7}$ & $15.7\pmstd{12.7}$ & $53.3\pmstd{9.5}$ & $29.6\pmstd{8.7}$ & $54.2\pmstd{1.8}$ & $31.3\pmstd{1.3}$ & $61.8\pmstd{0.1}$ \\
Delta-Matching (\textit{ours}) & $62.6\pmstd{19.7}$ & $14.3\pmstd{5.2}$ & $32.9\pmstd{0.4}$ & $30.8\pmstd{13.9}$ & $49.5\pmstd{4.7}$ & $31.6\pmstd{1.0}$ & $58.0\pmstd{1.4}$ \\
\bottomrule
\end{tabular*}
\end{benchmarktable}
\begin{benchmarktable}
\centering
\ContinuedFloat
\caption{\textbf{Per-task RULER scores for the Mamba-2/GQA hybrid (continued).} 8K context.}
\scriptsize
\setlength{\tabcolsep}{2pt}
\begin{tabular*}{\linewidth}{@{\extracolsep{\fill}}lccccccc@{}}
\toprule
\multicolumn{8}{c}{\textbf{8K context}} \\
Method & S1 & S2 & S3 & MK1 & MK2 & MK3 & MV \\
\midrule
BF16/FP32 MP & $100.0\pmstd{0.0}$ & $100.0\pmstd{0.0}$ & $67.7\pmstd{4.7}$ & $82.6\pmstd{9.0}$ & $17.3\pmstd{0.1}$ & $1.2\pmstd{1.0}$ & $66.9\pmstd{6.4}$ \\
Naive FP8 & $99.8\pmstd{0.2}$ & $99.4\pmstd{0.6}$ & $69.6\pmstd{1.8}$ & $80.0\pmstd{4.4}$ & $68.7\pmstd{19.9}$ & $2.5\pmstd{1.3}$ & $55.7\pmstd{0.3}$ \\
Delta-Matching (\textit{ours}) & $100.0\pmstd{0.0}$ & $99.9\pmstd{0.1}$ & $77.2\pmstd{2.8}$ & $79.6\pmstd{1.6}$ & $49.4\pmstd{20.6}$ & $4.4\pmstd{3.6}$ & $46.7\pmstd{9.2}$ \\
\bottomrule
\end{tabular*}
\par\vspace{0.6em}
\begin{tabular*}{\linewidth}{@{\extracolsep{\fill}}lccccccc@{}}
\toprule
\multicolumn{8}{c}{\textbf{8K context (continued)}} \\
Method & MQ & VT & CWE & FWE & QA1 & QA2 & Avg \\
\midrule
BF16/FP32 MP & $56.5\pmstd{3.0}$ & $3.8\pmstd{2.8}$ & $19.3\pmstd{2.6}$ & $25.0\pmstd{5.3}$ & $35.7\pmstd{0.5}$ & $27.5\pmstd{1.7}$ & $46.5\pmstd{1.2}$ \\
Naive FP8 & $48.6\pmstd{2.8}$ & $5.8\pmstd{4.5}$ & $18.2\pmstd{0.4}$ & $25.7\pmstd{1.4}$ & $37.0\pmstd{1.2}$ & $26.2\pmstd{1.0}$ & $49.0\pmstd{1.3}$ \\
Delta-Matching (\textit{ours}) & $47.7\pmstd{15.6}$ & $4.8\pmstd{0.7}$ & $19.1\pmstd{0.1}$ & $31.9\pmstd{5.2}$ & $35.4\pmstd{3.4}$ & $27.3\pmstd{1.3}$ & $48.0\pmstd{0.6}$ \\
\bottomrule
\end{tabular*}
\end{benchmarktable}

\begin{benchmarktable}
\centering
\caption{
\textbf{Per-task RULER scores across optimizers.}
4K context.
Results correspond to the single-initialization runs in
Table~\ref{tab:app_muon_full} (Appendix~\ref{app:muon_full_training}).
Scores are percentages and carry no across-seed standard errors.
}
\label{tab:app_ruler_muon}
\scriptsize
\setlength{\tabcolsep}{2pt}
\begin{tabular*}{\linewidth}{@{\extracolsep{\fill}}lccccccc@{}}
\toprule
\multicolumn{8}{c}{\textbf{4K context}} \\
Method & S1 & S2 & S3 & MK1 & MK2 & MK3 & MV \\
\midrule
\multicolumn{8}{l}{\textbf{Muon}} \\
BF16/FP32 MP & $100.0$ & $100.0$ & $85.0$ & $92.8$ & $91.4$ & $27.2$ & $75.8$ \\
Naive FP8 & $98.6$ & $99.8$ & $37.6$ & $55.4$ & $9.6$ & $1.4$ & $22.2$ \\
Delta-Matching (\textit{ours}) & $100.0$ & $89.4$ & $85.4$ & $90.4$ & $91.2$ & $28.8$ & $85.5$ \\
\midrule
\multicolumn{8}{l}{\textbf{AdamW}} \\
BF16/FP32 MP & $100.0$ & $100.0$ & $60.8$ & $98.2$ & $94.2$ & $25.2$ & $89.9$ \\
Naive FP8 & $97.8$ & $94.2$ & $16.2$ & $29.2$ & $6.0$ & $0.6$ & $15.0$ \\
Delta-Matching (\textit{ours}) & $100.0$ & $100.0$ & $87.6$ & $92.6$ & $63.8$ & $20.2$ & $87.9$ \\
\bottomrule
\end{tabular*}
\par\vspace{0.6em}
\begin{tabular*}{\linewidth}{@{\extracolsep{\fill}}lccccccc@{}}
\toprule
\multicolumn{8}{c}{\textbf{4K context (continued)}} \\
Method & MQ & VT & CWE & FWE & QA1 & QA2 & Avg \\
\midrule
\multicolumn{8}{l}{\textbf{Muon}} \\
BF16/FP32 MP & $77.9$ & $25.8$ & $33.8$ & $22.9$ & $44.8$ & $29.6$ & $62.1$ \\
Naive FP8 & $28.6$ & $3.0$ & $14.5$ & $4.9$ & $27.8$ & $19.0$ & $32.5$ \\
Delta-Matching (\textit{ours}) & $78.9$ & $24.2$ & $37.0$ & $1.3$ & $53.2$ & $29.8$ & $61.2$ \\
\midrule
\multicolumn{8}{l}{\textbf{AdamW}} \\
BF16/FP32 MP & $75.5$ & $3.0$ & $65.7$ & $44.8$ & $49.6$ & $32.0$ & $64.5$ \\
Naive FP8 & $16.7$ & $0.6$ & $20.8$ & $1.8$ & $16.2$ & $12.0$ & $25.2$ \\
Delta-Matching (\textit{ours}) & $80.3$ & $11.7$ & $23.7$ & $28.9$ & $52.4$ & $33.0$ & $60.2$ \\
\bottomrule
\end{tabular*}
\end{benchmarktable}

\begin{benchmarktable}
\centering
\ContinuedFloat
\caption{\textbf{Per-task RULER scores across optimizers (continued).} 8K context.}
\scriptsize
\setlength{\tabcolsep}{2pt}
\begin{tabular*}{\linewidth}{@{\extracolsep{\fill}}lccccccc@{}}
\toprule
\multicolumn{8}{c}{\textbf{8K context}} \\
Method & S1 & S2 & S3 & MK1 & MK2 & MK3 & MV \\
\midrule
\multicolumn{8}{l}{\textbf{Muon}} \\
BF16/FP32 MP & $100.0$ & $100.0$ & $93.8$ & $85.0$ & $75.8$ & $13.2$ & $49.5$ \\
Naive FP8 & $97.8$ & $93.0$ & $5.6$ & $48.8$ & $1.6$ & $0.2$ & $24.3$ \\
Delta-Matching (\textit{ours}) & $100.0$ & $100.0$ & $99.4$ & $87.6$ & $88.0$ & $13.4$ & $74.2$ \\
\midrule
\multicolumn{8}{l}{\textbf{AdamW}} \\
BF16/FP32 MP & $100.0$ & $100.0$ & $66.4$ & $88.4$ & $80.8$ & $7.2$ & $61.6$ \\
Naive FP8 & $95.4$ & $90.8$ & $7.4$ & $26.0$ & $1.8$ & $0.0$ & $14.8$ \\
Delta-Matching (\textit{ours}) & $100.0$ & $100.0$ & $67.4$ & $82.0$ & $23.4$ & $2.2$ & $60.2$ \\
\bottomrule
\end{tabular*}
\par\vspace{0.6em}
\begin{tabular*}{\linewidth}{@{\extracolsep{\fill}}lccccccc@{}}
\toprule
\multicolumn{8}{c}{\textbf{8K context (continued)}} \\
Method & MQ & VT & CWE & FWE & QA1 & QA2 & Avg \\
\midrule
\multicolumn{8}{l}{\textbf{Muon}} \\
BF16/FP32 MP & $57.8$ & $8.9$ & $19.8$ & $22.3$ & $37.0$ & $26.6$ & $53.0$ \\
Naive FP8 & $23.8$ & $1.0$ & $6.0$ & $2.6$ & $20.2$ & $16.2$ & $26.2$ \\
Delta-Matching (\textit{ours}) & $60.8$ & $9.0$ & $23.7$ & $39.5$ & $39.8$ & $28.2$ & $58.7$ \\
\midrule
\multicolumn{8}{l}{\textbf{AdamW}} \\
BF16/FP32 MP & $59.6$ & $5.2$ & $32.5$ & $42.6$ & $38.0$ & $26.8$ & $54.5$ \\
Naive FP8 & $11.2$ & $0.4$ & $9.4$ & $2.2$ & $11.6$ & $13.4$ & $21.9$ \\
Delta-Matching (\textit{ours}) & $64.3$ & $7.2$ & $9.4$ & $45.3$ & $38.4$ & $28.6$ & $48.3$ \\
\bottomrule
\end{tabular*}
\end{benchmarktable}

\let\addcontentsline\appendixoriginaladdcontentsline

\end{document}